\documentclass[journal]{IEEEtran}
\usepackage{amsmath,amssymb,amsfonts}
\usepackage{algorithmic}
\usepackage{algorithm}
\usepackage{graphicx}
\usepackage{textcomp}
\usepackage{subfig}
\usepackage{soul}

\DeclareMathOperator*{\argmin}{arg\,min}

\ifCLASSINFOpdf
\else
\fi
\begin{document}
%
\title{Non-learning Stereo-aided Depth Completion under Mis-projection via Selective Stereo Matching}
%
%
%

\author{Yasuhiro~Yao, Ryoichi~Ishikawa, Shingo~Ando, Kana~Kurata, Naoki~Ito, Jun~Shimamura, and Takeshi~Oishi
\thanks{Yasuhiro Yao was with the Institute of Industrial Science, the University of Tokyo, Tokyo 153-0041, Japan, and also with the Human Informatics Laboratories, Nippon Telegraph and Telephone Corporation, Tokyo 100-0004, Japan (e-mail: yao@cvl.iis.u-tokyo.ac.jp).

Ryoichi Ishikawa and Takeshi Oishi were with the Institute of Industrial Science, the University of Tokyo, Tokyo 153-0041, Japan (e-mail: ishikawa@cvl.iis.u-tokyo.ac.jp; oishi@cvl.iis.u-tokyo.ac.jp).

Shingo Ando, Kana Kurata, Naoki Ito, and Jun Shimamura were with the Human Informatics Laboratories, Nippon Telegraph and Telephone Corporation, Tokyo 100-0004, Japan (e-mail: shingo.andou.fv@hco.ntt.co.jp; jun.shimamura.ec@hco.ntt.co.jp).}}

%
%

\markboth{IEEE Access,~Vol.~9, October~2021}%
{Yao \MakeLowercase{\textit{et al.}}: Non-learning Stereo-aided Depth Completion under Mis-projection via Selective Stereo Matching}
%



\maketitle

\begin{abstract}
We propose a non-learning depth completion method for a sparse depth map captured using a light detection and ranging (LiDAR) sensor guided by a pair of stereo images.
Generally, conventional stereo-aided depth completion methods have two limiations.
(i) They assume the given sparse depth map is accurately aligned to the input image, whereas the alignment is difficult to achieve in practice.
(ii) They have limited accuracy in the long range because the depth is estimated by pixel disparity.
To solve the abovementioned limitations, we propose selective stereo matching (SSM) that searches the most appropriate depth value for each image pixel from its neighborly projected LiDAR points based on an energy minimization framework. 
This depth selection approach can handle any type of mis-projection.
Moreover, SSM has an advantage in terms of long-range depth accuracy because it directly uses the LiDAR measurement rather than the depth acquired from the stereo. 
SSM is a discrete process; thus, we apply variational smoothing with binary anisotropic diffusion tensor (B-ADT) to generate a continuous depth map while preserving depth discontinuity across object boundaries.
Experimentally, compared with the previous state-of-the-art stereo-aided depth completion, the proposed method reduced the mean absolute error (MAE) of the depth estimation to 0.65 times and demonstrated approximately twice more accurate estimation in the long range.
Moreover, under various LiDAR-camera calibration errors, the proposed method reduced the depth estimation MAE to 0.34-0.93 times from previous depth completion methods.
\end{abstract}


%
\IEEEpeerreviewmaketitle

\section{Introduction}
%
%
%
%

\begin{figure}
    \centering
    \includegraphics[width = \linewidth]{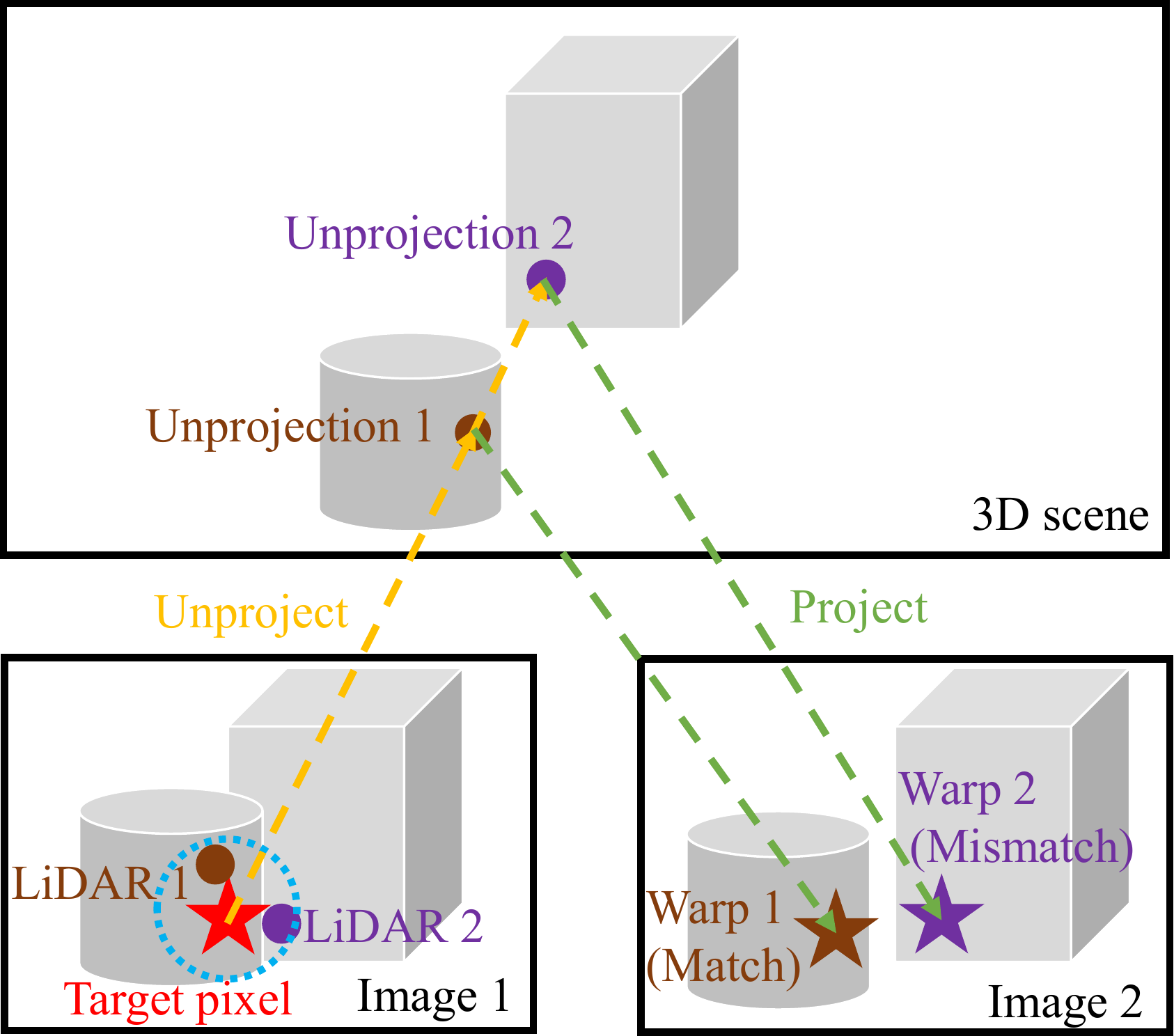}
    \caption{
    Conceptual diagram of SSM.
    SSM selects the most appropriate depth for each pixel from its nearby LiDAR projections within a search radius based on stereo correspondence and smoothness.
    The diagram ignores the smoothness for simplicity.
    Here, two depths (LiDAR 1 and 2) exist within the search radius from the target pixel in image 1.
    SSM warps the target pixel to image 2 using the depths of LiDAR 1 and 2.
    Then, SSM evaluates the correspondence of image 1 at the target pixel and image 2 at the warps (Warp 1 and 2).
    Here, the correspondence is higher for Warp 1, in other words, when LiDAR 1 is selected.
    }
    \label{fig:ssm_overview}
\end{figure}
\IEEEPARstart{D}{epth} measurement is conducted in several ways such as time-of-flight (ToF), stereo cameras, and structured light projection \cite{batlle1998recent}.
Stereo cameras and structured light projection estimate depth by pixel disparity.
Hence their precision dramatically reduces as the distance increases since a small disparity change indicates a large depth change in the long range.
In comparison, ToF sensors have a higher precision in the long range.
Among ToF sensors, light detection and ranging (LiDAR) is used in various systems that require adaptability to dynamic environments, e.g., automated driving and robots, because of its active sensing capability and robustness to environmental changes.
However, in terms of measurement density, LiDAR has a limitation because of the number of lasers in its array and the narrow beam measurement.

An alternative to address the sparsity of LiDAR is the depth completion, and the most common approach uses a single synchronized image as a guide \cite{kopf2007joint, ferstl2013image, diebel2006application, schneider2016semantically, ma2019self, wong2020unsupervised, yao2020discontinuous, liu2021fcfr, park2020non, cheng2020cspn++, zhao2021adaptive, chen2019learning, qiu2019deeplidar, li2020multi, van2019sparse, xu2019depth, yan2020revisiting, eldesokey2019confidence, schuster2021ssgp, bai2020depthnet, shivakumar2019dfusenet}.
These methods generate a sparse depth map by projecting LiDAR points onto the image, and then the depth map is completed using pixel intensities.
Such image-aided depth completion methods are extensively studied and range from non-learning to supervised methods. 

An important issue in the depth completion is mis-projection where LiDAR points are projected onto different objects in the image. 
Mis-projection often occurs because of spatial and temporal displacement of sensors, dynamic objects, decalibration, and calibration errors. 
Because camera and LiDAR are typically placed at different position, occlusion is unavoidable for near or dynamic objects.
The temporal displacement of each LiDAR beam also causes errors without precise synchronization.
Decalibration appears at run time because of oscillations of the vehicle or other mechanical reasons.
Furthermore, the extrinsic calibration error between sensors results in mis-projection over the entire image.

Generally, extrinsic LiDAR and camera calibration is difficult because of their different modalities. 
To build the KITTI dataset \cite{geiger2013vision}, Geiger et al. calibrated LiDAR and cameras using multiple calibration boards in a controlled garage environment \cite{geiger2012automatic} followed by the manual selection of corresponding points.
Although marker-less calibration methods have been examined \cite{pandey2012automatic, ishikawa2018lidar, john2015automatic}, it remains difficult to stably realize accurate extrinsic calibration in uncontrolled environments.

In recent studies, stereo images have been used for depth completion to solve the mis-projection issue rather than a single image.
This is because stereo camera systems are widely available and such systems can perceive 3D information with the help of stereo matching algorithms.
For example, a self-supervised method is applicable under mis-projection caused by displacements of sensors\cite{cheng2019noise}.
However, this method still requires a dataset measured with a well-calibrated LiDAR-stereo system to train the neural network (NN).
Furthermore, this method estimates depth by pixel disparity and suffers from low depth precision in the long range.

Therefore, in this study, we propose a non-learning depth completion method for a stereo-LiDAR system that is effective for the long range and robust to mis-projection.
The proposed method comprises two techniques, i.e., selective stereo matching (SSM) and binary anisotropic diffusion tensor (B-ADT) \cite{yao2020discontinuous}-aided smoothing.
An important proposal is SSM, which searches for an optimal depth value for each pixel from its neighborly projected LiDAR points using an energy minimization framework (Fig. \ref{fig:ssm_overview}).
This energy minimization approach can handle any type of mis-projection.
Furthermore, SSM directly uses LiDAR depths and is advantageous in long-range accuracy. 
SSM is discrete optimization; thus, we apply B-ADT-aided smoothing for continuous depth estimation while preserving discontinuity between different objects. 

Our contributions are summarized as follows.
\begin{itemize}
    \item
    We propose SSM, which performs stereo matching in a selective manner to upsample LiDAR depths while maintaining the depth precision of LiDAR in the long range and considering mis-projection of LiDAR points.
    \item We propose a non-learning depth completion framework that combines SSM and B-ADT-aided smoothing.
    The framework achieves boundary-aware continuous depth estimation in addition to the SSM effects (long-range depth accuracy and robustness to mis-projection).
\end{itemize}

The rest of the paper is organized as follows.
Section \ref{sec:related} reviews the related works and limitations of existing depth completion methods.
Section \ref{sec:proposed} explains the proposed method.
Section \ref{sec:eval} shows the evaluations.
Section \ref{sec:limitation} gives the conclusion and summarizes limitations and future works.
\section {Related work}
\label{sec:related}
\subsection{Stereo matching}
\begin{figure*}[t]
    \centering
    \includegraphics[width=\linewidth]{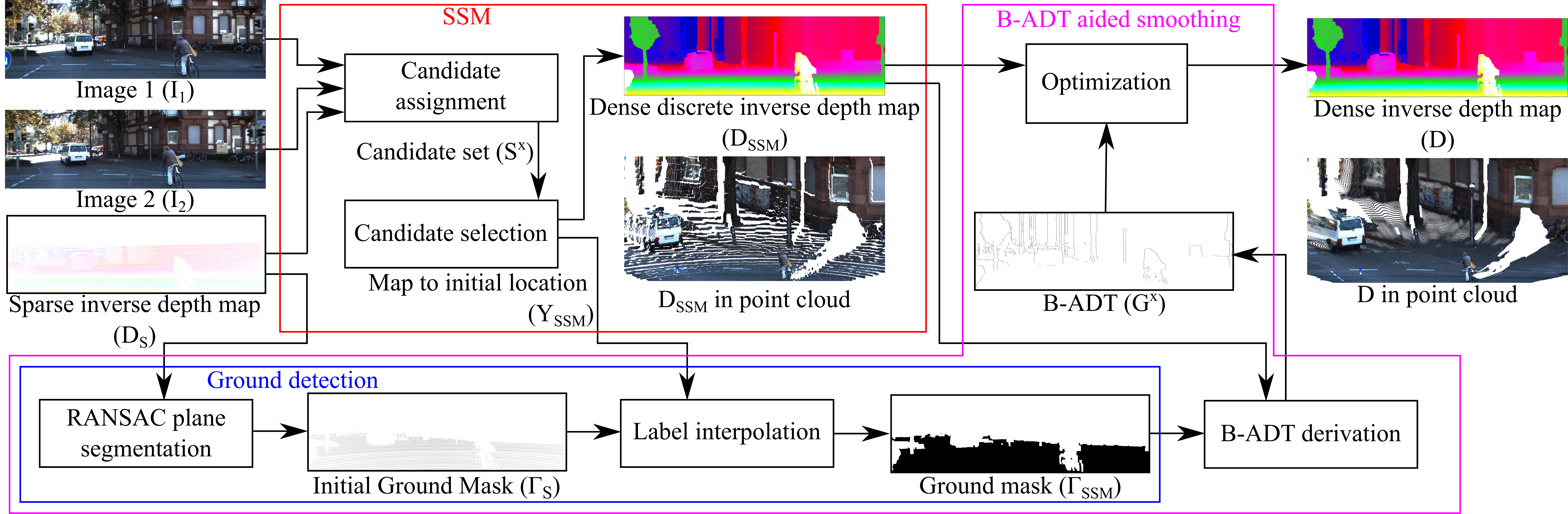}
    \caption{
    Proposed Framework.
    SSM is composed of candidate assignment and candidate selection via optimization.
    The candidate assignment takes stereo images ($I_1$ and $I_2$) and a sparse inverse depth map ($D_S$) to derive a candidate set ($S^{\mathbf{x}}$).
    The candidate selection via optimization derives a dense discrete inverse depth map ($D_{\mathrm{SSM}}$) and a map to the initial inverse depth location ($Y_{\mathrm{SSM}}$).
    B-ADT-aided smoothing first performs the ground detection to derive the ground mask ($\Gamma_{\mathrm{SSM}})$, then derives B-ADT($G^{\mathbf{x}})$, and finally performs optimization to derive a dense inverse depth map ($D$).
    }
    \label{fig:flow}
\end{figure*}
Stereo matching is extensively studied for 3D scanning because of the availability of stereo camera systems.
The methods span from non-learning \cite{zhang2005parameter, hernandez2016embedded, yamaguchi2014efficient, muresan2015improving, spangenberg2014large, zureiki2007stereo} to NN-based self-supervised \cite{wang2021pvstereo} and supervised \cite{cheng2020hierarchical} methods.

In terms of accuracy, the supervised methods perform the best among them.
The supervised methods also have the potential to perform in the long range because the precision of the ground truth disparity, which is usually made by other sensors such as LiDAR, is high and sub-pixel-level accurate.
However, it is challenging to prepare a large dataset with ground truth disparity to train NNs.

Non-learning and self-supervised methods have difficulties achieving high precision in the long range because their depth estimation is limited by pixel-level stereo matching.
As mentioned, a small disparity change indicates a large depth change in the long range.
Therefore, there is uncertainty in the depth estimation even if the matching is accurate at the pixel level.
Moreover, stereo matching methods still suffer from challenging scenarios such as repetitive pattern, low texture, discontinuity to cause occlusion, and specular reflection conditions.
\subsection{Single-image-aided depth completion}
Depth completion methods generate high-resolution and dense depth maps from sparse or low-resolution depth maps captured using LiDAR or depth cameras.
The most common approach uses a single image as guidance.
Kopf et al. \cite{kopf2007joint} proposed a method to interpolate low-resolution depth values based on the joint distance of color and space in a high-resolution image.
Diebel and Thrun performed upsampling using a Markov random field (MRF) formulation \cite{diebel2006application}.
In this method, the smoothness term is weighted as per texture derivatives; however, the results suffer from surface over-flattening.
To address this issue, Ferstl et al. formalized depth completion into ADT-aided and TGV-regularized energy minimization \cite{ferstl2013image}; and their method has been successfully used to smooth and optimize depth maps in more recent methods \cite{chen2016transforming, hirata2019real}.
Recently, Yao et al. proposed B-ADT to achieve depth completion to preserve discontinuity between different objects \cite{yao2020discontinuous}.

In addition, NNs have been applied to depth completion tasks.
The most common approach is to train networks with ground truth dense depth maps \cite{liu2021fcfr, park2020non, cheng2020cspn++, zhao2021adaptive, chen2019learning, qiu2019deeplidar, li2020multi, van2019sparse, xu2019depth, yan2020revisiting, eldesokey2019confidence, schuster2021ssgp, bai2020depthnet, shivakumar2019dfusenet}.
Recently, self-supervised and semi-supervised methods have been examined because it is difficult to acquire the dense ground truth.
Ma et al. \cite{ma2019self} and Wong et al. \cite{wong2020unsupervised} proposed methods with NNs that can be self-supervised using monocular camera frames and sparse depth maps from LiDAR with motion.
Yang et al. proposed a method that can train a NN by the likelihood of the observed sparse point cloud under a hypothesized depth map \cite{yang2019dense}.

A major limitation of the single-image-aided depth completion is that mis-projection of LiDAR points is not considered.
\subsection{Stereo-aided depth completion}
Stereo images have been used as guides to complete the sparse measurements of LiDAR.
These methods have been developed based on stereo matching, and they perform dense stereo matching using the accurate sparse depth value by LiDAR as a clue.

Badino et al. used LiDAR measurements to reduce the search space for stereo matching and provided predefined paths for dynamic programming \cite{huber2011integrating}.
Maddern et al. proposed a probabilistic model to fuse LiDAR and disparities by combining prior from each sensor \cite{maddern2016real}, and Park et al. used NNs to learn such a model, which takes two disparities as input, i.e., one from the interpolated LiDAR and the other from semi-global matching \cite{park2018high}.
Choe et al. recently proposed a geometry-aware stereo-LiDAR fusion network for long-range depth estimation \cite{choe2021volumetric}.
As the same as single-image-aided methods, these methods do not consider mis-projection in given sparse depth maps.

Several recent methods have attempted to infer dense disparity maps from inaccurately projected LiDAR points with the help of stereo images.
For example, Cheng et al. proposed a self-supervised method to train a NN to remove occluded background projection of LiDAR points to infer dense disparity maps \cite{cheng2019noise}; however, this method does not handle incorrect projection caused by extrinsic calibration errors between the LiDAR and camera.
Park et al. proposed a supervised method to train a NN to infer dense disparity maps from LiDAR inputs with extrinsic calibration errors between LiDAR and the camera\cite{park2019high}; however, this method requires accurately calibrated LiDAR and cameras to acquire effective training data.

Furthermore, the previous non-learning and self-supervised methods \cite{huber2011integrating, maddern2016real, cheng2019noise} estimate the depth by pixel disparity; thus, their depth precision is limited in the long range.

In summary, there are two major limitations in existing stereo-aided depth completion methods.
\begin{itemize}
\item These methods require accurate LiDAR-camera extrinsic calibration at some point in their process, which is often difficult to realize.
\item The precision in the depth estimation is dramatically reduced as the distance increases because of the nature of disparity estimation using images.
\end{itemize}

The cost of the proposed method is similar to the previous studies \cite{zhang2005parameter, cheng2019noise}, whereas the approach to minimize the cost is different.
The proposed method searches the minimizer by the selection from projected LiDAR depth values.
The approach can handle any type of mis-projection without requiring accurate LiDAR-camera extrinsic calibration in any part of the process.
Moreover, this selective approach has an advantage in the long-range precision because it directly uses LiDAR depth values.
\section{Proposed method}
\label{sec:proposed}
\begin{figure}[t]
    \centering
    \includegraphics[width=\linewidth]{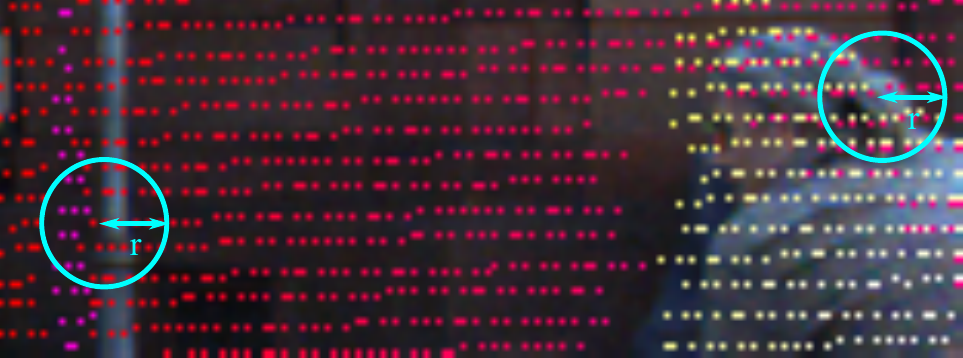}
    \caption{
    Candidate assignment of SSM.
    The circles indicate the areas to create the candidate sets for the centered pixels.
    Although there is mis-projection, the appropriate LiDAR values are located in the areas, e.g., purple for the pole at the left, and yellow for the person at the right.
    }
    \label{fig:concept}
\end{figure}
As shown in Fig. \ref{fig:flow}, the proposed method applies SSM followed by B-ADT-aided smoothing\cite{yao2020discontinuous}.
SSM is a discrete optimization, and its output is discrete; thus, smoothing improves the quality of the result.
B-ADT allows us to incorporate boundary-direction-aware discontinuity in a variational approach.
Both SSM and B-ADT-aided smoothing; thus, the proposed method as a whole, preserve depth discontinuity between different objects.

We give the problem statement in Section \ref{sec:prob}, introduce SSM in Section \ref{sec:ssm}, explain B-ADT-aided smoothing in Section \ref{sec:b-adt}, and describe a practical parameter tuning approach for SSM in Section \ref{sec:param}.
\subsection{Problem statement}
\label{sec:prob}
Our problem settings assume a stereo camera and LiDAR are used to capture the scene; however, the camera and LiDAR calibration contains errors.
Such conditions are possibly occur because of the difficulty associated with calibration, particularly when a calibration target is not available.
Our aim is to estimate a dense depth map that is aligned with the image.
Using mathematical notations, the target problem is defined as follows.

We are given a pair of stereo images ($I_1:\Omega_1 \to \mathbb{R}$, $I_2:\Omega_2 \to \mathbb{R}$) and a sparse inverse depth map captured by LiDAR ($D_S:\Omega_1 \to \mathbb{R}\bigcup\left\{\phi\right\} $) with $\Omega_1\subset \mathbb{R}^2$ and $\Omega_2\subset \mathbb{R}^2$ being the image domains and $\phi$ indicating an empty depth.
Here, $D_S$ is determined by projecting the LiDAR points to the image $I_1$; however, $D_S$ is not accurately aligned with $I_1$ because we expect LiDAR-camera miscalibration and occlusions.
Our aim is to derive a dense inverse depth map $D:\Omega_1 \to \mathbb{R}$ aligned with the input image $I_1$.
Throughout this paper, we normalize $I_1$ and $I_2$ to the range $\left[0,1\right]$, and the unit of the depth is meter.

Note that we derive an inverse depth map $D$ rather than directly deriving the depth map.
This is performed to balance the contribution of both near and distant depths \cite{newcombe2011dtam, yao2020discontinuous}. 
Deriving a dense inverse depth map $D$ is equivalent to deriving a dense depth map $D^{-1}$ or dense disparity map $D^{\prime}$.
The conversions at $\mathbf{x} \in \Omega_1$ are given as Eq. \eqref{eq:conv_depth}, \eqref{eq:conv_disp} using camera focal length $f$ and stereo baseline $b$.
\begin{align}
\label{eq:conv_depth}
    D^{-1}(\mathbf{x}) & = D(\mathbf{x}) ^ {-1}\\
\label{eq:conv_disp}
    D^{\prime}(\mathbf{x}) & = fb D(\mathbf{x}) 
\end{align}

The proposed method is applicable to motion stereo images as input as in the evaluation using the Komaba dataset (Section \ref{sec:eval_kitti_error}).

In this paper, we use $|\cdot|$ to denote the vector norm.
In particular, given a vector $\mathbf{p}\in\mathbf{R}^K$ with $K$ being the arbitrary number of dimensions, the norm is given as follows:
\begin{equation}
|\mathbf{p}| = \sqrt{\sum_{i}^K{p_i}^2}
\end{equation}
\subsection{Selective Stereo Matching}
\label{sec:ssm}
SSM searches the most appropriate inverse depth value for every $\mathbf{x} \in \Omega_1$ from its neighborly projected LiDAR points.
SSM comprises the candidate assignment and candidate selection steps.
In the candidate assignment step, each pixel in the image is assigned a set of LiDAR inverse depth values.
In the candidate selection step, SSM selects the most appropriate value from the candidate set using an energy minimization framework.
Here, the energy is defined as the sum of the stereo matching cost and the smoothness regularization term.

Implementation-wise, both the candidate assignment and the candidate selection are composed of pixel-wise calculations which are parallelized on GPU. 
\subsubsection{Candidate assignment}
\paragraph{Initial map of candidate sets}
First, SSM constructs a candidate set $S^{\mathbf{x}}$ ($\mathbf{x} \in \Omega_1$), which is a set of inverse depth values in the surrounding pixels within a pre-defined radius $r$ from $\mathbf{x}$ (Eq. \eqref{eq:set}), as shown in Fig. \ref{fig:concept}.
\begin{align}
\label{eq:set}
    S^{\mathbf{x}} = \{d \mid & d = D_S(\mathbf{y})\land d\neq \phi \mathrm{~where~} \left|\mathbf{y} - \mathbf{x} \right| < r \}
\end{align}
Note that we introduce an empirical approach to set $r$ in Section \ref{sec:param}.
If the cardinality of $S^{\mathbf{x}}$ is less than predefined threshold $m$, the set is assumed to be empty ($S^{\mathbf{x}} = \emptyset$) to avoid selecting a value from a small number of candidates (we used $m=4$ in our evaluations).
\paragraph{Candidate set interpolation}
To fill the pixels with non-empty candidate sets, we interpolate the candidate sets using image-guided nearest neighbor search (IGNNS) \cite{yao2020discontinuous}.
IGNNS searches the nearest neighbor by the cumulative distance of image gradients.
Here, let $\Psi$ be the set of pixels where the candidate set is empty ($\Psi=\left\{ \mathbf{x} \mid S^\mathbf{x} = \emptyset \right\}$), and let $\bar{\Psi}$ be the complement of $\Psi$ ($\bar{\Psi} = \Omega_1 \backslash \Psi$).
We search an image-guided nearest neighbor of every $\mathbf{x} \in \Psi$ from $\bar{\Psi}$ (denoted $\mathbf{z}$) and update the candidate set by $S^{\mathbf{x}} = S^{\mathbf{z}}$.

We derive $\mathbf{z}$ as Eq. \eqref{eq_nearest} by denoting $\pi(\bar{\mathbf{x}},\mathbf{x})$ being the set of pixels on the grid path from a surrounding non-empty pixel $\bar{\mathbf{x}} \in \bar{\Psi}$ to $\mathbf{x}$.
\begin{equation}
\label{eq_nearest}
\mathbf{z} = \argmin_{\bar{\mathbf{x}} \in \bar{\Psi}} \min_{\pi(\bar{\mathbf{x}},\mathbf{x})} 
{\sum_{\mathbf{y}\in \pi(\bar{\mathbf{x}},\mathbf{x})} \left\{ \left| \nabla I_1 (\mathbf{y}) \right|^2 +c \right\}},
\end{equation}
where $c$ is the constant cost of the unit path length, and we set $c=0.04$ in our evaluations following the parameter study in the literature \cite{yao2020discontinuous}.
\paragraph{Correspondence search}
Following candidate set interpolation, we identify the correspondence $\mathbf{x}^{\prime}(d) \in \Omega_2$ of $\mathbf{x}\in \Omega_1$.
$\mathbf{x}^{\prime}(d)$ is the location on $I_2$ where $\mathbf{x}$ on $I_1$ is warped with inverse depth value $d$.
We calculate $\mathbf{x}^{\prime}(d)$ for all combinations of $\mathbf{x}$ and $d \in S^{\mathbf{x}}$.
Below we denote $\mathbf{x} = (x_0\:x_1)^T$.

If $I_1$ and $I_2$ are a pair of rectified binocular stereo images, using the floor function denoted as $\lfloor \cdot \rfloor$, $\mathbf{x}^{\prime}(d)$ is derived with camera focal length $f$ and baseline $b$ as follows:
\begin{equation}
\label{eq:warp_stereo}
    \mathbf{x}^{\prime}(d) = \left( \begin{array}{c} \lfloor x_0 - fbd \rfloor \\ x_1 \end{array} \right).
\end{equation}

For motion stereo images, $\mathbf{x}^{\prime}(d)$ is calculated using the camera intrinsic parameter ($K\in\mathbb{R}^{3\times3}$), rotation ($R_C \in SO(3)$), and the translation ($t_C \in \mathbb{R}^3$) of the camera motion as follows:
\begin{align}
\label{eq:warp_motion}
    &\mathbf{x}^{\prime}(d) = \left( \begin{array}{c} \lfloor \tilde{x}^{\prime}_0(d) / \tilde{x}^{\prime}_2(d) \rfloor \\ \lfloor \tilde{x}^{\prime}_1(d) / \tilde{x}^{\prime}_2(d) \rfloor \end{array} \right) \mathrm{,} \nonumber\\
    &\mathrm{where\:}
    \left( \begin{array}{c} \tilde{x}^{\prime}_0(d) \\ \tilde{x}^{\prime}_1(d) \\ \tilde{x}^{\prime}_2(d) \end{array} \right) 
    = K \left( R_C K^{-1} \left( \begin{array}{c} x_0 \\ x_1 \\ d^{-1} \end{array} \right) + \mathbf{t}_C \right).
\end{align}
If there are two or more $d \in S^{\mathbf{x}}$ to derive the same $\mathbf{x}^{\prime}(d)$, we only maintain the nearest from \textbf{x} among those in $S^{\mathbf{x}}$.
Note that this pruning process is performed to realize computational efficiency of the following optimization process.

\subsubsection{Candidate selection via optimization}
\paragraph{Stereo matching cost}
The stereo cost $L^{\mathbf{x}}(d)$ evaluates the consistency of the inverse depth $d$ and the pair of input images at location $\mathbf{x}\in \Omega_1$.

Similar to the literature \cite{cheng2019noise}, we compose the stereo cost $L^{\mathbf{x}}(d)$ using the sum of the photometric loss ${L_P}^{\mathbf{x}}(d)$, the census loss ${L_C}^{\mathbf{x}}(d)$, and the image gradient loss ${L_G}^{\mathbf{x}}(d)$ with weights $\alpha_C$ and $\alpha_G$ as follows: 
\begin{equation}
\label{eq:lxd}
    L^{\mathbf{x}}(d) = {L_P}^{\mathbf{x}}(d) + \alpha_C {L_C}^{\mathbf{x}}(d) + \alpha_G {L_G}^{\mathbf{x}}(d).
\end{equation}
Here, ${L_P}^{\mathbf{x}}(d)$, ${L_C}^{\mathbf{x}}(d)$, and ${L_G}^{\mathbf{x}}(d)$ are calculated using the warped coordinates $\mathbf{x}^{\prime}(d)$ in Eq. \eqref{eq:warp_stereo} or \eqref{eq:warp_motion} with the predefined window $W$ as follows:
\begin{align}
    &{L_P}^{\mathbf{x}}(d)= \sum_{{\mathbf{\delta}} \in W} \min{\left( \left| I_1\left(\mathbf{x} + \mathbf{\delta}\right) - I_2\left(\mathbf{x}^{\prime}(d) + \mathbf{\delta} \right) \right|, l_P \right)}
\label{eq:lp}\\
    &{L_C}^{\mathbf{x}}(d) = \min{\left( \left\| C_1\left(\mathbf{x} \right) - C_2\left(\mathbf{x}^{\prime}(d) \right) \right\|_H, l_C \right)}
\label{eq:lc}\\
    &{L_G}^{\mathbf{x}}(d) = \sum_{\mathbf{\delta} \in W} \min{ \left( \left| \nabla I_1\left(\mathbf{x} + \mathbf{\delta} \right) - \nabla I_2\left(\mathbf{x}^{\prime}(d) + \mathbf{\delta}\right) \right| , l_G \right)},
    \label{eq:lg} 
\end{align}
where $C_1$ and $C_2$ respectively represent the census transformation of $I_1$ and $I_2$ with window $W$, $\|\cdot\|_H$ denotes the Hamming distance, and $l_P$, $l_C$, and $l_G$ are the maximum cost values.
We set the window $W$ to be an $11\times11$ square centered at $\mathbf{x}$.
In our evaluations, we set $\alpha_C =\alpha_G=1$ and $l_P=l_C=l_G=0.5$.
\paragraph{Energy definition}
SSM searches the optimal depth value from $S^{\mathbf{x}}$ for every $\mathbf{x} \in \Omega_1$, which is performed using an energy minimization.
Here, the energy follows the conventional stereo disparity estimation \cite{zhang2005parameter}.
We construct an MRF whose nodes are $\mathbf{x} \in \Omega_1$, and the edges $\mathcal{E}$ comprise all the pairs of adjacent pixels.
The energy $E_\mathrm{SSM}$ is defined by the addition of the stereo matching cost $L^{\mathbf{x}}(d)$ defined in Eq. \eqref{eq:lxd} and a smoothness regularization term for the inverse depth as follows:
\begin{align}
\label{eq:essm}
    &E_{\mathrm{SSM}} 
    =\sum_{\mathbf{x} \in \Omega_1}{ L^\mathbf{x}(d) + \lambda_{\mathrm{SSM} } \sum_{e \in \mathcal{E}} {\min{\left(\left|\Delta_{e} d\right|, l_d\right)} }} \mathrm{,} \nonumber
    \\  &\mathrm{\:with\:} d \in S^{\mathbf{x}}.
\end{align}
Here, $\Delta_e$ represents taking the difference across the edge $e$, and $\lambda_{\mathrm{SSM}}$ is the regularization weight.
We empirically set $\lambda_{\rm SSM}=10^2$.
\paragraph{Optimization}
SSM derives a discrete inverse dense depth map ($D_{\mathrm{SSM}}$) by minimizing the energy ($E_\mathrm{SSM}$) in Eq. \eqref{eq:essm}.

The minimization of $E_{\mathrm{SSM}}$ is an optimization of MRF, which we solve by Loopy Belief Propagation (LBP) \cite{yedidia2000generalized}.
In particular, by setting $X \subset \Omega_1$ as the set of four adjacent pixels of $\mathbf{x} \in \Omega_1$, we iteratively update the message from $\mathbf{x}$ to one of its adjacent pixels $\mathbf{y} \in X$ by the min-sum algorithm as shown in Eq. \eqref{eq:update} and \eqref{eq:normalize}.
Here, we denote the iteration index as $n$, the normalized message from $\mathbf{x}$ to $\mathbf{y}$ as $\mathrm{msg}^n_{\mathbf{x} \to \mathbf{y}}(d)$, and the message prior to normalization as $\overline{\mathrm{msg}}^n_{\mathbf{x} \to \mathbf{y}}(d)$.
\begin{align}
\label{eq:update}
\overline{\mathrm{msg}}^n_{\mathbf{x} \to \mathbf{y}}(d) = \min_{d^{\prime} \in S^{\mathbf{x}}}& 
L^{\mathbf{x}}\left(d^{\prime}\right) + \lambda_{\mathrm{SSM}} \min{\left(\left| d - d^{\prime} \right|, l_d \right)} \nonumber \\
&+ \sum_{{\mathbf{z}} \in X \backslash {{\mathbf{y}}}}{\mathrm {msg}^{n-1}_{\mathbf{z}\to \mathbf{x}} }(d^{\prime})
\end{align}
\begin{equation}
\label{eq:normalize}
    \mathrm{msg}^n_{\mathbf{x} \to \mathbf{y}}(d) = \overline{\mathrm {msg}}^n_{\mathbf{x} \to \mathbf{y}}(d) - \log{\sum_{d \in S^{\mathbf{x}}}{\exp{\left(\overline{\mathrm{msg}}_{\mathbf{x} \to \mathbf{y}}\left(d\right)\right)}}}.
\end{equation}
Denoting the message after convergence as $\mathrm{msg}^{\infty}$, the optimal inverse depth value $d_{\mathrm{SSM}}^{\mathbf{x}}$ at $\mathbf{x}$ is expressed as follows:
\begin{equation}
d_{\mathrm{SSM}}^{\mathbf{x}} = \argmin_{d \in S^{\mathbf{x}} } { \left\{L^{\mathbf{x}}(d) + \sum_{\mathbf{y} \in X} {\mathrm{msg}^{\infty}_{\mathbf{y} \to \mathbf{x}}(d)}\right\}}.
\end{equation}

The output inverse depth map $D_{\mathrm{SSM}}$ is assigned based on the optimal values as Eq. \eqref{eq:dssm}.
\begin{equation}
\label{eq:dssm}
D_{\mathrm{SSM}}(\mathbf{x}) = d_{\mathrm{SSM}}^{\mathbf{x}} 
\end{equation}
$D_\mathrm{SSM}$ is visually shown in Fig. \ref{fig:flow}.

In addition, for the ground mask creation in later process (Section \ref{sec:ground_detection}), we construct a map $Y_{\mathrm{SSM}}:\Omega_1\to\Omega_1$ to indicate the original location of the inverse depth map.
Because $D_{\mathrm{SSM}}$ is created by the selection, we know where each value of $D_{\mathrm{SSM}}$ initially located in $D_S$.
In particular, if the value of $D_{\mathrm{SSM}}$ at $\mathbf{x}\in \Omega_1$ is originally at $\mathbf{y}\in \Omega_1$ in $D_S$, we set $Y_{\mathrm{SSM}} = \mathbf{y}$.
By using equations, this assignment is expressed as Eq. \eqref{eq:y_ssm}.
\begin{align}
    \label{eq:y_ssm}
    &Y_\mathrm{SSM}\left(\mathbf{x}\right) = \mathbf{y} \nonumber \\
    &\mathrm{such~that}\: D_S(\mathbf{y})\in S^{\mathbf{x}} \land D_S(\mathbf{y}) = D_{\mathrm{SSM}}(\mathbf{x}).
\end{align}
\subsection{B-ADT aided smoothing}
\label{sec:b-adt}
\begin{table*}[t]
\centering
\caption{Depth completion results on KITTI dataset with the accurate calibration}
\label{tab:eval_kitti}
\begin{tabular}{c|cc|c|cc}
\hline
     Method
     & Input & $^{\ast}$Supervised & Processing time [s]
     & Error rate [\%] & MAE [m] \\
     \hline
      Hernandez et al. \cite{hernandez2016embedded} & Stereo & No & 0.003 & 6.59& 0.977 \\
      Yamaguchi et al. \cite{yamaguchi2014efficient} & Stereo & No & 1.947 & 4.31 & 1.021 \\
     Kopf et al. 
     \cite{kopf2007joint} 
     & Monocular + LiDAR & No & 1.650 & 9.70 & 0.819 \\
     Ferstl et al.
     \cite{ferstl2013image}
     & Monocular + LiDAR & No & 0.064 & 7.45 & 0.578 \\
     Yao et al.
     \cite{yao2020discontinuous} 
     & Monocular + LiDAR & No & 0.065 & 4.47 & 0.413 \\
     Maddern et al. 
     \cite{maddern2016real} 
     & Stereo + LiDAR & No & n/a & 5.91 & n/a \\
     Ours (SSM only) 
     &Stereo + LiDAR & No & 0.968 & 3.43 & 0.399 \\
     Ours 
     &Stereo + LiDAR & No & 0.999 & 3.32 & \textbf{0.356} \\
     \hline
     Park et al. \cite{park2018high} 
     & Stereo + LiDAR & Yes & n/a & 4.84 & n/a \\
     Cheng et al. 
     \cite{cheng2019noise} 
     & Stereo + LiDAR & Yes & 1.721 & \textbf{2.17} & 0.548 \\
     \hline
\multicolumn{5}{c}{$^{\ast}$`Yes" if the method requires accurate LiDAR camera extrinsic calibration parameters during training.}
\end{tabular}
\end{table*}
\setlength{\fboxsep}{0pt}
\begin{figure*}[t]
    \centering
\subfloat[Input and ground truth depth (Scene \#3)]{\includegraphics[width=59mm]{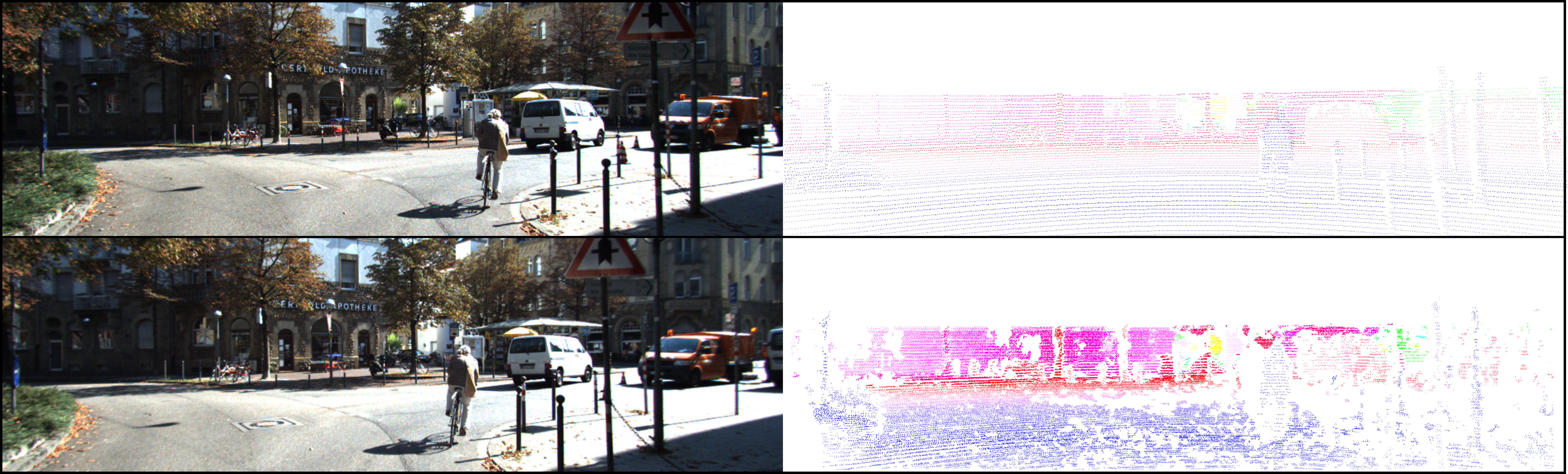}}
\subfloat[Input depth in point cloud (Scene \#3)]{\includegraphics[width=59mm]{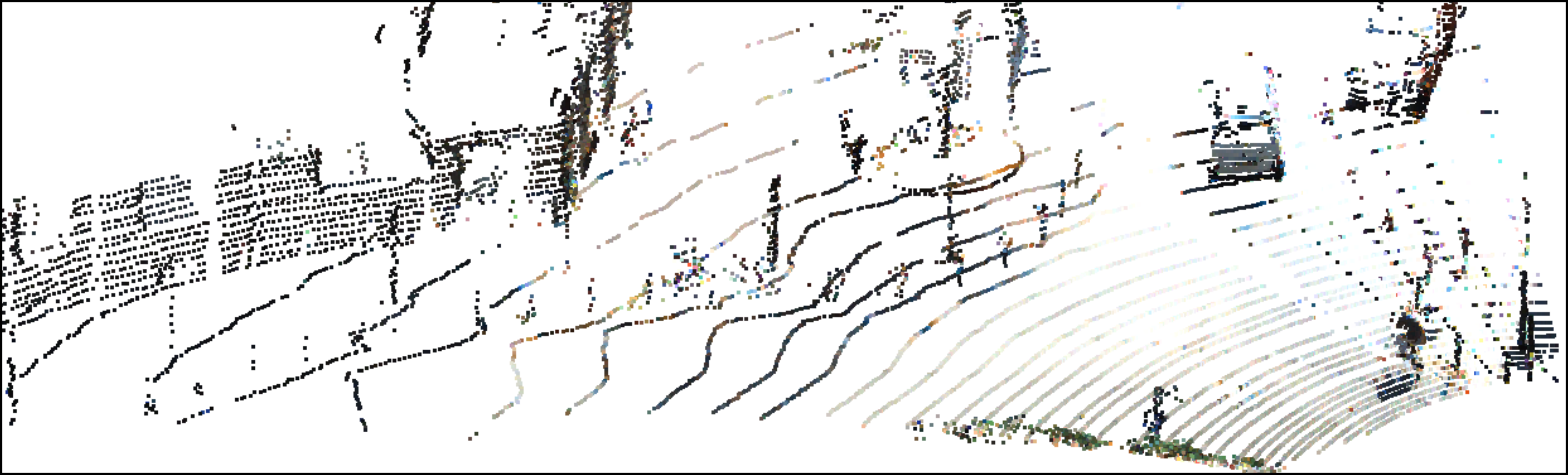}}
\subfloat[Ground truth depth in point cloud (Scene \#3)]{\includegraphics[width=59mm]{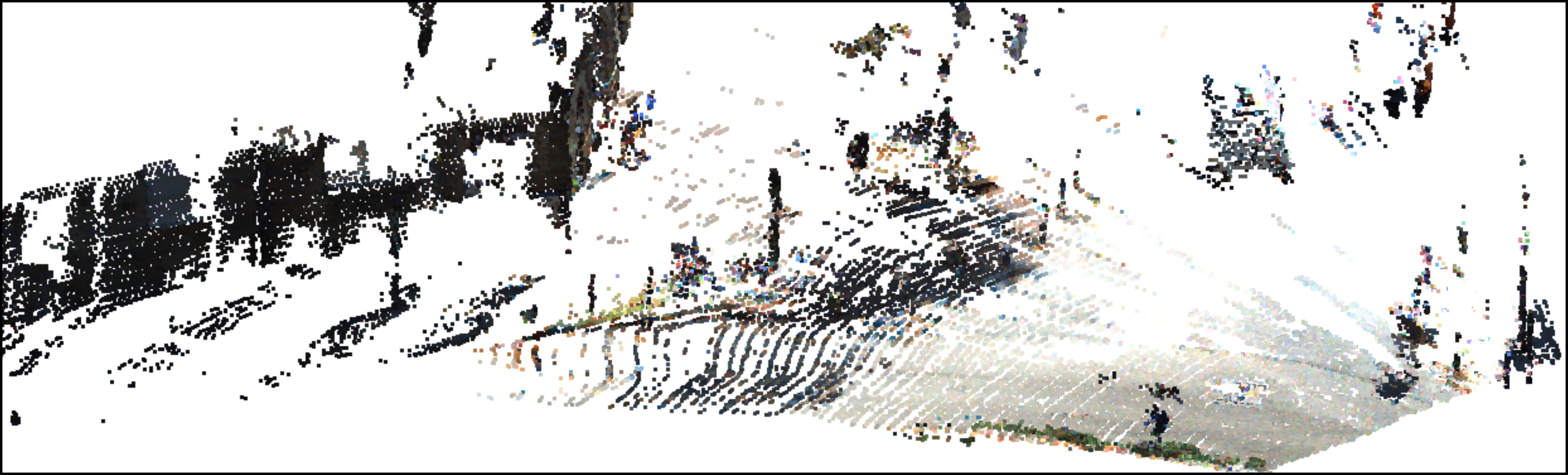}}
\vfil
\vspace{-8pt}
\subfloat[Result of Cheng et al. \cite{cheng2019noise} (Scene \#3)]{\includegraphics[width=59mm]{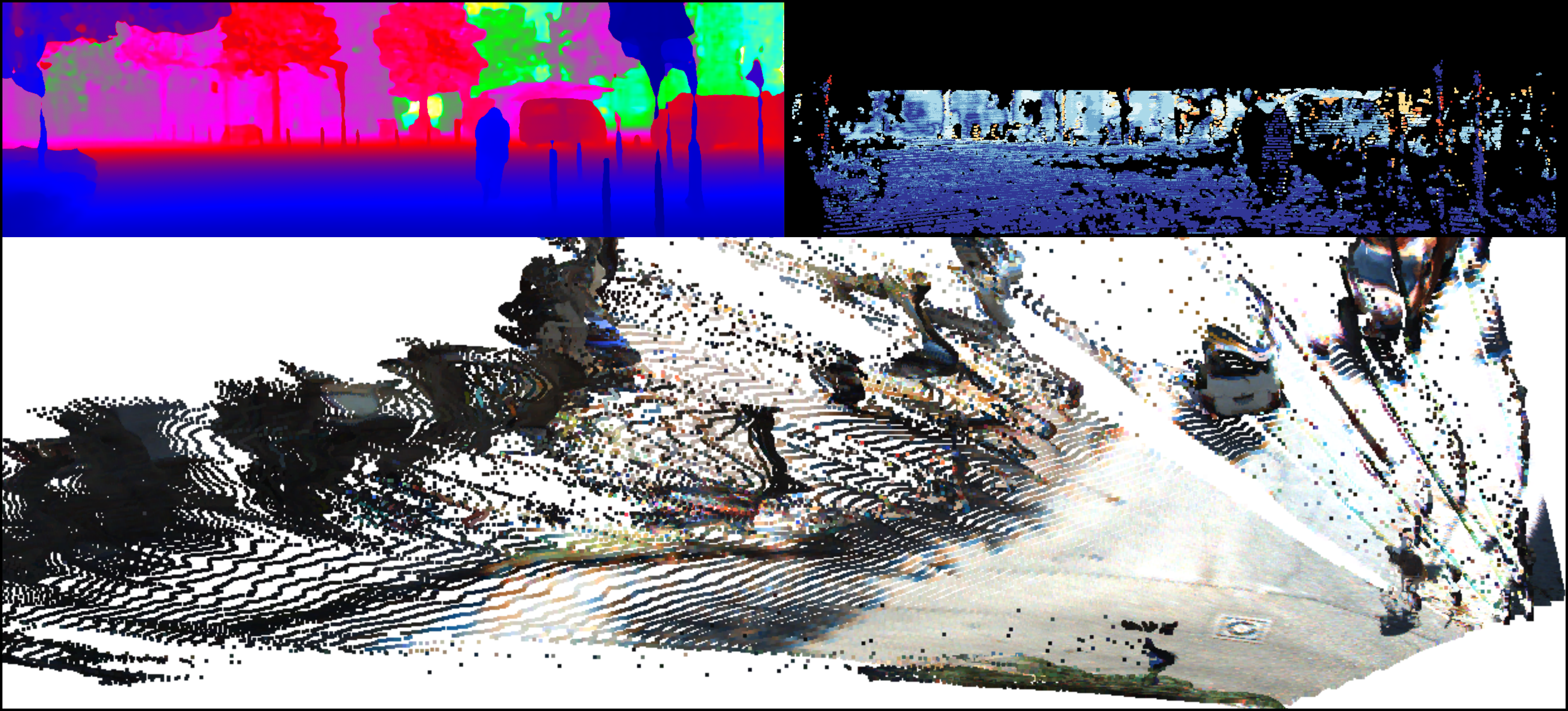}}
\subfloat[Result of Ours (SSM only) (Scene \#3)]{\includegraphics[width=59mm]{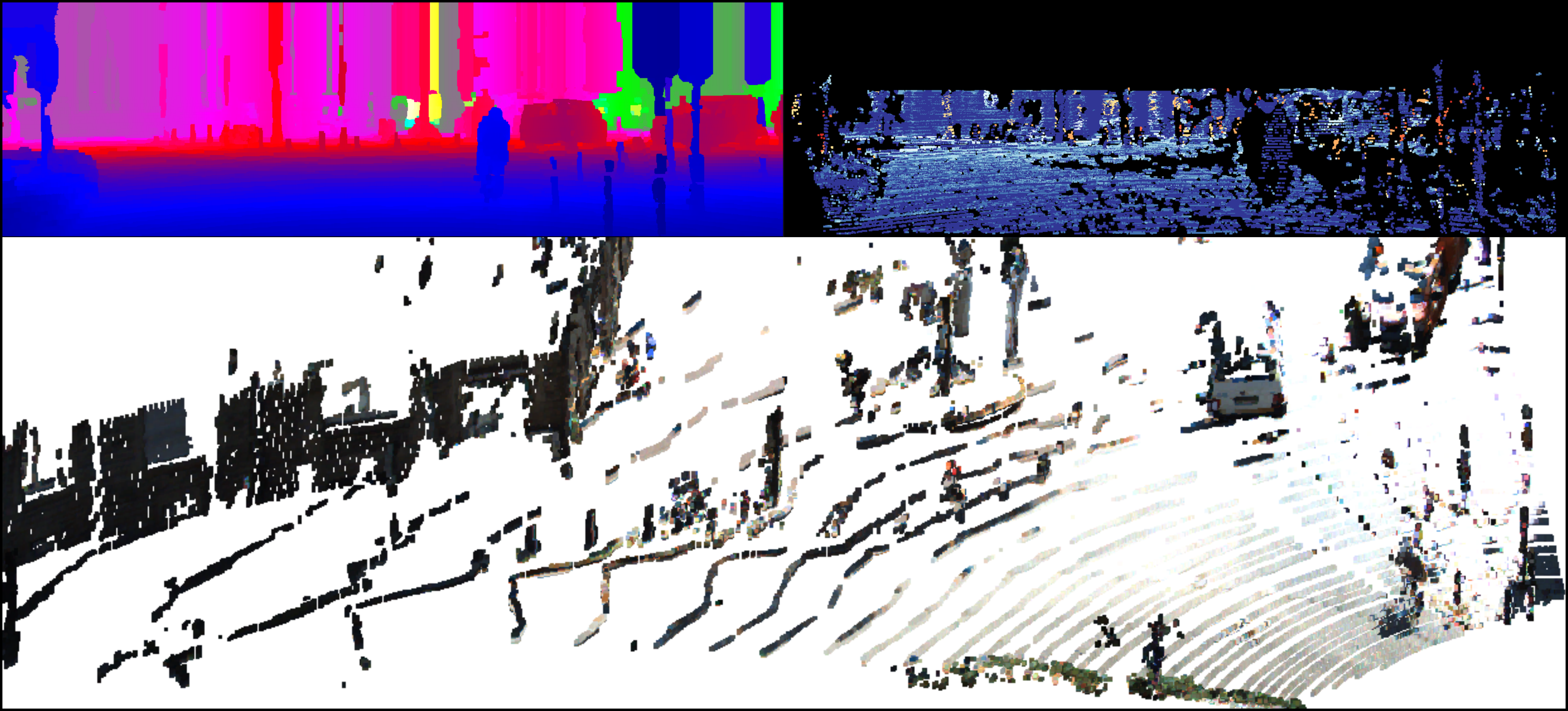}}
\subfloat[Result of Ours (Scene \#3)]{\includegraphics[width=59mm]{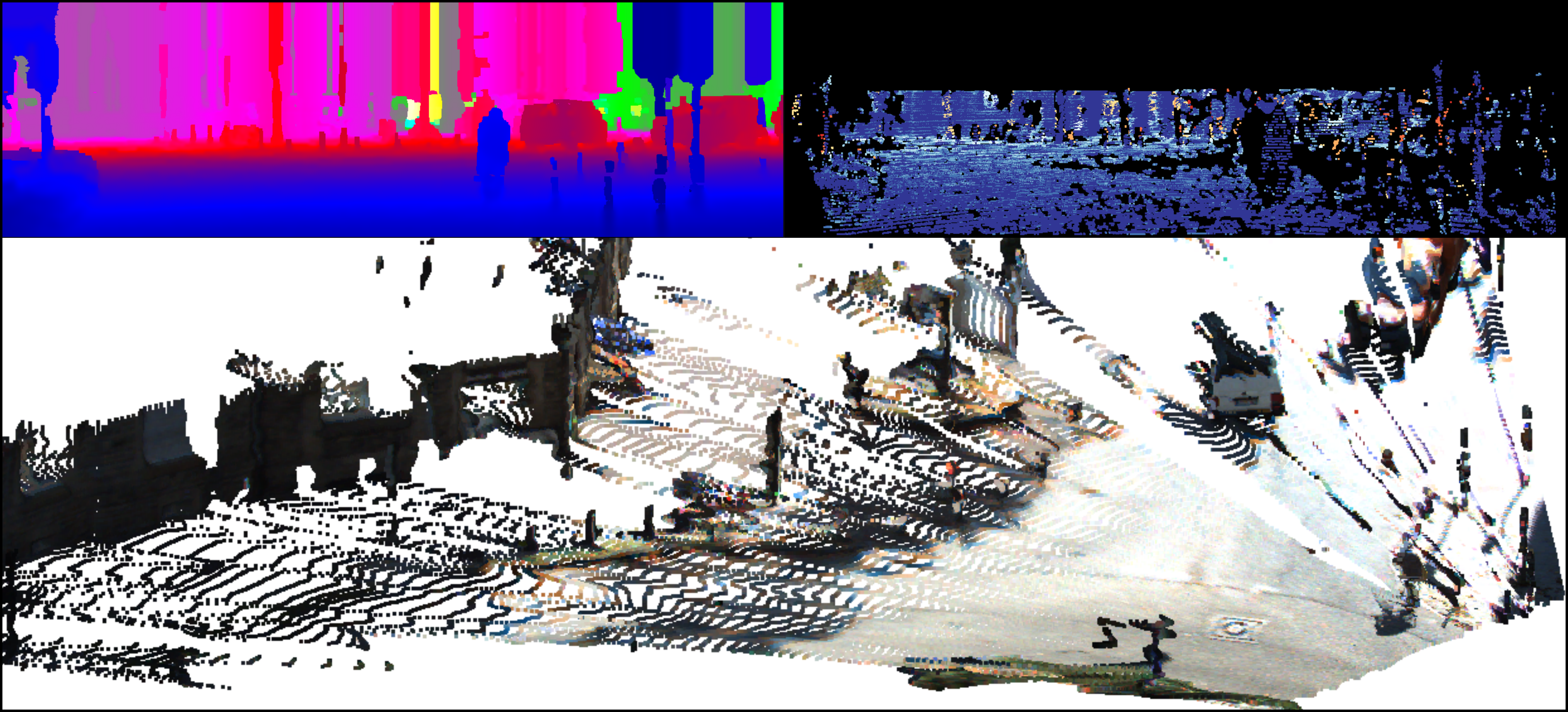}}
\vfil
\subfloat[Input and ground truth depth (Scene \#199)]{\includegraphics[width=59mm]{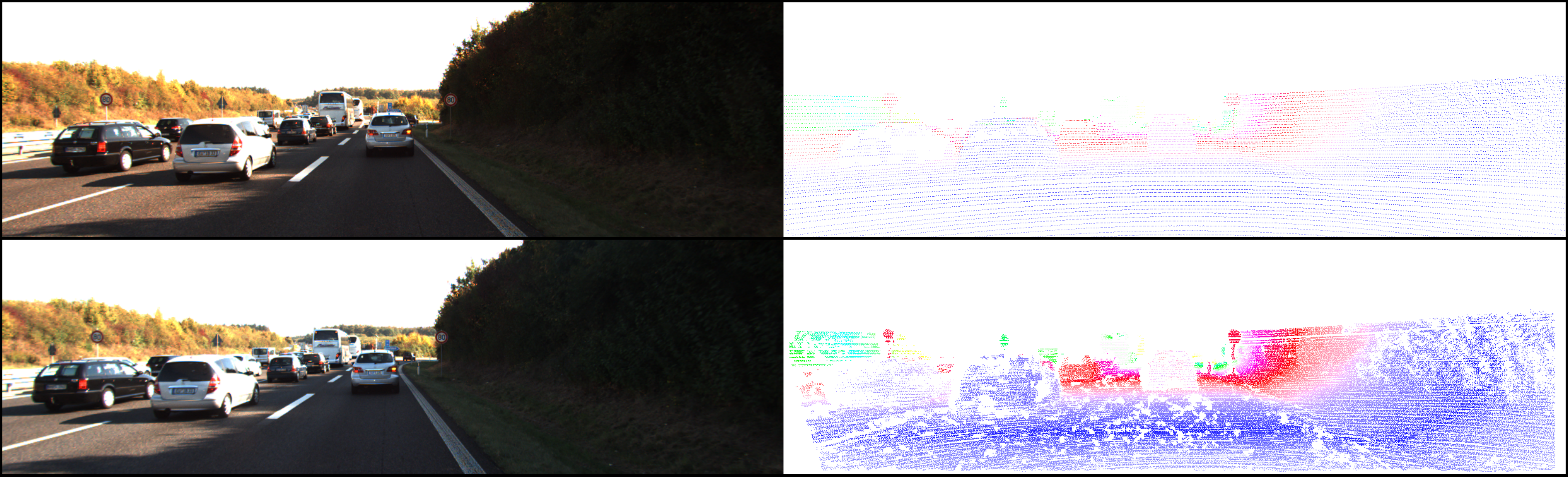}}
\subfloat[Input depth in point cloud (Scene \#199)]{\includegraphics[width=59mm]{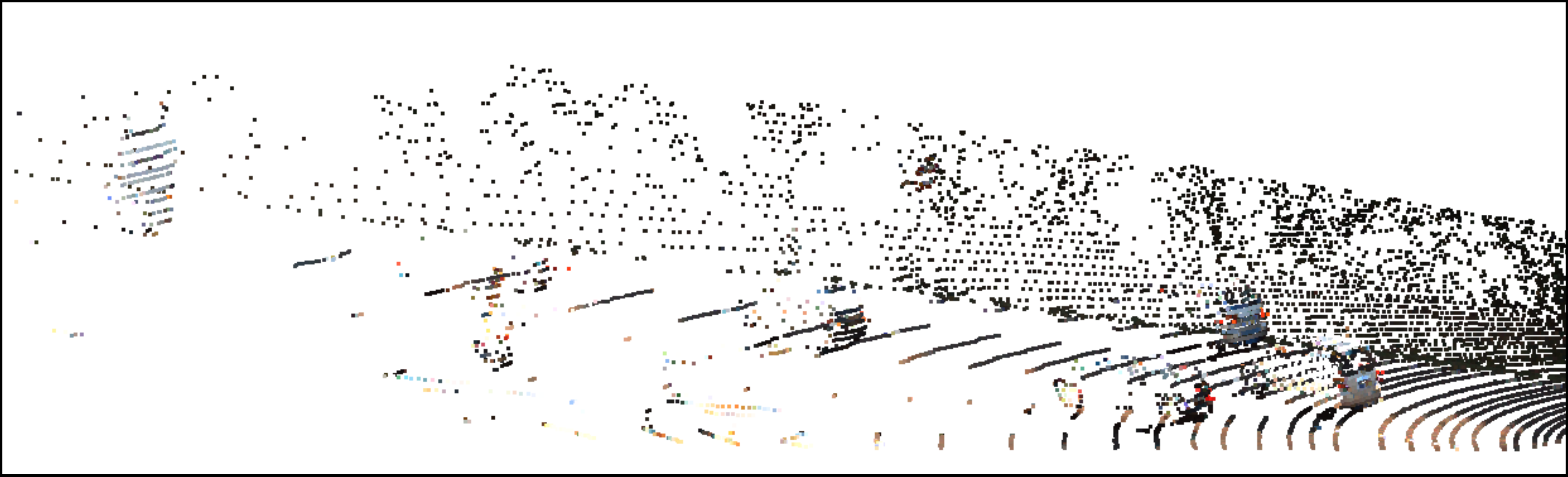}}
\subfloat[Ground truth depth in point cloud (Scene \#199)]{\includegraphics[width=59mm]{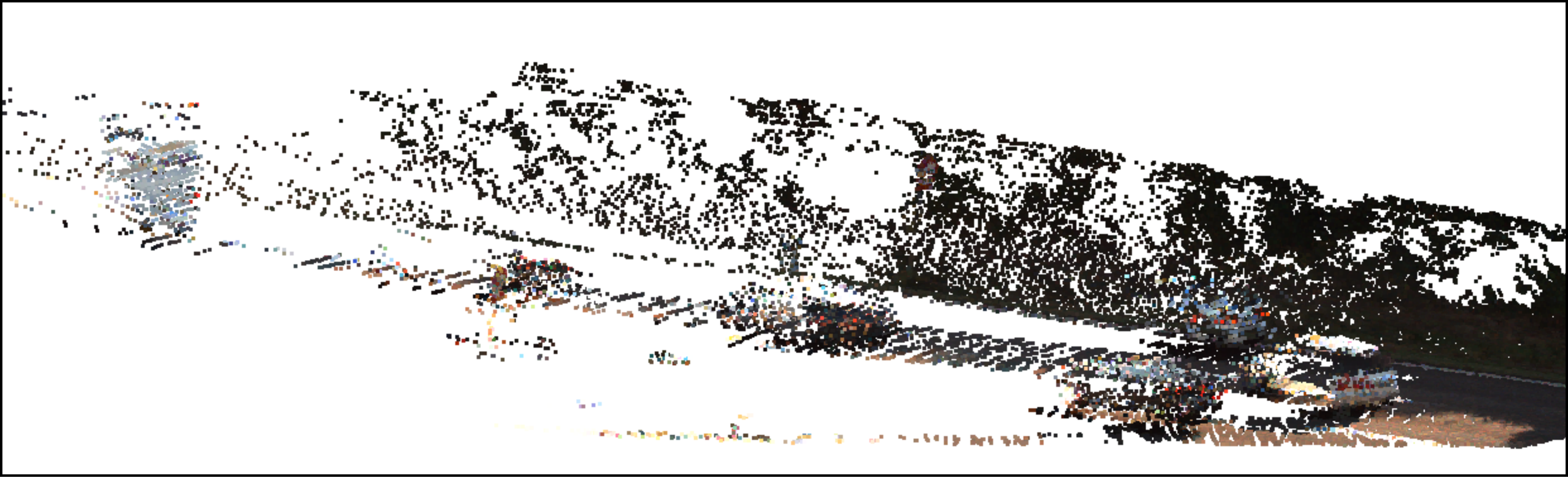}}
\vfil
\vspace{-8pt}
\subfloat[Result of Cheng et al. \cite{cheng2019noise} (Scene \#199)]{\includegraphics[width=59mm]{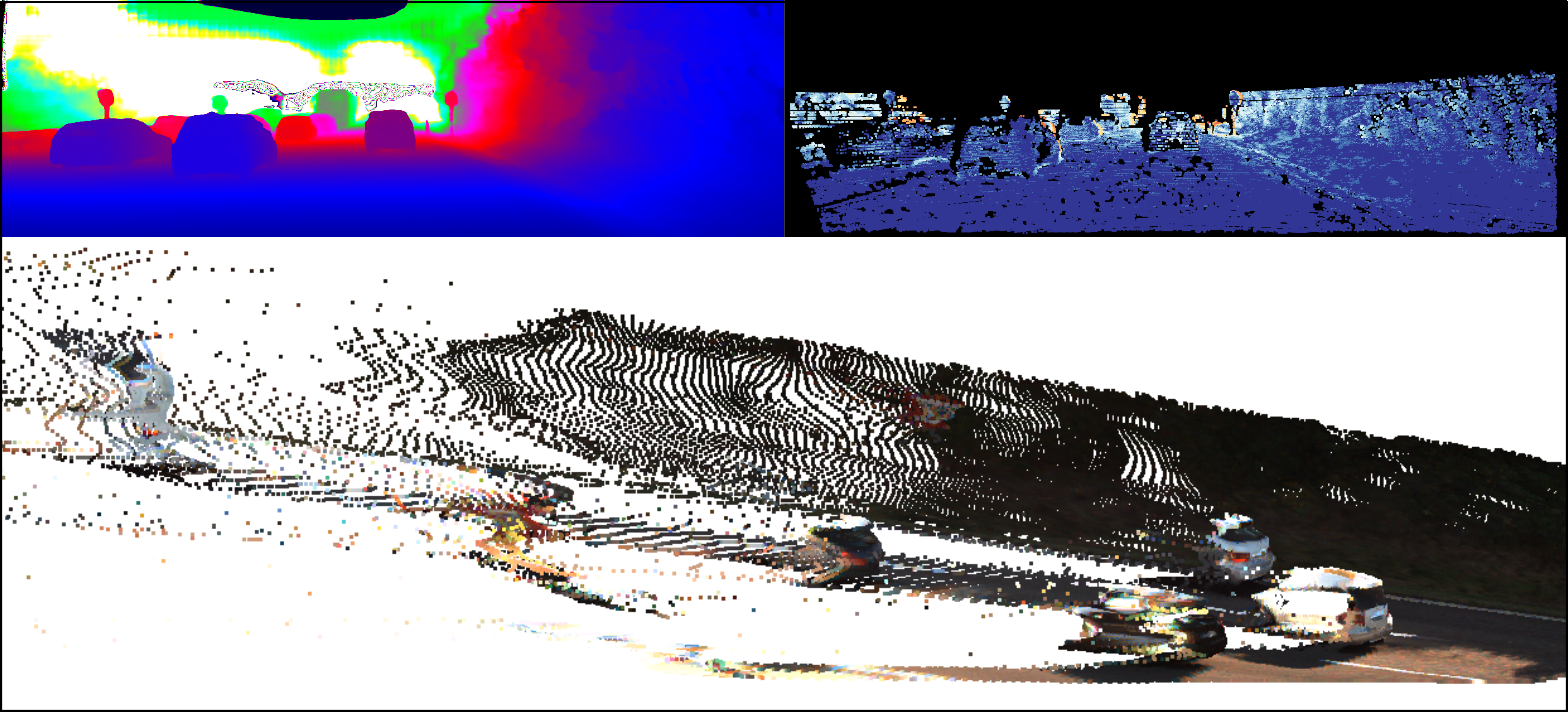}}
\subfloat[Result of Ours (SSM only) (Scene \#199)]{\includegraphics[width=59mm]{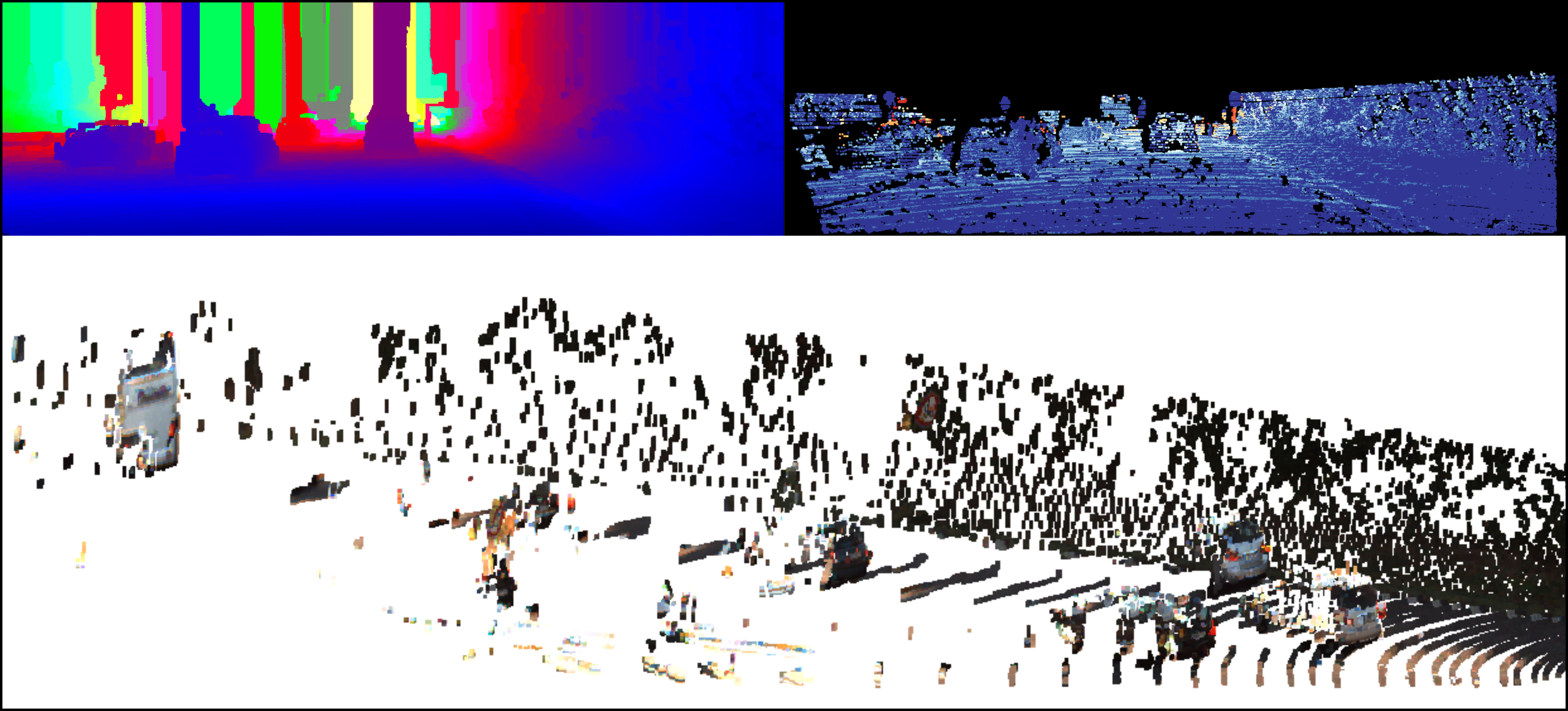}}
\subfloat[Result of Ours (Scene \#199)]{\includegraphics[width=59mm]{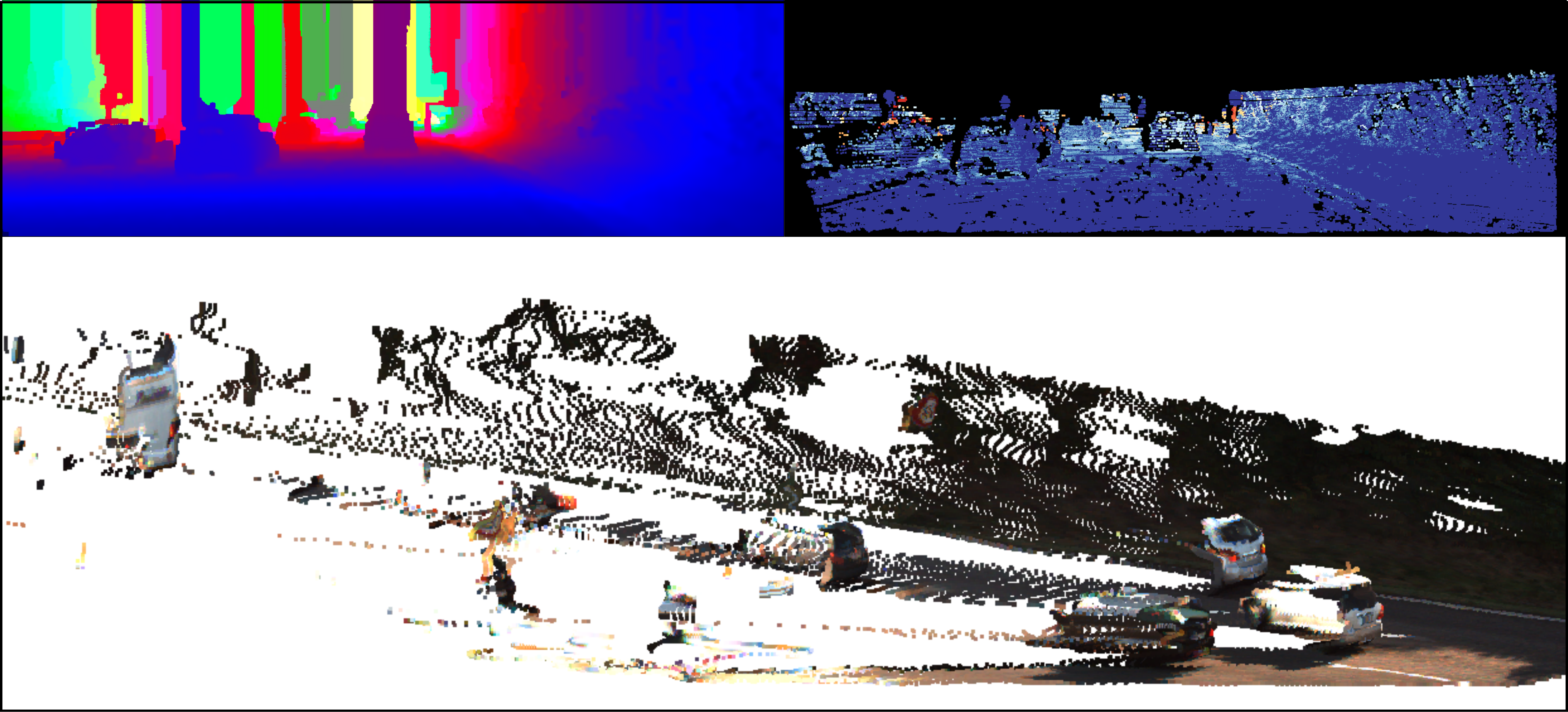}}
\caption{
    Inputs and results of depth completion on KITTI with the accurate calibration.
    The scene numbers are from the KITTI stereo training set \cite{menze2018object}.
    (a), (g) Top left: the input left images, bottom left: input right images, top right: input sparse depth maps, bottom right: ground truth dense depth maps.
    (b), (h) The point clouds created from the input sparse depth maps.
    (c), (i) The point clouds created from the ground truth dense depth maps.
    (d), (e), (f), (j), (k), (l) Top left: the depth completion results, top right: the error maps, bottom: the point clouds generated from the depth completion results.
    As expected, SSM gives discrete point clouds, and our whole framework gives continuous point clouds.
    The surface shapes of the poles and walls in the background are preserved by the proposed method, while they are missing in the results obtained using Cheng's method \cite{cheng2019noise}.
    In addition, the error maps demonstrate smaller errors of our method in the long range compared to Cheng's method \cite{cheng2019noise}.
    }
    \label{fig:result_kitti}
\end{figure*}
$D_{\mathrm{SSM}}$ is discrete because it is generated by the selection from a finite number of candidates.
Here, we apply B-ADT weighted TGV smoothing from \cite{yao2020discontinuous} to derive smooth depth with discontinuity preservation at boundaries.
Below, we explain the ground detection to create the filter for B-ADT derivation, the B-ADT derivation, and the optimization.
Implementation-wise, the ground detection is the RANSAC plane segmentation, B-ADT derivation is a single-step calculation, and the optimization is iterative pixel-wise and parallelized on the GPU.
\subsubsection{Ground detection}
\label{sec:ground_detection}
We create a ground mask to filter out occlusion boundaries that are faultily detected on the ground.

First, we detect the ground points for the input LiDAR depth map.
We convert the LiDAR depth map to the point cloud and apply the RANSAC plane segmentation \cite{fischler1981random}.
The RANSAC plane segmentation iteratively searches the coefficients of a plane having the maximum number of inlier points within the given threshold $d_{\mathrm{rand}}$, by randomly sampling three points from the point cloud to derive the plane coefficients in every iteration.
For RANSAC parameters, we set $d_{\mathrm{rand}}=0.2$ [m] and the number of iterations as 100.

Then, we project the inlier points of the derived plane to the image domain $\Omega_1$ and acquire the ground mask $\Gamma_S:\Omega_1 \to \left\{0,1\right\}$, where $\Gamma_S \left( \mathbf{x} \right)=1$ if $\mathbf{x}$ is the ground.

Finally, we create a dense ground mask $\Gamma_{\mathrm{SSM}}:\Omega_1 \to \left\{0,1\right\}$ which is aligned with $D_{\mathrm{SSM}}$.
$\Gamma_{\mathrm{SSM}}$ is derived by interpolating $\Gamma_S$ based on the selection performed in SSM.
In particular, using $Y_{\mathrm{SSM}}$ in Eq. \eqref{eq:y_ssm}, $\Gamma_{\mathrm{SSM}}$ is assigned as follows:
\begin{equation}
\label{eq:gamma_SSM}
    \Gamma_{\mathrm{SSM}}\left(\mathbf{x}\right) = \Gamma_S\left(Y_{\mathrm{SSM}}\left(\mathbf{x}\right)\right)
\end{equation}
In Fig \ref{fig:flow}, $\Gamma_S$ and $\Gamma_{\mathrm{SSM}}$ are visualized by indicating the ground pixels in black.
\subsubsection{B-ADT derivation}
B-ADT is pixel-wise weighting for the variational regularization term.
Here, B-ADT is derived based on the occlusion boundary conditions in $D_{\mathrm SSM}$ and the ground mask $\Gamma_{\mathrm{SSM}}$.
Occlusion boundaries are boundaries where objects are not in contact; thus, the depth values at the occlusion boundaries immediately change.

The B-ADT for each pixel is assigned based on the following two conditions: $A$, i.e., the pixel is on a vertical occlusion boundary, and $B$, i.e., the pixel is on a horizontal occlusion boundary.
Here, a vertical occlusion boundary is a vertical line segment across which the depth is horizontally discontinuous, and a horizontal occlusion boundary is a horizontal line segment across which the depth is vertically discontinuous.

In particular, with predefined threshold $t$, we determine a pixel $\mathbf{x}\in \Omega_1$ is in $A$ if $\left|\partial_x D_{\mathrm{SSM}}^{-1}(\mathbf{x})\right| > t$, and in $B$ if $\left|\partial_y D_{\mathrm{SSM}}^{-1}(\mathbf{x})\right| > t$.
To make occlusion boundaries where the adjacent depths change more than 2 [m], we used $t=2$ in our evaluations.
Because images are defined on 2D grids, every pixel belongs to one of four sets, i.e., neither $A$ nor $B$ ($\bar{A} \cap \bar{B}$), $A$ but not $B$ ($A \cap \bar{B}$), not $A$ but $B$ ($\bar{A} \cap B$), and $A$ and $B$ ($A \cap B$).

Boundary detection by a single threshold can be faulty, particularly in the ground region because the ground is often a large plane parallel to the view direction with a wide depth range.
Thus, we filter out occlusion boundaries that are detected on the ground by $\Gamma_{\mathrm{SSM}}$ in Eq. \eqref{eq:gamma_SSM}.

Finally, by denoting B-ADT at pixel $\mathbf{x} \in \Omega_1$ as $G^{\mathbf{x}}$, we set $G^{\mathbf{x}}$ based on the boundary conditions and the ground mask as follows:.
\begin{align}
\label{eq:b-adt}
G^{\mathbf{x}} = \left\{ \begin{array}{ll}
      \left(
      \begin{array}{cc}
           1 & 0 \\
           0 & 1 
      \end{array}\right) &\mathrm{if}\ \mathbf{x} \in \bar{A} \cap \bar{B}\:\mathrm{or}\: \Gamma_{\mathrm{SSM}}(\mathbf{x}) = 1 \\
      \left(
      \begin{array}{cc}
           0 & 0 \\
           0 & 1 
      \end{array}\right) &\mathrm{if}\ \mathbf{x} \in A \cap \bar{B}\:\mathrm{and}\: \Gamma_{\mathrm{SSM}}(\mathbf{x}) = 0 \\
      \left(
      \begin{array}{cc}
           1 & 0 \\
           0 & 0 
      \end{array}\right) &\mathrm{if}\ \mathbf{x} \in \bar{A} \cap B\:\mathrm{and}\: \Gamma_{\mathrm{SSM}}(\mathbf{x}) = 0 \\
      \left(
      \begin{array}{cc}
           0 & 0 \\
           0 & 0 
      \end{array}\right) &\mathrm{if}\ \mathbf{x} \in A \cap B\:\mathrm{and}\: \Gamma_{\mathrm{SSM}}(\mathbf{x}) = 0\\
\end{array}\right.
\end{align}
\subsubsection{Optimization}
We minimize the energy with B-ADT weighted TGV regularization to acquire the output of the proposed framework.
By denoting the inverse depth map during optimization as $u : \Omega_1 \to \mathbb{R}$ and the relaxation variable as $\mathbf{v} : \Omega_1 \to \mathbb{R}^2$, we define the energy $E_{\mathrm{TGV}}$ as the sum of the data term $C\left[u\right]$ and the smoothness term $R\left[u, \mathbf{v}\right]$ as follows:
\begin{align}
&E_{\mathrm{TGV}}
=  C\left[u\right] + R\left[u, \mathbf{v}\right]\\
&C\left[u\right] = \int_{\Omega_1} w\left| u - D_{\mathrm{SSM}} \right|^2 d{\mathbf{x}} \\
\label{eq_our_tgv_term}
&R\left[u, \mathbf{v}\right] = \int_{\Omega_1} \lambda_A \left|G^{\mathbf{x}} \left(\nabla u - \mathbf{v} \right)\right| + \lambda_B \left|\nabla \mathbf{v}\right| d \mathbf{x},
\end{align}
where $w$ is the pixel-wise weight for the data term, and $\lambda_A$ and $\lambda_B$ are weights for the energy terms.
We set $w=D_\mathrm{SSM}^{-2.5}$, $\lambda_A=1.0$, and $\lambda_B=8.0$ based on the literature \cite{yao2020discontinuous}.

Note that $E_{\mathrm{TGV}}$ is convex, and we can derive the optimums of $u$ and $\mathbf{v}$ using the primal dual algorithm \cite{chambolle2011first}.
In the end, the output of the proposed framework (a dense inverse depth map $D$) is the optimum of $u$ as Eq. \eqref{eq:tgv_min}.
\begin{equation}
    \label{eq:tgv_min}
    D = \argmin_{u} E_{\mathrm{TGV}}.
\end{equation}
\subsection{Parameter setting for SSM}
\label{sec:param}
SSM introduces a parameter $r$ as the radius for candidate set search.
Here, we present a practical approach to select the $r$ value. 
$r$ should be as small as possible to cover the nearest appropriate depth because the number of candidates increases as $r$ increases, which generally leads to inappropriate selections. 
Furthermore, we consider two primary causes for mis-projection, i.e., LiDAR-camera calibration error and occlusion.

The mis-projection caused by calibration errors is primarily attributed to rotational errors.
At the center of the image, projection error $\sigma_{\mathrm{calib}}$ caused by rotation error $\theta_{\mathrm{calib}}$ can be calculated as follows:
\begin{equation}
\label{eq:sigma}
    \sigma_{\mathrm{calib}} = f \tan{\theta_{\mathrm{calib}}} \; \mathrm{[pixel]},
\end{equation}
where $f$ is the camera focal length. 
Although the exact value of $\theta_{\mathrm{calib}}$ cannot be known, it can be practically given in several ways, e.g., the error range presented in the reference of the original calibration method or by visually observing the LiDAR points projected onto the image.

To handle mis-projection by occlusion, all pixels typically should have several candidates in the range. 
Empirically, we found that this can be achieved when the radius is set to cover two scanlines. 
The pixel distance between two scanlines $\sigma_{\mathrm{scan}}$ is estimated using Eq. \eqref{eq:sigma} with angle $\theta_{\mathrm{scan}}$ between the scanlines as follows:
\begin{equation}
    \sigma_{\mathrm{scan}} = f \tan{\theta_{\mathrm{scan}}} \mathrm{[pixel]}.
\end{equation}

We set the optimal radius $r^{\ast}$ to be maximum of $\sigma_{\mathrm{calib}}$ and $\sigma_{\mathrm{scan}}$ to cover mis-projection caused by calibration errors and occlusion as follows:
\begin{equation}
\label{eq:max_r}
r^{\ast} = \max\left(\sigma_{\mathrm{calib}}, \sigma_{\mathrm{scan}} \right) \mathrm{[pixel]}.
\end{equation}
\section{Evaluation}
\label{sec:eval}
\begin{figure*}
    \centering
    \subfloat{
    \includegraphics[width=17.8mm]{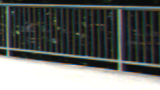}}
    \subfloat{
    \includegraphics[width=17.8mm]{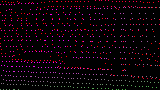}}
    \subfloat{
    \includegraphics[width=17.8mm]{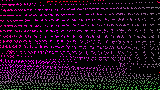}}
    \subfloat{
    \includegraphics[width=17.8mm]{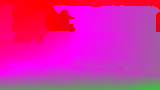}
    \includegraphics[width=17.8mm]{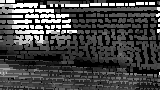}}
    \subfloat{
    \includegraphics[width=17.8mm]{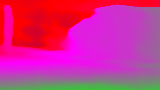}
    \includegraphics[width=17.8mm]{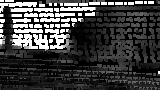}} 
    \subfloat{
    \includegraphics[width=17.8mm]{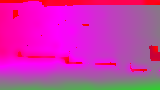}
    \includegraphics[width=17.8mm]{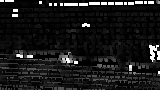}}
    \vfil
    \vspace{-8pt}
    \subfloat{
    \includegraphics[width=17.8mm]{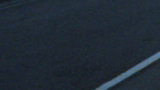}}
    \subfloat{
    \includegraphics[width=17.8mm]{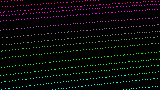}}
    \subfloat{
    \includegraphics[width=17.8mm]{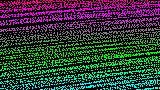}}
    \subfloat{
    \includegraphics[width=17.8mm]{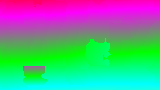}
    \includegraphics[width=17.8mm]{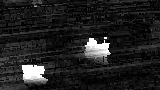}}
    \subfloat{
    \includegraphics[width=17.8mm]{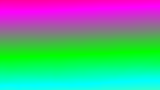}
    \includegraphics[width=17.8mm]{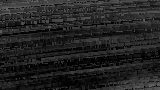}} 
    \subfloat{
    \includegraphics[width=17.8mm]{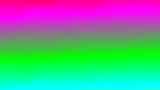}
    \includegraphics[width=17.8mm]{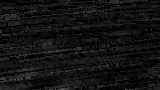}}
    \vfil
    \vspace{-8pt}
    \subfloat{
    \includegraphics[width=17.8mm]{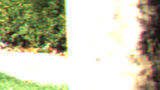}}
    \subfloat{
    \includegraphics[width=17.8mm]{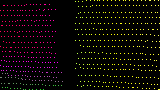}}
    \subfloat{
    \includegraphics[width=17.8mm]{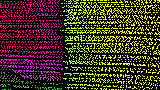}}
    \subfloat{
    \includegraphics[width=17.8mm]{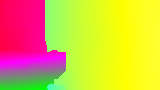}
    \includegraphics[width=17.8mm]{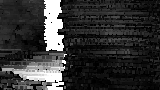}}
    \subfloat{
    \includegraphics[width=17.8mm]{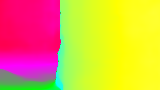}
    \includegraphics[width=17.8mm]{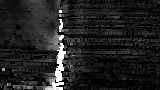}} 
    \subfloat{
    \includegraphics[width=17.8mm]{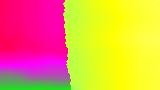}
    \includegraphics[width=17.8mm]{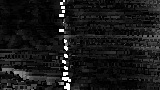}}
    \vfil
    \vspace{-8pt}
    \setcounter{subfigure}{0}
    \subfloat[Image]{
    \includegraphics[width=17.8mm]{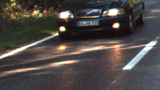}}
    \subfloat[LiDAR]{
    \includegraphics[width=17.8mm]{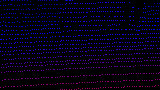}}
    \subfloat[G.t.]{
    \includegraphics[width=17.8mm]{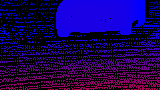}}
    \subfloat[Yamaguchi et al. \cite{yamaguchi2014efficient}]{
    \includegraphics[width=17.8mm]{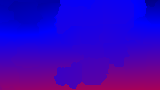}
    \includegraphics[width=17.8mm]{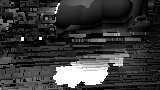}}
    \subfloat[Cheng et al. \cite{cheng2019noise}]{
    \includegraphics[width=17.8mm]{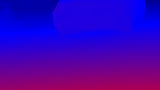}
    \includegraphics[width=17.8mm]{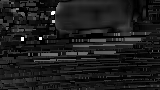}}
    \subfloat[Ours]{
    \includegraphics[width=17.8mm]{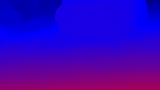}
    \includegraphics[width=17.8mm]{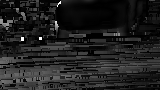}}
    \caption{
    Disparity estimation in challenging conditions: from top to bottom, repetitive pattern, low texture, discontinuity, and specular reflection.
    (a) Input left images.
    (b) Input LiDAR disparity maps.
    (c) The ground truth dense disparity maps.
    (d), (e), (f) left: disparity estimations, right: error maps with white indicating large error.
    The stereo-only method \cite{yamaguchi2014efficient} has larger disparity error than LiDAR-aided methods.
    Moreover, our method has less error in the repetitive pattern and discontinuity conditions than the state-of-the-art stereo-aided depth completion\cite{cheng2019noise}.
    }
    \label{fig:result_disparity}
\end{figure*}
\begin{figure}[t]
    \centering
    \includegraphics[width=3.3in]{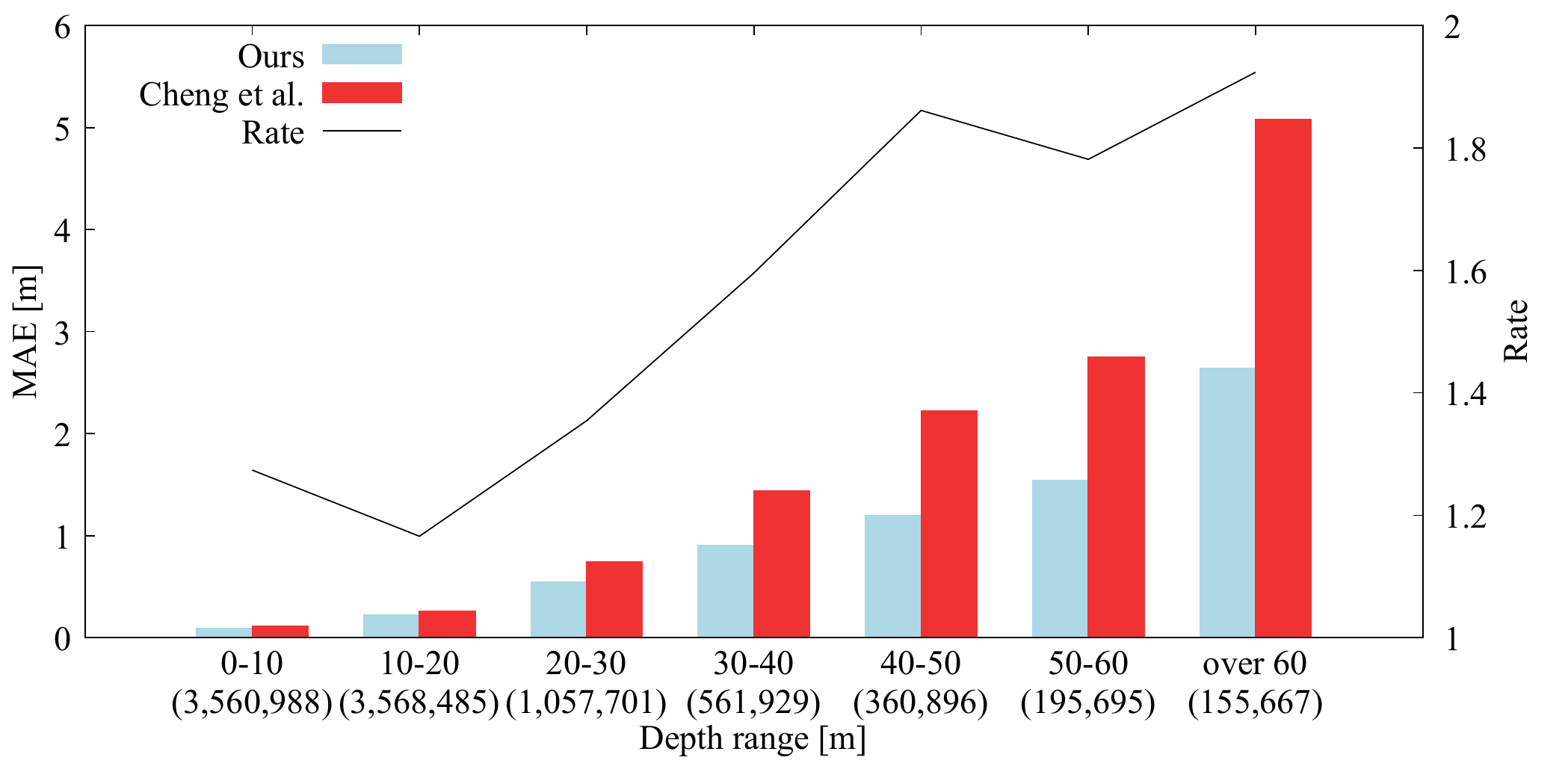}
    \caption{Plot of MAEs against depth ranges on KITTI with the accurate calibration.
    The numbers in parentheses indicate the number of occurrences.
    The proposed method obtained smaller MAE than Cheng's method \cite{cheng2019noise} in all the ranges.
    Moreover, the proposed method approximately achieved half MAE in longer ranges than 40 [m].}
    \label{fig:depth_wise}
\end{figure}
We performed an evaluation that used the accurate LiDAR-camera extrinsic calibration (Section \ref{sec:eval_kitti}), another that used erroneous LiDAR-camera extrinsic calibration (Section \ref{sec:eval_kitti_error}), and the other for the parameter study (Section \ref{sec:study}).
In the first evaluation, we compared the accuracy of the proposed method to that of existing state-of-the-art methods under common experimental conditions \cite{maddern2016real, park2018high, cheng2019noise}.
Moreover, we analyzed the accuracy distribution over the depth range to demonstrate the advantage of the proposed method in the long range.
In the second evaluation, we examined the robustness of the proposed method against LiDAR-camera extrinsic calibration errors.
In this experiment, we used the KITTI \cite{geiger2012automatic} and Komaba datasets \cite{hirata2019real} with added calibration errors.

In all evaluations, we implemented SSM and B-ADT aided smoothing on GPU by CUDA, and used RANSAC plane segmentation from PCL library \cite{radu20113d} for the ground detection.

\subsection{Evaluation with accurate calibration}
\label{sec:eval_kitti}
We evaluated the proposed method on a subset of the KITTI dataset, which is commonly used to evaluate stereo-LiDAR fusion \cite{cheng2019noise,maddern2016real, park2018high}.
These data comprise 141 sets of left and right images, sparse LiDAR depth maps, dense disparity maps, and dense depth maps.
The figure of an example frame of the KITTI dataset is in the supplementary material.
Here, we used the ground truths of the dense disparity map \cite{menze2018object} and dense depth map \cite{uhrig2017sparsity} for the evaluation.
Note that the input sparse depth maps still have mis-projection caused by occlusions, although the extrinsic calibration is accurate, as shown in Fig. \ref{fig:kitti_error} (a).
In this evaluation, we set the radius for the candidate search to $r=5$ [pixel] for our method.

We compared the proposed method to non-learning stereo methods \cite{hernandez2016embedded, yamaguchi2014efficient}, non-learning single-image-aided depth completion methods \cite{ferstl2013image, yao2020discontinuous}, non-learning stereo-aided depth completion methods \cite{maddern2016real}, and supervised stereo-aided depth completion methods \cite{park2018high, cheng2019noise}.
Note that we assume Cheng's method \cite{cheng2019noise} is a supervised because it uses an accurately calibrated dataset during training. 

Implementation conditions are as follows. 
We used our own CUDA implementations for several methods \cite{kopf2007joint, ferstl2013image, yao2020discontinuous}.
We used the authors' implementation for the non-learning stereo methods \cite{hernandez2016embedded, yamaguchi2014efficient}.
We used the authors' implementation and their trained model for Cheng's method \cite{cheng2019noise}.
We referred to the results in the original papers with the same experimental conditions for methods \cite{maddern2016real, park2018high}.%
%
\subsubsection{Overall accuracy}
Table \ref{tab:eval_kitti} compares the accuracy of each method, and Fig. \ref{fig:result_kitti} shows the visualized results.
This evaluation was based on the error rate measured with the dense disparity maps and the Mean Absolute Error (MAE) measured with the dense depth maps. 
Here, the error rate is defined as per the literature \cite{menze2018object} and is the percentage of stereo disparity outliers that have errors greater than or equal to three pixels.
The proposed method outperformed the compared methods in terms of MAE.
Moreover, although the proposed method is a non-learning method, it demonstrated a competitive error rate compared to the supervised stereo-aided depth completion \cite{park2018high, cheng2019noise}.

In addition, Table \ref{tab:eval_kitti} demonstrates the general advantage of LiDAR-aided methods, including our method, in relation to stereo-only methods \cite{hernandez2016embedded, yamaguchi2014efficient} in terms of depth accuracy.
Furthermore, Fig. \ref{fig:result_disparity} shows the advantage of LiDAR-aided methods in challenging conditions as a repetitive pattern, low texture, discontinuity, and specular reflection.
\subsubsection{Long-range accuracy}
Figure \ref{fig:depth_wise} shows a breakdown of MAE against the depth range compared to Cheng's method \cite{cheng2019noise}.
As shown, the difference in MAE between the proposed method and Cheng's method \cite{cheng2019noise} increased as the distance increased. 
Cheng's method estimate depth by pixel disparity, and its depth precision decreases as the distance increases. 
In contrast, the proposed method is based on the selection of LiDAR depths and does not lose the depth precision in the long range. 
The accuracy in the long range is also visible by the point clouds in Fig. \ref{fig:result_kitti}.
The method of Cheng et al. \cite{cheng2019noise} lost the shapes of background objects, e.g., the poles, walls, and cars, whereas the shapes of these objects were retained in the results of the proposed method.
\subsubsection{Processing time}
Table \ref{tab:eval_kitti} also shows the processing time in our environment, which is a laptop computer running Intel Core i9 and GeForce RTX 2080.
Although the processing time of our method is not in real time as methods \cite{hernandez2016embedded, ferstl2013image, yao2020discontinuous}, our method performed faster than the state-of-the-art non-learning stereo matching \cite{yamaguchi2014efficient} and stereo-aided depth completion \cite{cheng2019noise}.
\subsection{Evaluations with calibration errors}
\label{sec:eval_kitti_error}
\begin{figure}[t]
\centering
\subfloat[Calibrated]{\includegraphics[width=41mm]{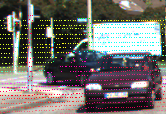}
}
\hfil
\subfloat[\textit{buleprint}]{\includegraphics[width=41mm]{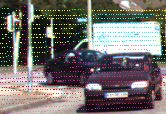}
}
\vfil
\vspace{-8pt}
\subfloat[\textit{error-1}]{\includegraphics[width=41mm]{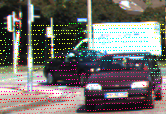}
}
\hfil
\subfloat[\textit{error-2}]{\includegraphics[width=41mm]{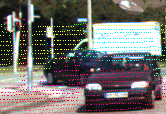}
}
\caption{
Mis-projection in the KITTI dataset.
(a) Mis-projection is caused by occlusion although the calibration is accurate.
(b), (c), (d) Calibration errors cause mis-projection in addition to the occlusion (see the poles).
}
\label{fig:kitti_error}
\end{figure}
\begin{table*}[t]
    \centering
    \caption{Input calibration error details}
    \begin{tabular}{c|c|cc|cc}
    \hline
    Dataset & error type& Rot. axis (x,y,z) & Rot. error [deg.] & Trans. direction (x,y,z) & Trans. error [m] \\
    \hline
        & \textit{blueprint} & (0.04,-0.89,0.45) & 0.952 & (0.03,-0.05,-0.99) & 0.076 \\
        KITTI
        & \textit{error-1} (avg.)$^{\ast}$ 
        & (-0.79,0.44,0.43) & 0.675 & (0.51,0.54,0.67) & 0.155 \\
        &\textit{error-2} (avg.)$^{\ast}$ 
        & (-0.94,0.33,-0.03) & 0.667 & (0.36,0.83,0.42)& 0.207 \\
    \hline
    Komaba & - & (0.51,-0.11,0.85) & 1.096 & (-0.26, -0.73, 0.63) & 0.207 \\ 
    \hline
    \multicolumn{6}{l}{$^{\ast}$ Averages are shown since every frame has different errors.}
    \end{tabular}
    \label{tab:data_errors}
\end{table*}
\begin{table*}[t]
\centering
\caption{Depth completion results obtained on KITTI dataset with calibration errors}
     \begin{tabular}{c|cc|cc|cc}
     \hline
     & \multicolumn{2}{c|}{\it{blueprint} }& \multicolumn{2}{c|}{\it{error-1} }& \multicolumn{2}{c}{\it{error-2}}\\
     \cline{2-7}
     Method & Error rate [\%] & MAE [m] & Error rate [\%] & MAE [m] & Error rate [\%] & MAE [m] \\ 
     \hline
     Kopf et al. \cite{kopf2007joint} & 9.93 & 1.584 & 15.65 & 1.094 & 14.97 & 1.083 \\
     Ferstl et al. \cite{ferstl2013image} & 9.72 & 1.600 & 15.17 & 1.025 & 14.34 & 1.022 \\
     Yao et al. \cite{yao2020discontinuous} & 7.66 & 1.576 & 12.91 & 1.023 & 12.08 & 0.986 \\
     Ours(SSM only) & 4.53 & 0.577 & 4.90 & 0.577 & 5.00 & 0.584 \\
     Ours & \textbf{4.11} & \textbf{0.528} & \textbf{4.43} & \textbf{0.527} & \textbf{4.50} & \textbf{0.532} \\
\hline
\end{tabular}
\label{tab:result_kitti_error}
\end{table*}
\begin{figure*}
    \centering
    \subfloat{\fbox{\includegraphics[width=33mm]{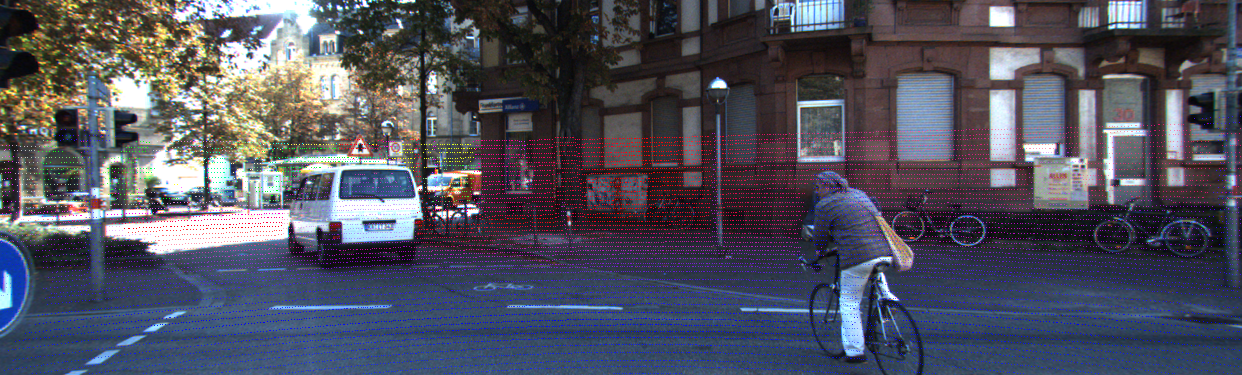}}}
    \subfloat{
    \fbox{\includegraphics[width=33mm]{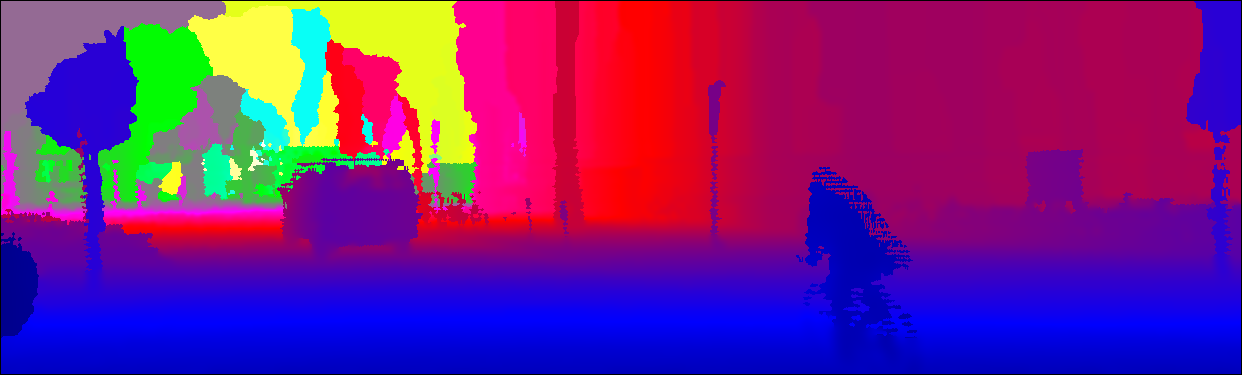}}
    \fbox{\includegraphics[width=33mm]{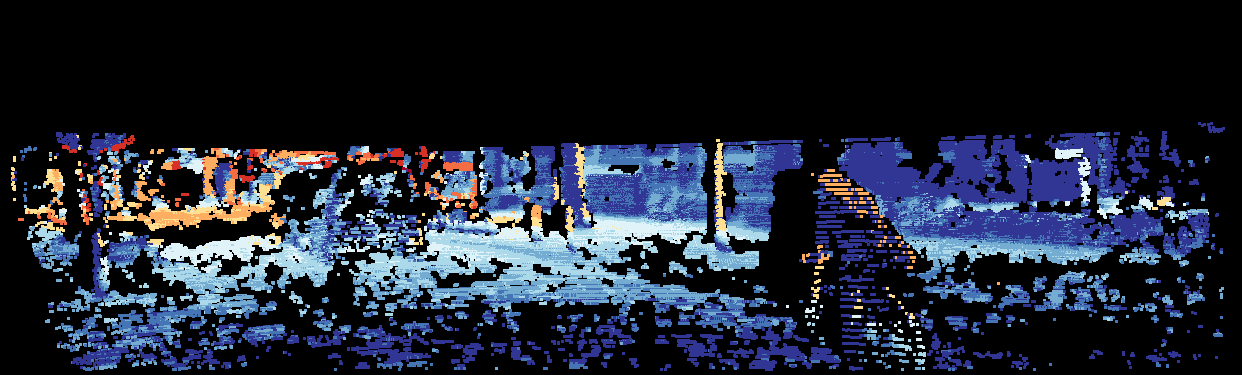}} }
    \subfloat{
    \fbox{\includegraphics[width=33mm]{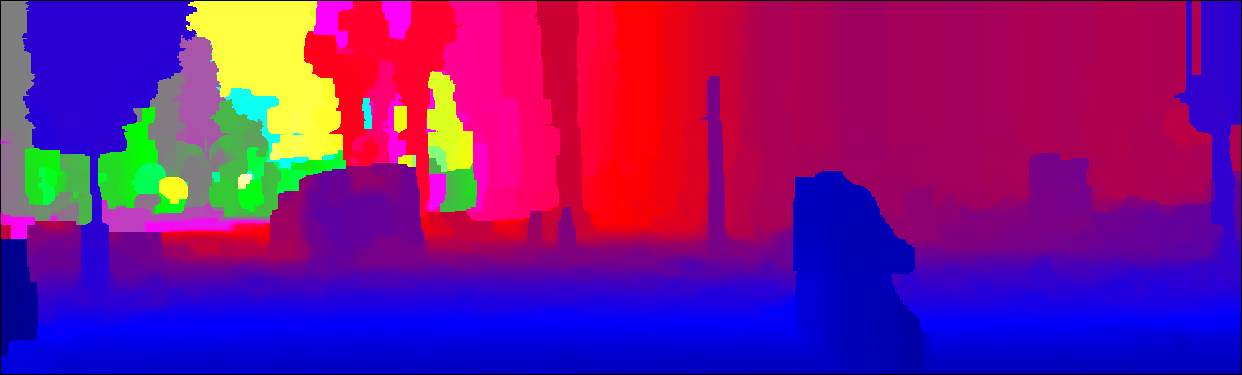}}
    \fbox{\includegraphics[width=33mm]{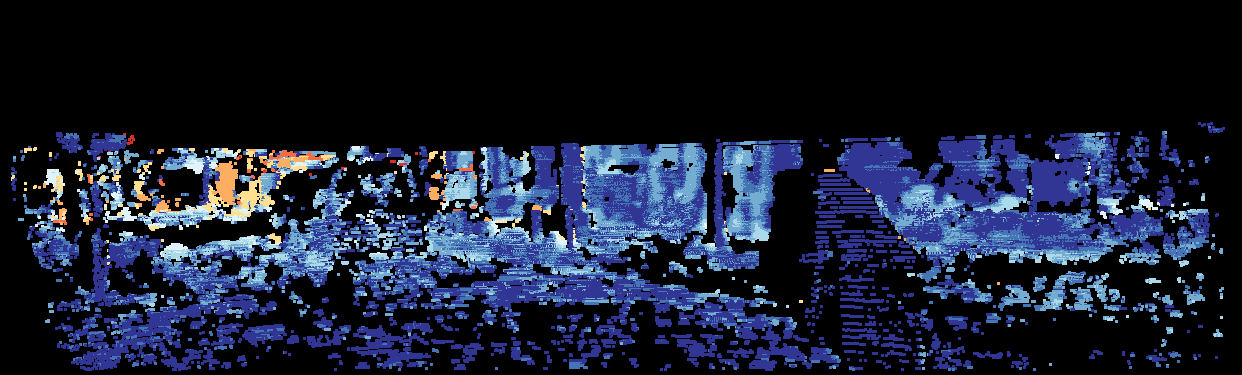}} }
    \vfil
    \vspace{-8pt}
    \subfloat{\fbox{\includegraphics[width=33mm]{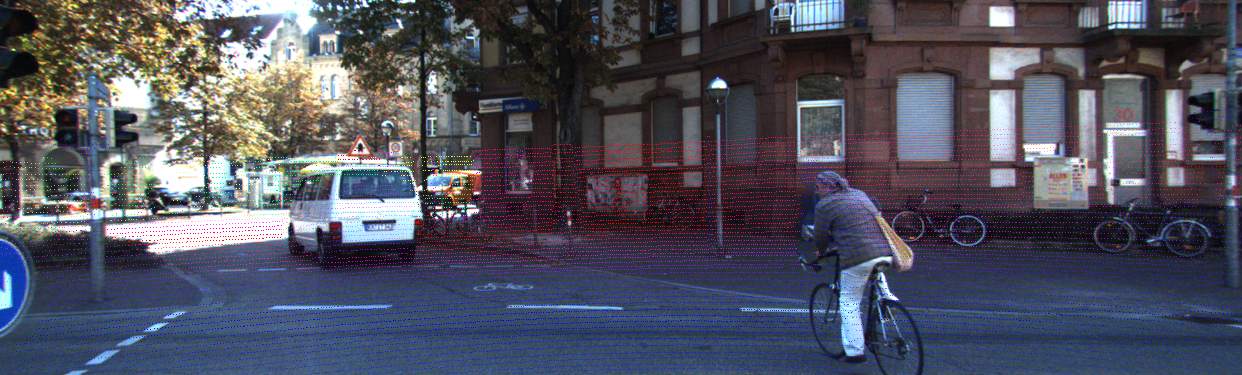}}}
    \subfloat{
    \fbox{\includegraphics[width=33mm]{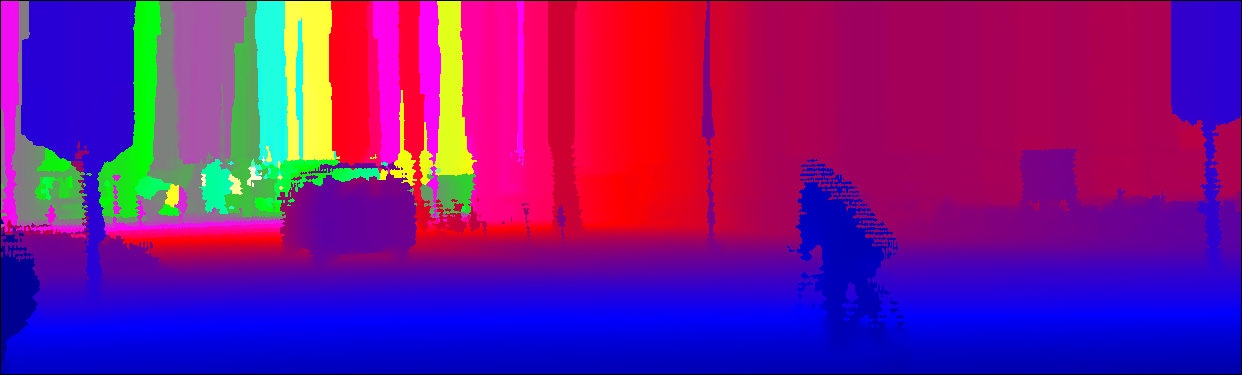}}
    \fbox{\includegraphics[width=33mm]{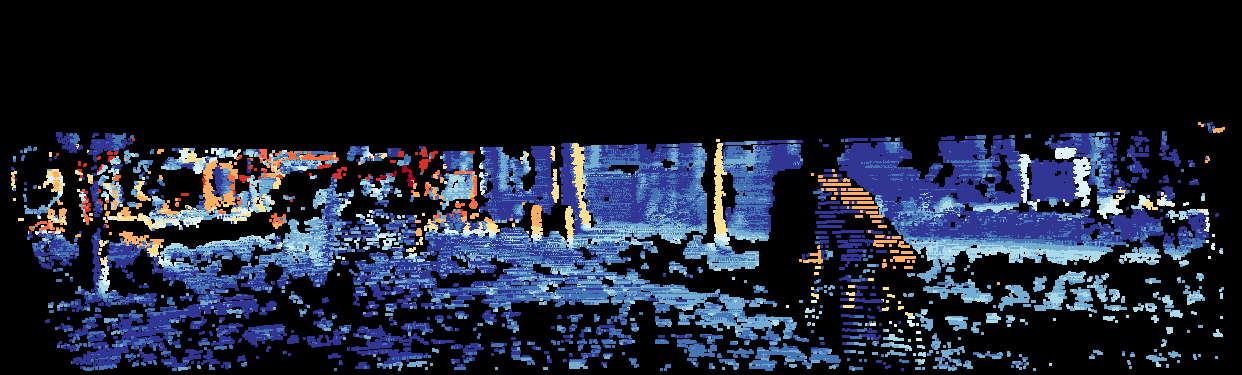}} }
    \subfloat{
    \fbox{\includegraphics[width=33mm]{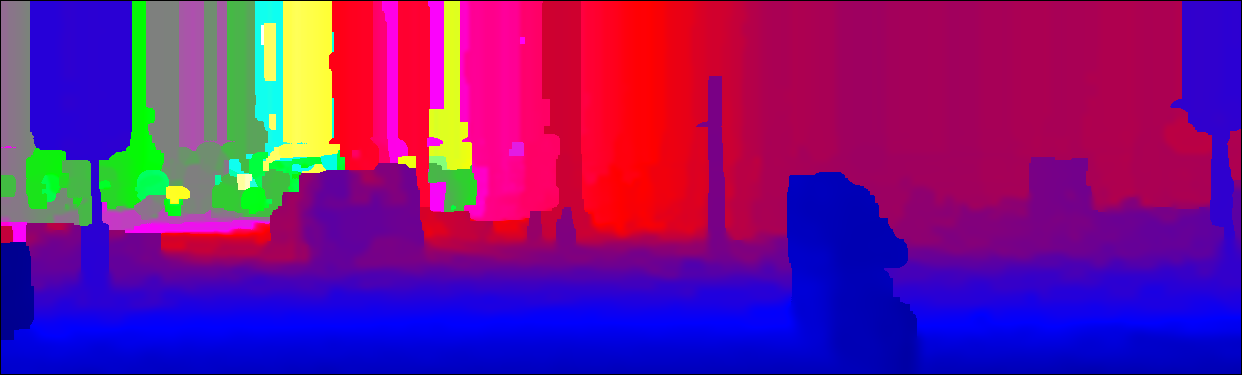}}
    \fbox{\includegraphics[width=33mm]{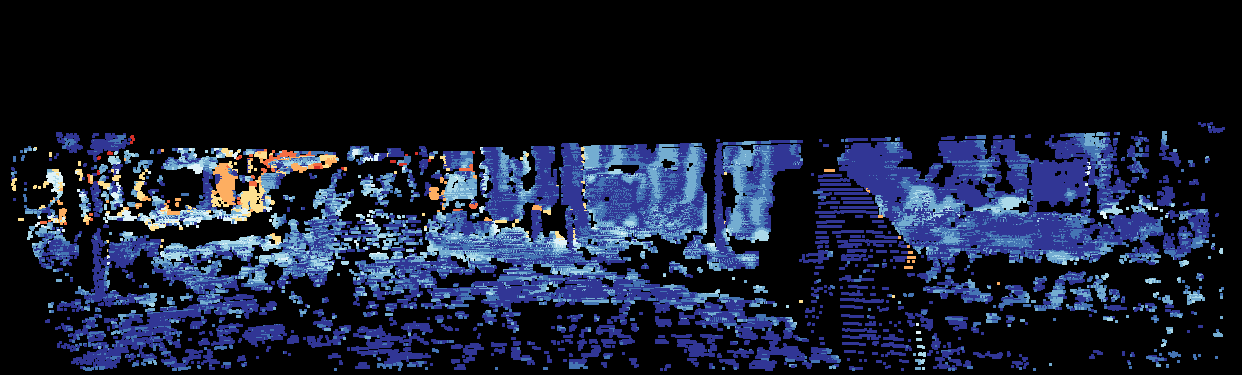}} }
    \vfil
    \vspace{-8pt}
    \setcounter{subfigure}{0}
    \subfloat[Inputs (Scene \#2)]{\fbox{\includegraphics[width=33mm]{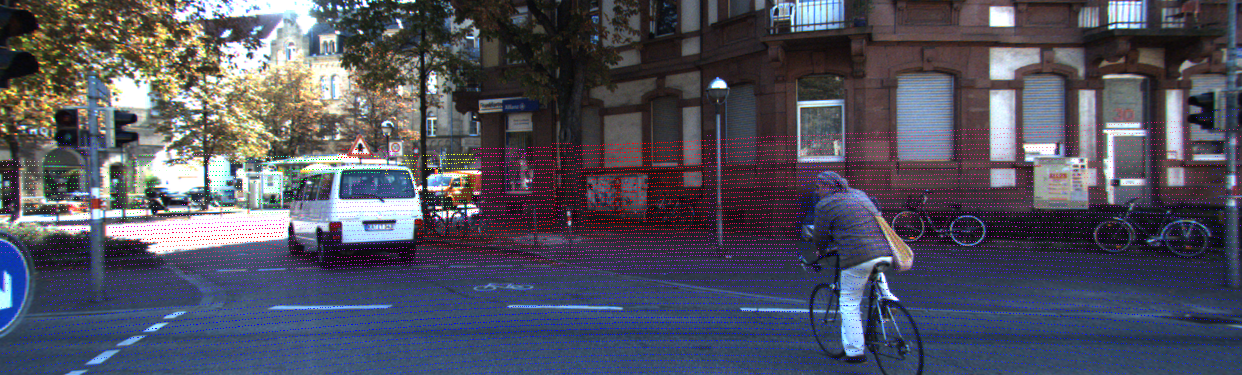}}}
    \subfloat[Results and error map of Yao et al. \cite{yao2020discontinuous} (Scene \#2)]{
    \fbox{\includegraphics[width=33mm]{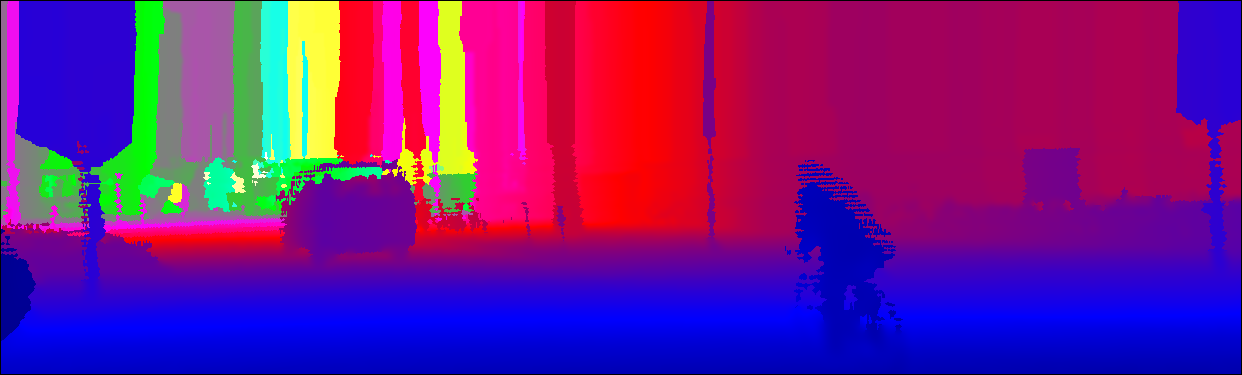}}
    \fbox{\includegraphics[width=33mm]{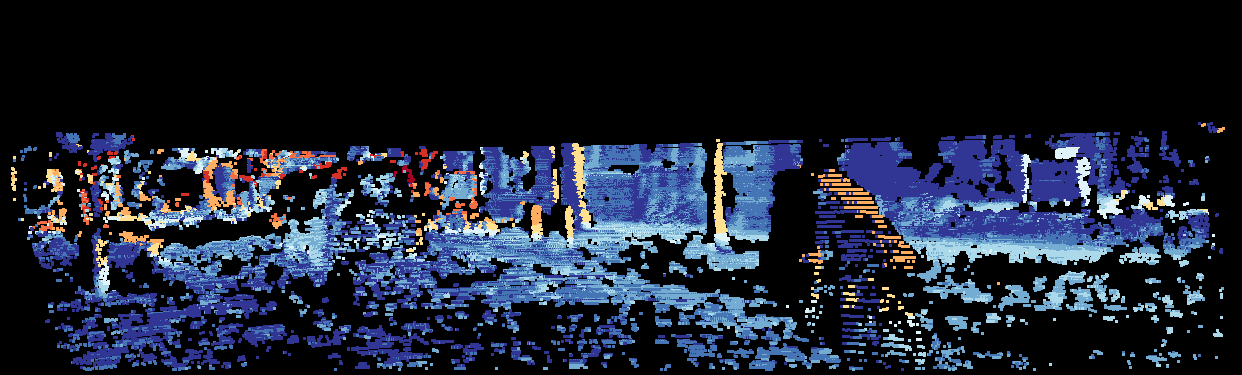}} }
    \subfloat[Results and error maps of ours (Scene \#2)]{
    \fbox{\includegraphics[width=33mm]{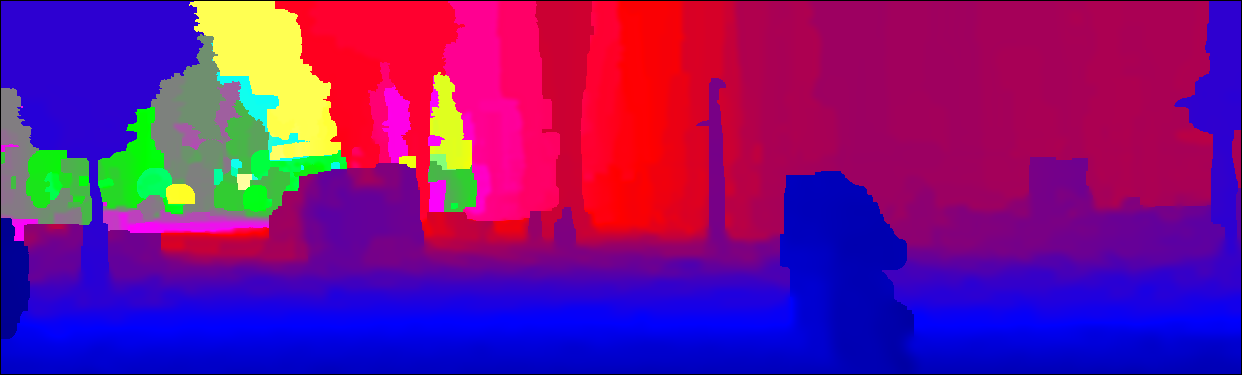}}
    \fbox{\includegraphics[width=33mm]{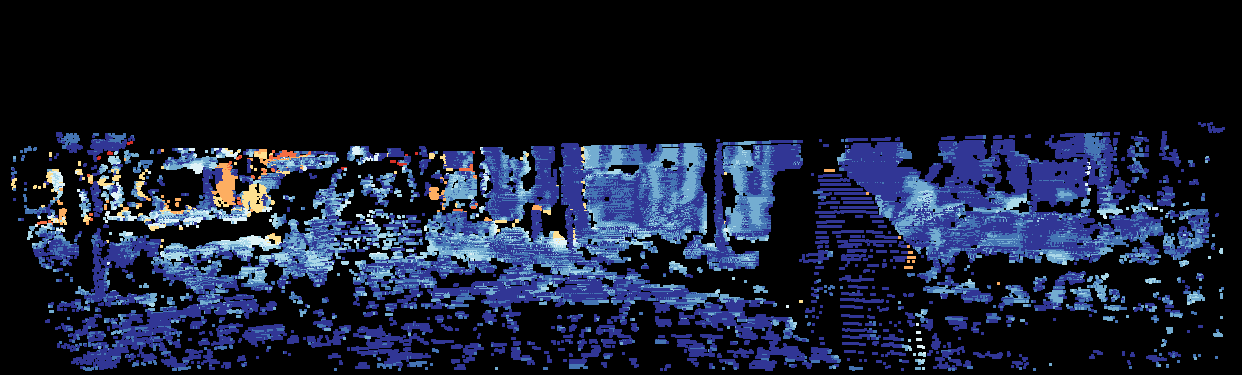}} }
    \vfil
    \subfloat{\fbox{\includegraphics[width=33mm]{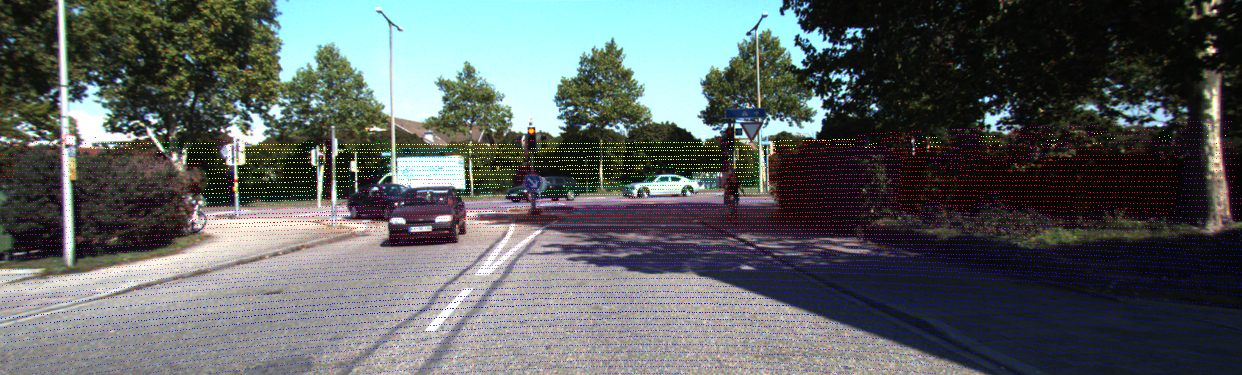}}}
    \subfloat{
    \fbox{\includegraphics[width=33mm]{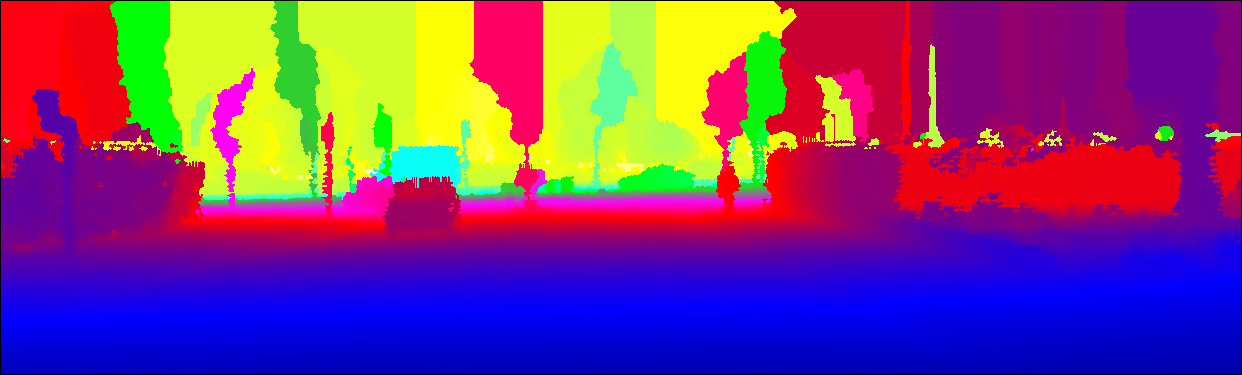}}
    \fbox{\includegraphics[width=33mm]{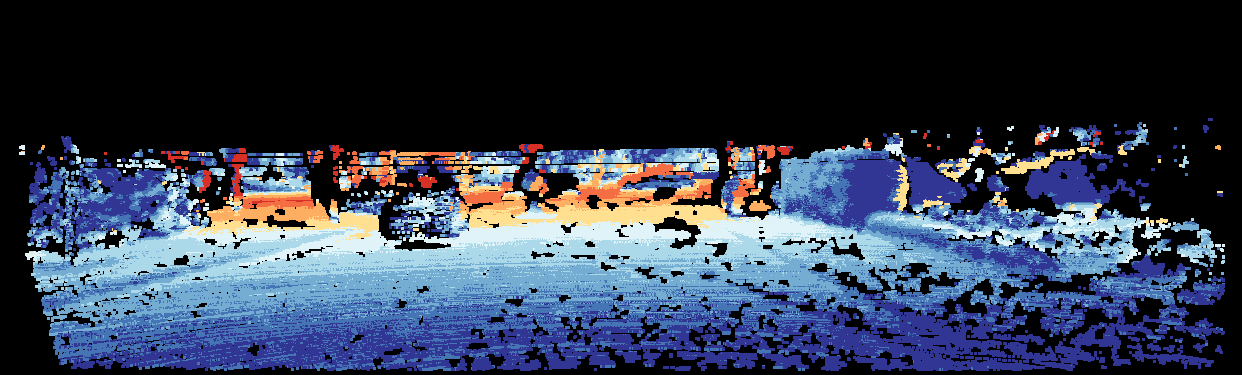}} }
    \subfloat{
    \fbox{\includegraphics[width=33mm]{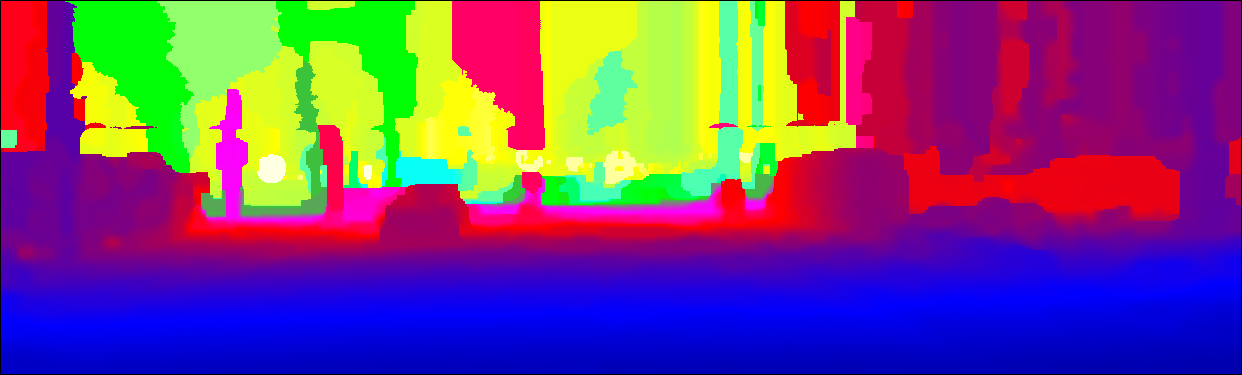}}
    \fbox{\includegraphics[width=33mm]{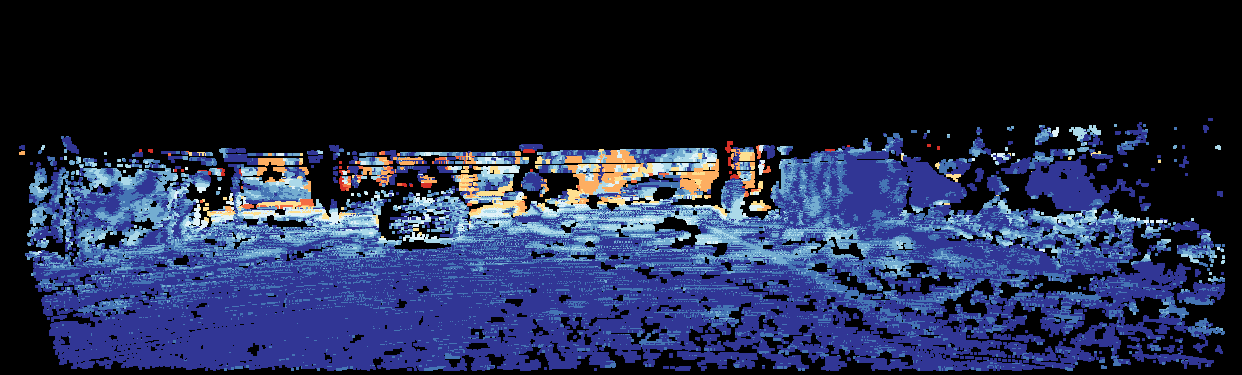}} }
    \vfil
    \vspace{-8pt}
    \subfloat{\fbox{\includegraphics[width=33mm]{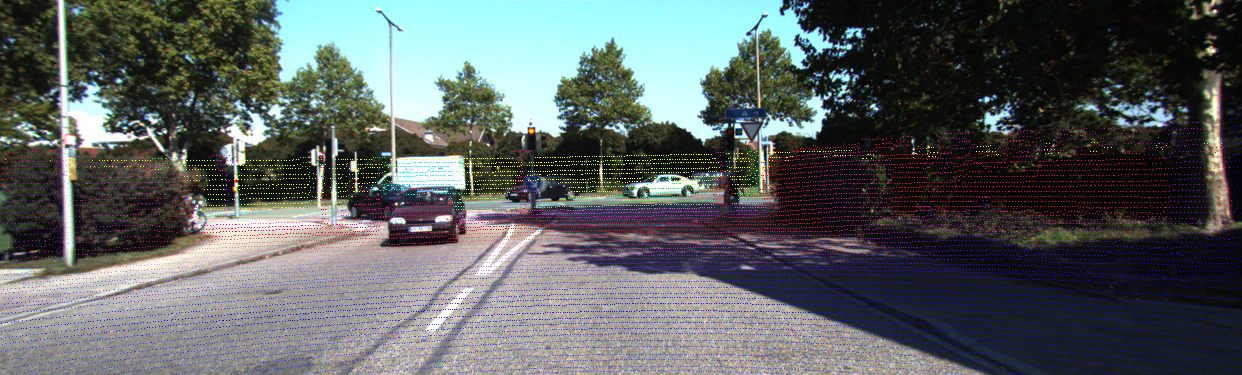}}}
    \subfloat{
    \fbox{\includegraphics[width=33mm]{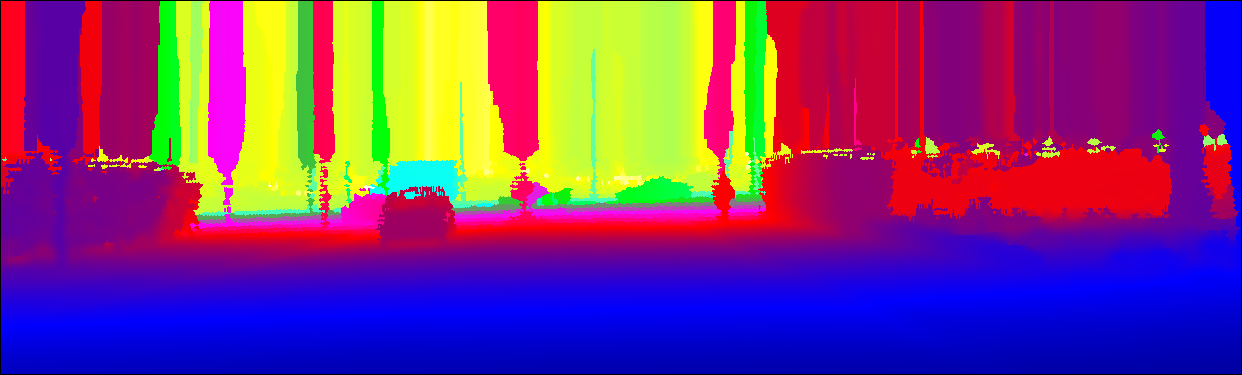}}
    \fbox{\includegraphics[width=33mm]{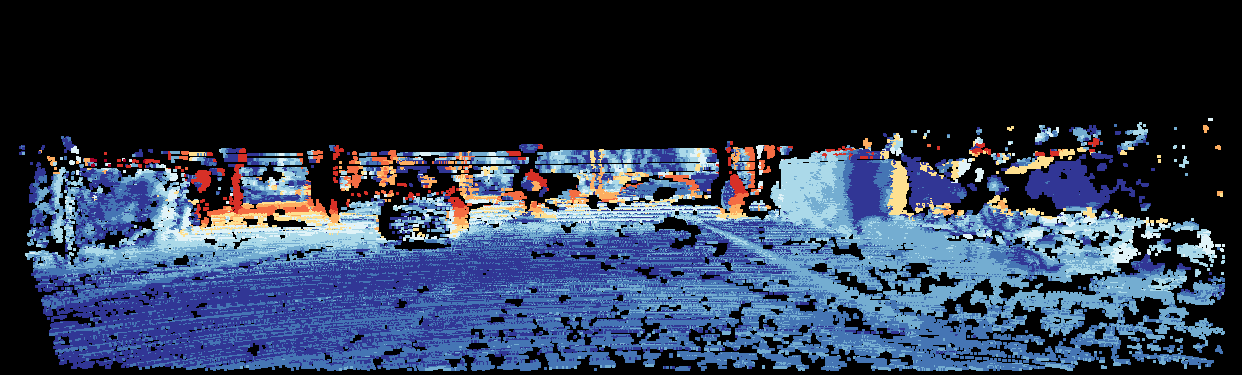}} }
    \subfloat{
    \fbox{\includegraphics[width=33mm]{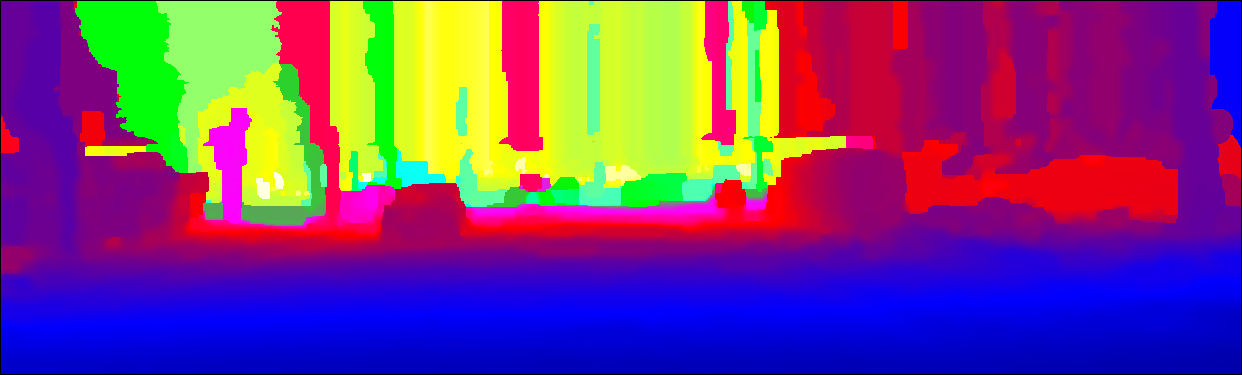}}
    \fbox{\includegraphics[width=33mm]{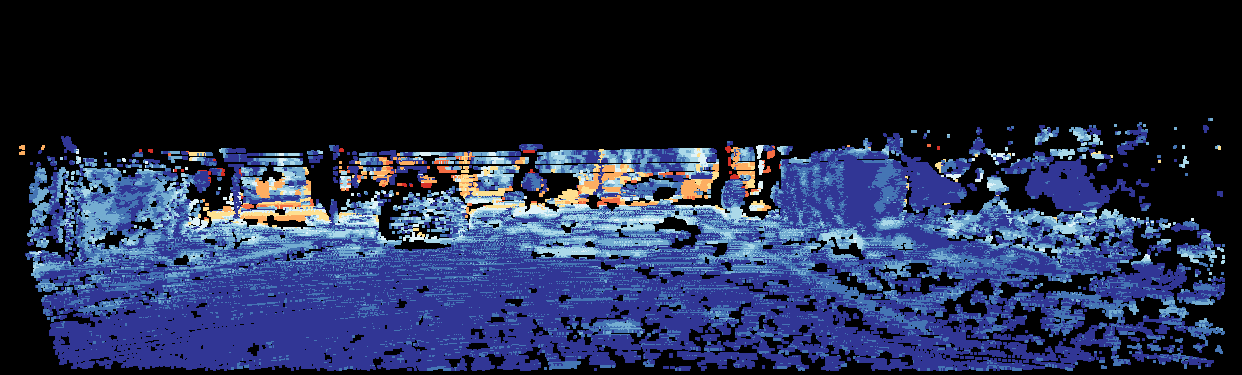}} }
    \vfil
    \vspace{-8pt}
    \setcounter{subfigure}{3}
    \subfloat[Inputs (Scene \#7)]{\fbox{\includegraphics[width=33mm]{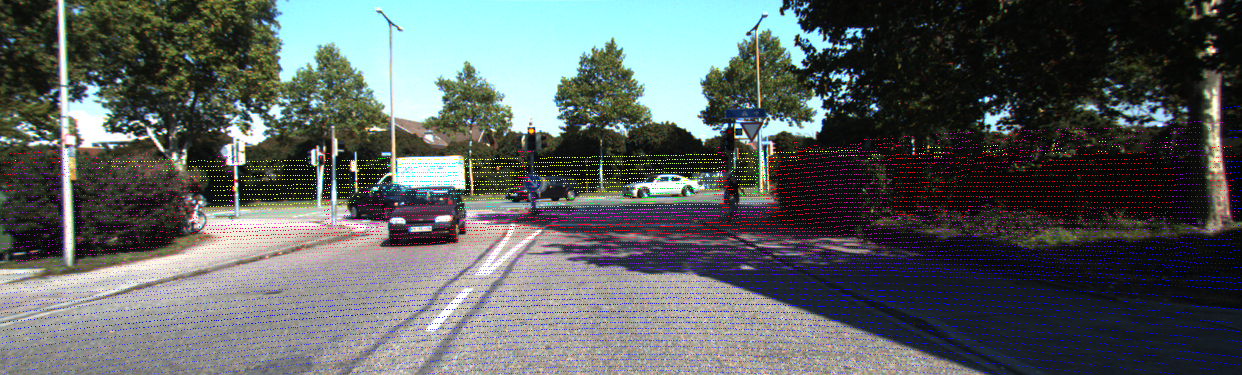}}}
    \subfloat[Results and error map of Yao et al. \cite{yao2020discontinuous} (Scene \#7)]{
    \fbox{\includegraphics[width=33mm]{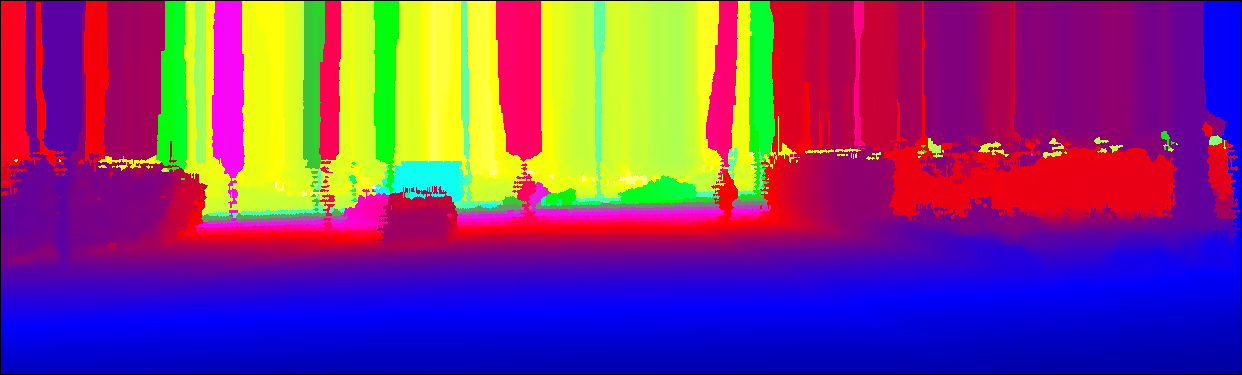}}
    \fbox{\includegraphics[width=33mm]{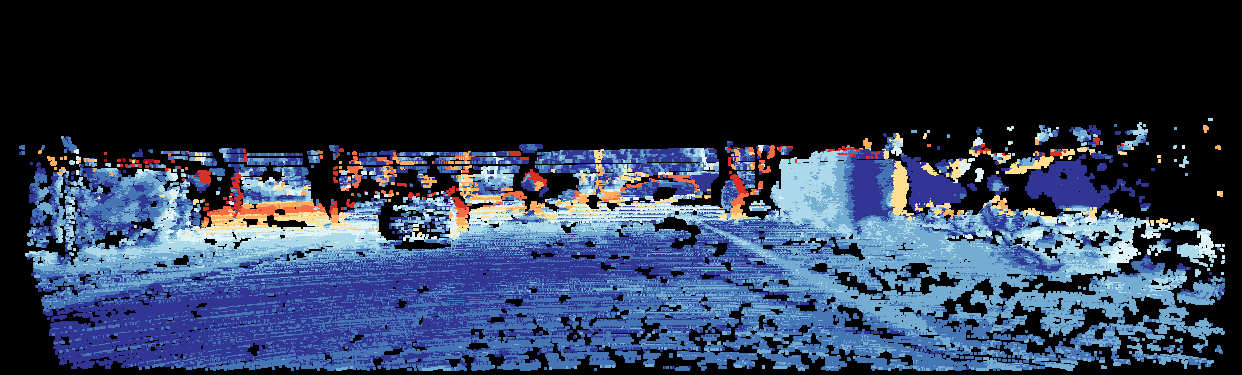}} }
    \subfloat[Results and error maps of ours (Scene \#7)]{
    \fbox{\includegraphics[width=33mm]{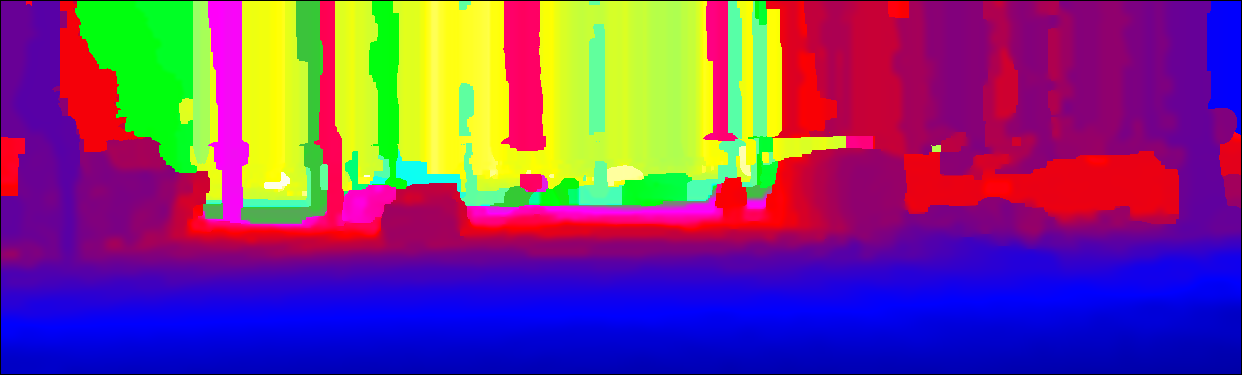}}
    \fbox{\includegraphics[width=33mm]{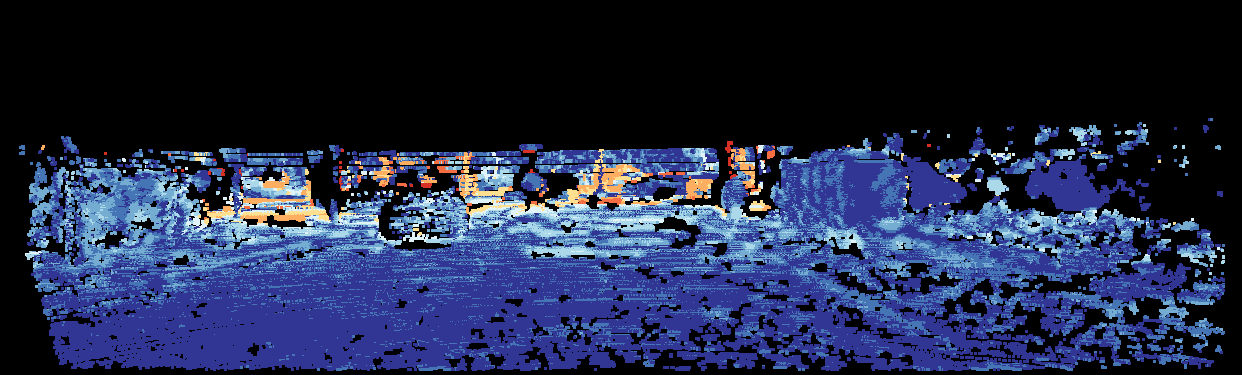}} }
    \vfil
    \caption{Inputs and results on KITTI dataset with calibration errors (top: \textit{blueprint}; middle: \textit{error-1}; and bottom: \textit{error-2}).
    The scene numbers are from the KITTI stereo dataset \cite{menze2018object}.
    Refer to Fig. \ref{fig:ref_gt} for the ground truths.
    (a), (d) The input sparse depth maps projected onto the input left image.
    (b), (c), (e), (f) The depth completion results and error maps.
    The areas of poles have smaller errors in (c), (f) compared to (b), (e).
    }
    \label{fig:result_kitti_error}
\end{figure*}
\setlength{\fboxsep}{0pt}
\begin{figure}[t]
\centering
\subfloat[Dense depth map]{\includegraphics[width=40mm]{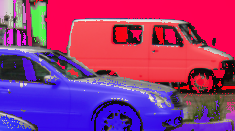}
}
\hfil
\subfloat[\textit{lines-16}]{\includegraphics[width=40mm]{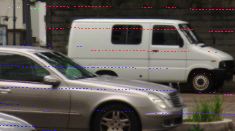}
}
\vfil
\vspace{-8pt}
\subfloat[\textit{lines-32}]{\includegraphics[width=40mm]{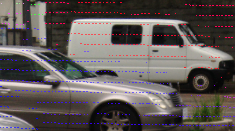}
}
\hfil
\subfloat[\textit{lines-64}]{\includegraphics[width=40mm]{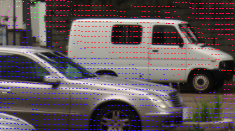}
}
\caption{
Mis-projection in the Komaba dataset.
(a) There is no mis-projection in the ground truth.
(b), (c), (d) Calibration error cause mis-projection (see the edges of vehicles).
}
\label{fig:komaba_zoom}
\end{figure}
\begin{table*}[t]
\centering
\caption{Depth completion results obtained on Komaba dataset with calibration errors}
     \begin{tabular}{c|cc|cc|cc}
     \hline
      & \multicolumn{2}{c}{\textit{lines-16}} & \multicolumn{2}{|c}{\textit{lines-32}} & \multicolumn{2}{|c}{\textit{lines-64}} \\
     \cline{2-7}
     Method & iMAE [1/m] & MAE [m] & iMAE [1/m] & MAE [m] & iMAE [1/m] & MAE [m]\\
     \hline
     Kopf et al.\cite{kopf2007joint}
     & 6.80 $\times 10^{-3}$ & 0.924 & 6.70 $\times 10^{-3}$ & 0.921 & 6.63 $\times 10^{-3}$ & 0.910\\
     Ferstl et al. 
     \cite{ferstl2013image}
      & 7.76 $\times 10^{-3}$& 1.001 & 7.68 $\times 10^{-3}$ & 1.034 & 6.64 $\times 10^{-3}$ & 1.013\\
     Yao et al.
     \cite{yao2020discontinuous}
     & 7.16 $\times 10^{-3}$ & 0.917 & 7.24 $\times 10^{-3}$ & 0.938 & 7.23 $\times 10^{-3}$ & 0.933 \\
     Ours (SSM only)
     & 6.52 $\times 10^{-3}$ & 0.866 & 6.17 $\times 10^{-3}$ & 0.827 & 6.05 $\times 10^{-3}$ & 0.816 \\
     Ours
     & \textbf{6.44} $\mathbf{\times 10^{-3}}$ & \textbf{0.857} & \textbf{6.09} $\mathbf{\times 10^{-3}}$ & \textbf{0.816} & \textbf{6.00} $\mathbf{\times 10^{-3}}$ & \textbf{0.809} \\
     \hline
\end{tabular}
\label{tab:result_komaba}
\end{table*}
\begin{figure*}
    \centering
    \subfloat{\fbox{\includegraphics[width=33mm]{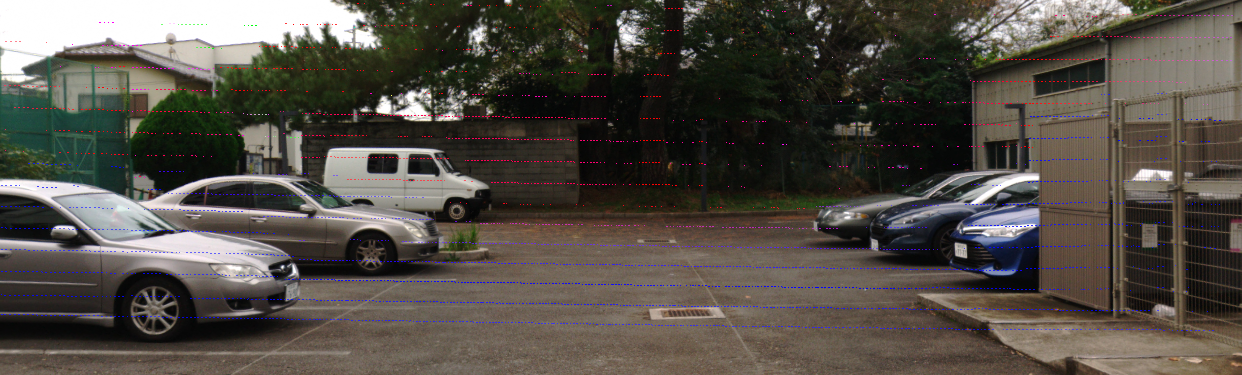}}}
    \subfloat{
    \fbox{\includegraphics[width=33mm]{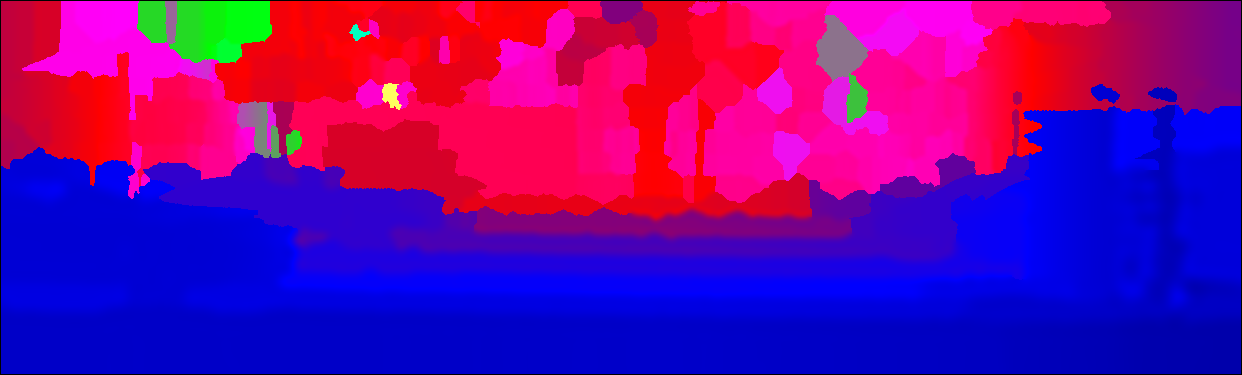}}
    \fbox{\includegraphics[width=33mm]{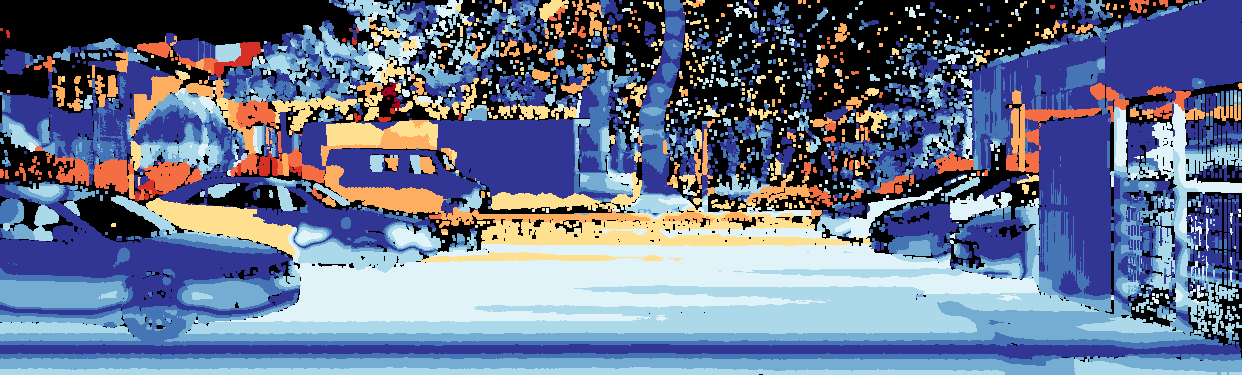}} }
    \subfloat{
    \fbox{\includegraphics[width=33mm]{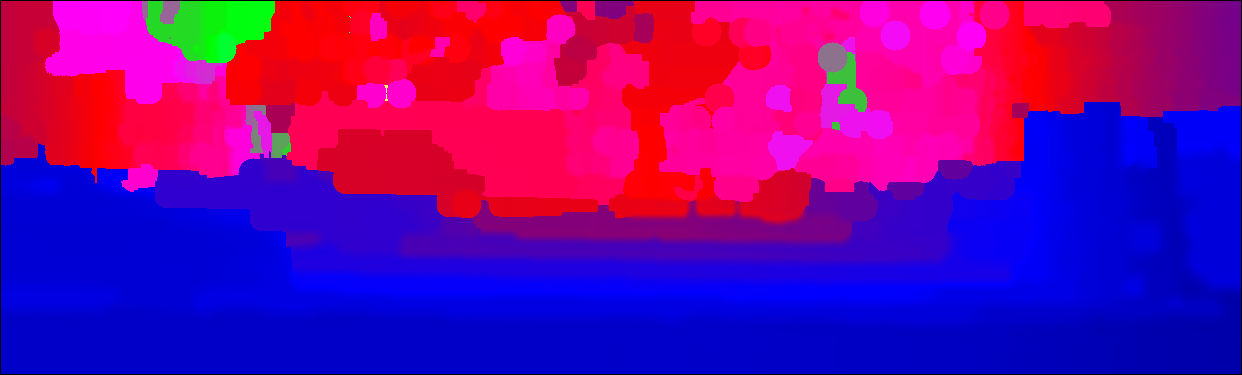}}
    \fbox{\includegraphics[width=33mm]{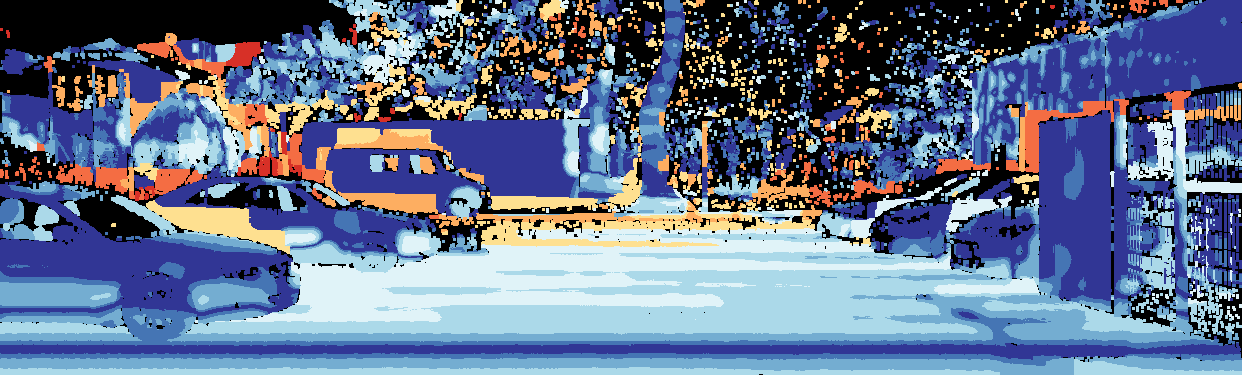}} }
    \vfil
    \vspace{-8pt}
    \subfloat{\fbox{\includegraphics[width=33mm]{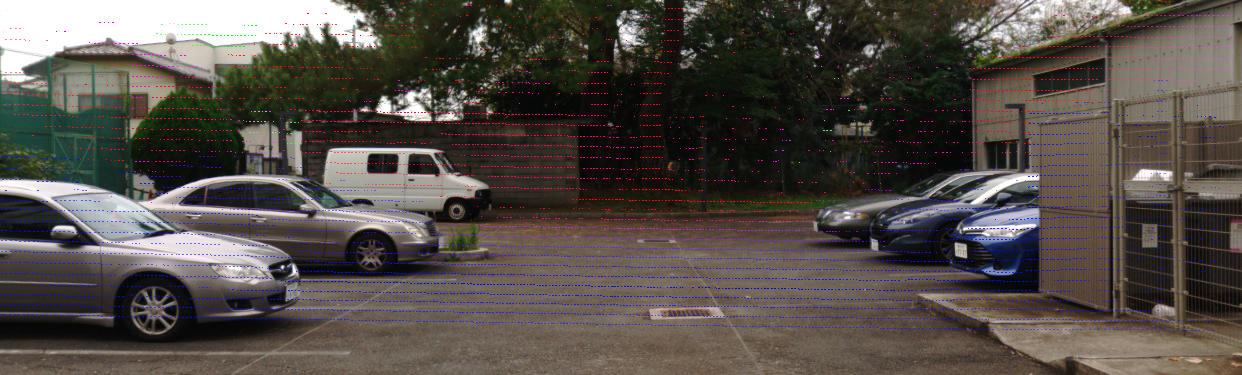}}}
    \subfloat{
    \fbox{\includegraphics[width=33mm]{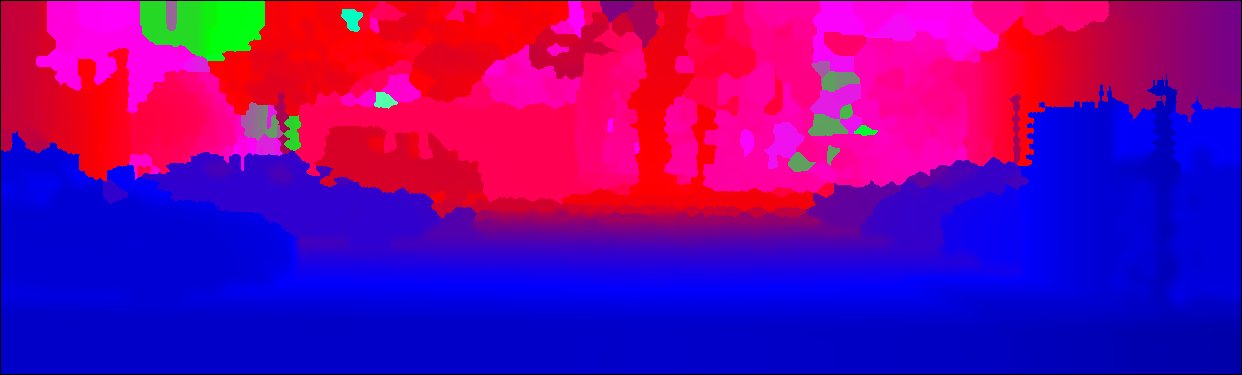}}
    \fbox{\includegraphics[width=33mm]{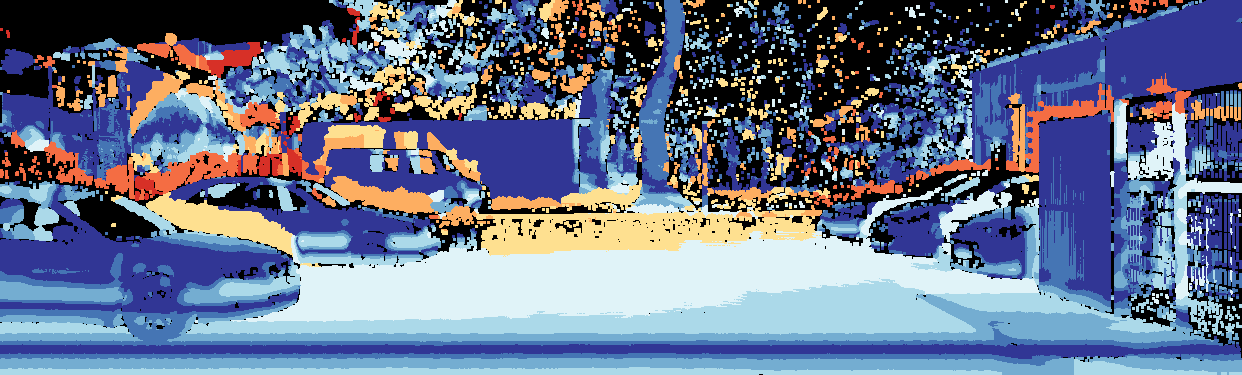}} }
    \subfloat{
    \fbox{\includegraphics[width=33mm]{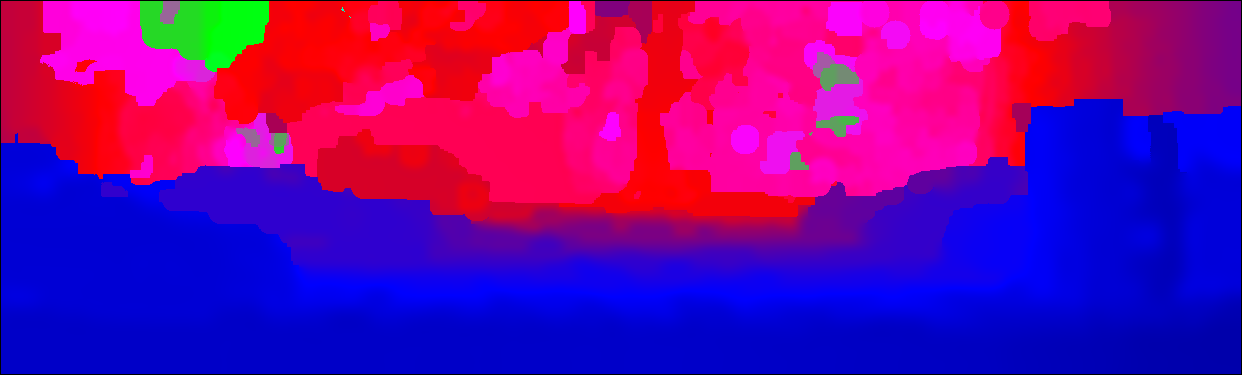}}
    \fbox{\includegraphics[width=33mm]{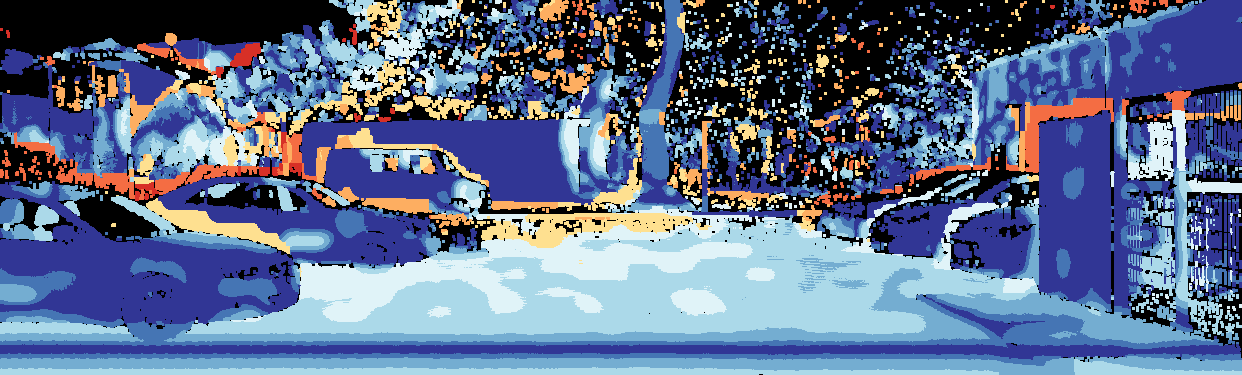}} }
    \vfil
    \vspace{-8pt}
    \setcounter{subfigure}{0}
    \subfloat[Inputs]{\fbox{\includegraphics[width=33mm]{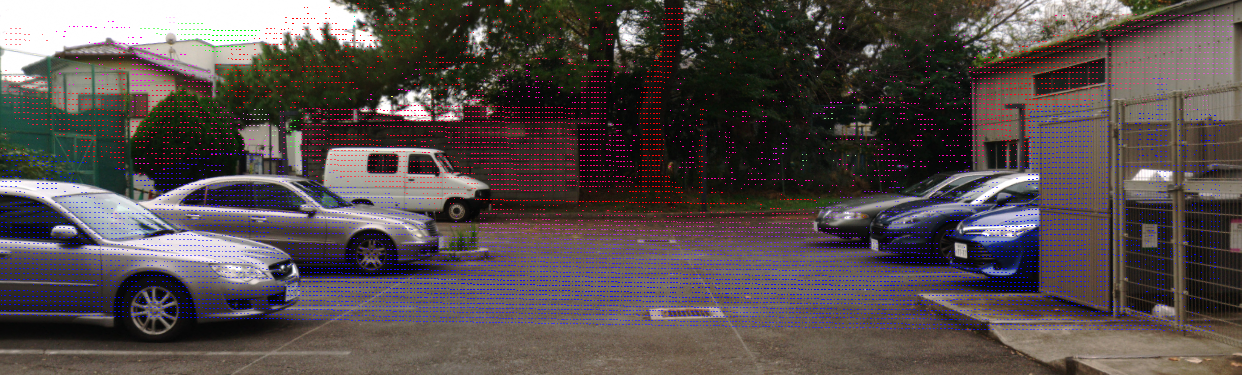}}}
    \subfloat[Results and error map of Yao et al. \cite{yao2020discontinuous}]{
    \fbox{\includegraphics[width=33mm]{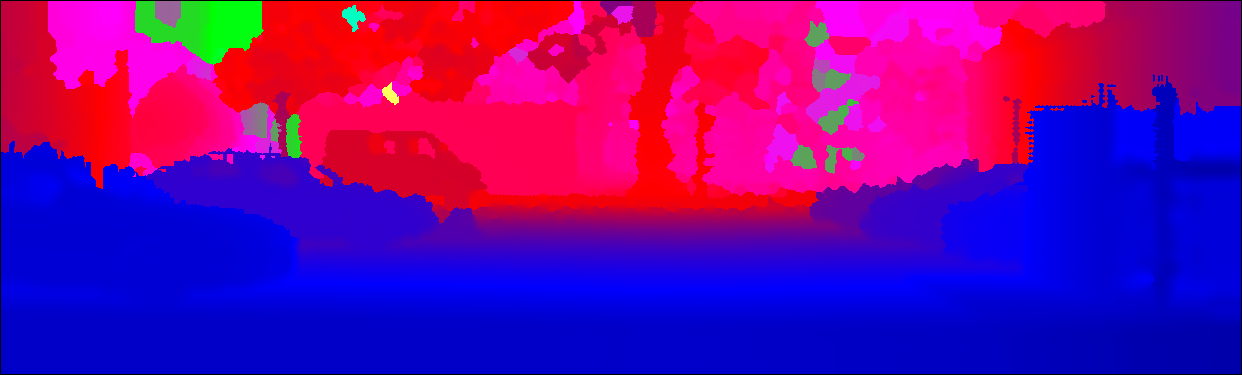}}
    \fbox{\includegraphics[width=33mm]{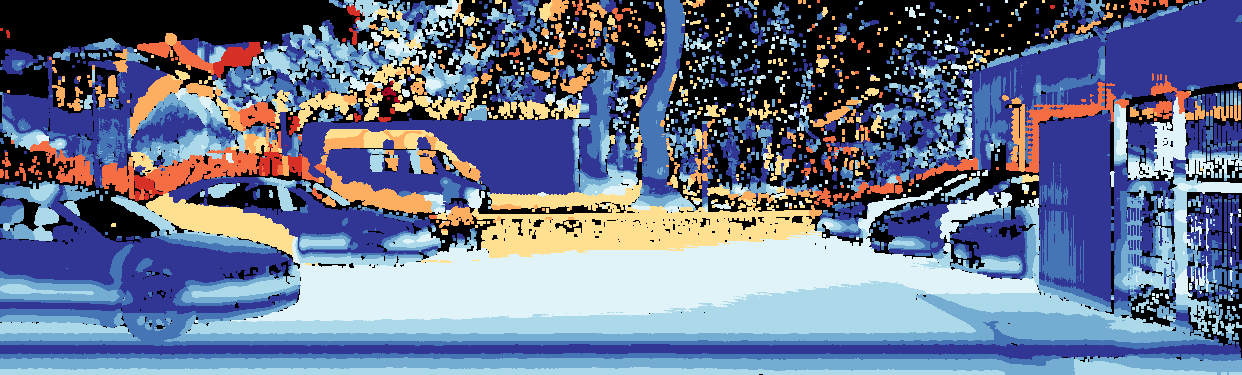}} }
    \subfloat[Results and error maps of ours]{
    \fbox{\includegraphics[width=33mm]{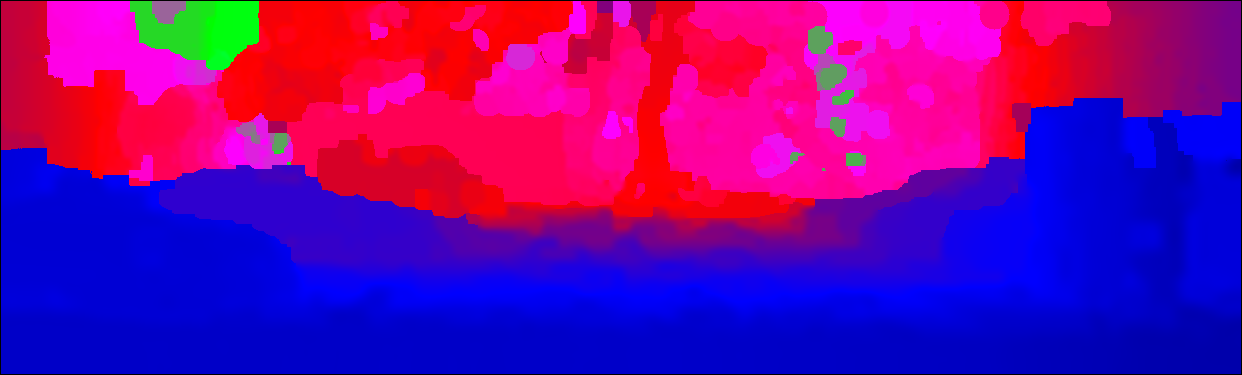}}
    \fbox{\includegraphics[width=33mm]{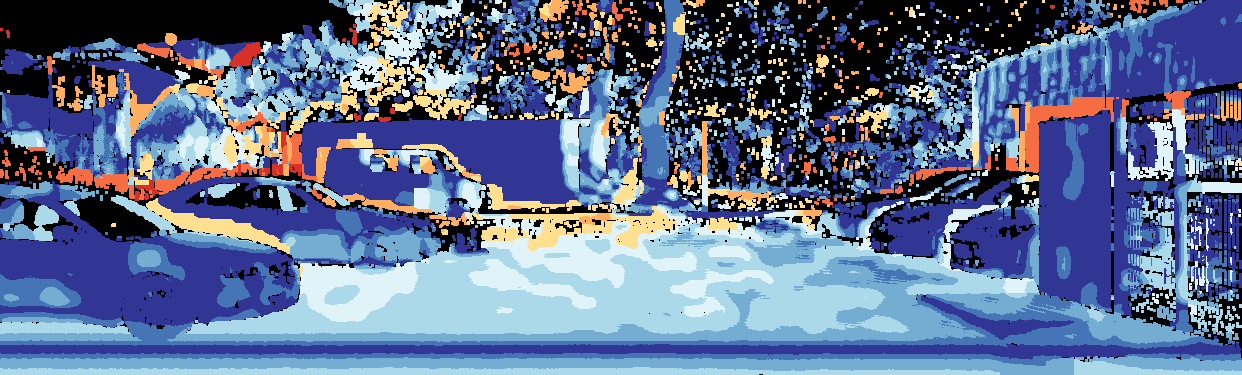}} }
    \caption{Inputs and results on Komaba dataset with calibration errors (top: \textit{lines-16}, middle: \textit{lines-32}, and bottom: \textit{lines-64}).
    Refer Fig. \ref{fig:ref_gt} for the ground truth.
    (a) The input sparse depth maps projected onto the input image.
    (b), (c) Depth completion results and error maps.
    Around the boundary of the van in the rear left, the depth is not correctly estimated by Yao's method \cite{yao2020discontinuous} and the proposed method with \textit{lines-16}.
    With more scanlines (\textit{lines-32} and \textit{lines-64}), the proposed method estimated depth more accurately around the boundary.
    }
    \label{fig:result_komaba}
\end{figure*}
We evaluated the proposed method with LiDAR camera extrinsic calibration errors.
Here, we applied random errors to the KITTI and Komaba datasets.
This comparison was performed against unsupervised methods \cite{kopf2007joint, ferstl2013image, yao2020discontinuous}.
Supervised stereo-LiDAR fusion methods were not applied because accurately calibrated scans for training are not available.
In this evaluation, the parameter settings were the same as those discussed in Section \ref{sec:eval_kitti}, except for the candidate search radius, which was set to $r=15$ [pixel].
\subsubsection{KITTI dataset}
We applied the following three error types to the KITTI dataset used in Section \ref{sec:eval_kitti}. 
\begin{itemize}
 \item \textit{blueprint} represents the extrinsic parameters before calibration, derived by the sensor setup blueprint of the KITTI dataset.
 \item \textit{error-1} represents parameters calibrated by a single-frame-marker-less method from the initial 2 [deg.] of rotation and 0.2 [m] of translation errors.
 After calibration, the average error was 0.675 [deg.] and 0.155 [m].
 \item \textit{error-2} represents parameters calibrated by a single-frame-marker-less method from the initial 4 [deg.] of rotation and 0.4 [m] of translation errors.
 After calibration, the average error was 0.667 [deg.] and 0.207 [m].
\end{itemize}
 The single-frame-marker-less calibration method to derive \textit{error-1} and \textit{error-2} is explained in the supplementary material. 
The intention of \textit{error-1} and \textit{error-2} is to emulate the worst-case calibration error expected in practical cases.

Table \ref{tab:data_errors} shows the details of the errors, and Fig \ref{fig:kitti_error} shows the visual of the calibration error in the input data.
Note that KITTI with calibration errors also has mis-projection caused by temporal and spatial occlusions in the original KITTI dataset.

Table \ref{tab:result_kitti_error} shows the results obtained on the KITTI dataset. 
The proposed method outperformed the baselines under all experimental conditions.
Moreover, by comparing with those in Table \ref{tab:eval_kitti} (the same dataset with accurate calibration), the proposed method outperformed Park's method \cite{park2018high} in terms of error rate and Cheng's method \cite{cheng2019noise} in terms of MAE, although the proposed method was applied to data with calibration errors.
The results indicate that the proposed method is robust to LiDAR-camera extrinsic calibration errors.
Figure \ref{fig:result_kitti_error} shows the results and error maps.
Figure \ref{fig:result_kitti_error} indicates that the proposed method successfully densified the depth of thin objects, e.g., poles, although the LiDAR points were not projected onto thin objects in the image.
\subsubsection{Komaba dataset}
The Komaba dataset was introduced in the literature \cite{hirata2019real} and has been used in a previous study \cite{yao2020discontinuous}.
The figure of an example frame of the Komaba dataset is in the supplementary material.
This dataset includes five frames of data comprising motion stereo image pairs and dense depth maps captured by FARO FocusS 150.
The motion between two scans is estimated by aligning the LiDAR point clouds.
There is no spatial and temporal displacement between the camera and LiDAR, and occlusions are not expected in the Komaba dataset. 

To create input sparse depth maps, we sampled the original dense depth maps and applied the randomly generated calibration errors (Table \ref{tab:data_errors}).
Here, three sampling patterns were applied to simulate different LiDAR resolutions.
\begin{itemize}
    \item \textit{lines-16} sampled 16 scanlines.
    \item \textit{lines-32} sampled 32 scanlines.
    \item \textit{lines-64} sampled 64 scanlines.
    This condition is similar to the KITTI dataset, which has 64 scanlines.
\end{itemize}
The density and mis-projection in input data are visualized in Fig. \ref{fig:komaba_zoom}.

Table \ref{tab:result_komaba} shows the results obtained on the Komaba dataset.
Here, rather than the error rate, we evaluated the inverse MAE (iMAE) because the Komaba dataset does not provide the ground truth of the disparity maps. 
The iMAE evaluates the accuracy of the inverse of the depth, which is proportional to the disparity.
The proposed method outperformed the baselines under all experimental conditions.
However, we observed a performance degradation with the proposed method as the number of scanlines decreased.
This reduction in performance occurred because there was less possibility to find an appropriate value near the target pixel if scanlines are sparse.
The relationship between the number of scanlines and performance is visually confirmed in Fig. \ref{fig:result_komaba}.
\subsection{Parameter study}
\label{sec:study}
We evaluated the effect on MAE of the value of $r$ using the KITTI dataset (Section \ref{sec:eval_kitti}), the KITTI dataset with the \textit{blueprint} condition, and the Komaba dataset with the \textit{lines-64} condition (Section \ref{sec:eval_kitti_error}).
The results are shown in Table \ref{tab:parameter}.

In Table \ref{tab:parameter}, the value of $r^{\ast}$ derived from Eq. \eqref{eq:max_r} reside close to the $r$ value to give the minimum MAE for every data.
Here, we derived $r^{\ast}$ for each data using Eq. \eqref{eq:max_r} as follows.
\begin{itemize}
\item KITTI: $r^{\ast}=6.68$ with $f=959.7915$, $\theta_{\mathrm{calib}}=0$, and $\theta_{\mathrm{scan}}=0.4$. 
\item KITTI (\textit{blueprint}): $r^{\ast}=15.90$ with $f=959.791$, $\theta_{\mathrm{calib}}=0.952$, and $\theta_{\mathrm{scan}}=0.4$.
\item Komaba (\textit{lines-64}): $r^{\ast}=16.453$ with $f=956.925$ and $\theta_{\mathrm{calib}}=1.096$. 
Note that, in this case, we ignored $\sigma_{\mathrm{scan}}$ because occlusion was not expected.
\end{itemize}
Hence, the results supports our approach to set $r$ in Section \ref{sec:param}.
\begin{figure}[t]
    \centering
    \subfloat[KITTI (Scene \#2)]{\includegraphics[width=40mm]{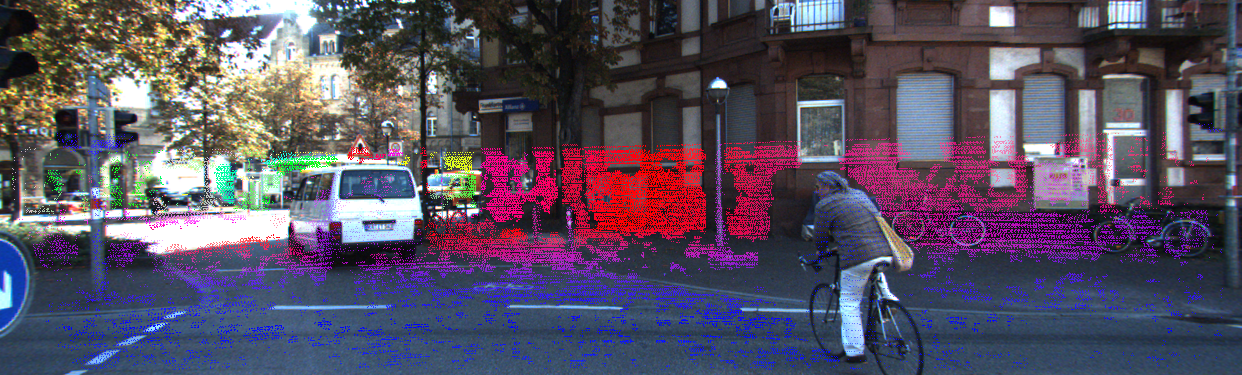}}
    \hfil
    \subfloat[KITTI (Scene \#7)]{\includegraphics[width=40mm]{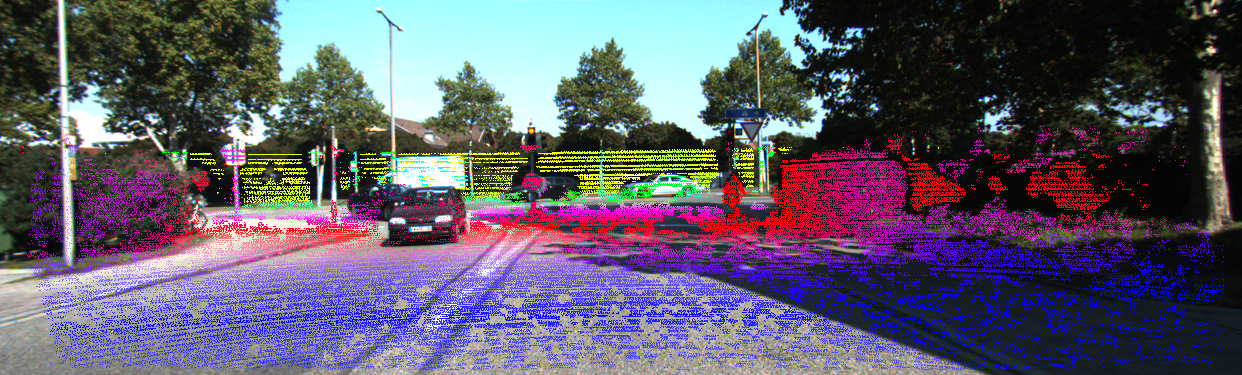}}
    \vfil \vspace{-8pt}
    \subfloat[Komaba]{\includegraphics[width=33mm]{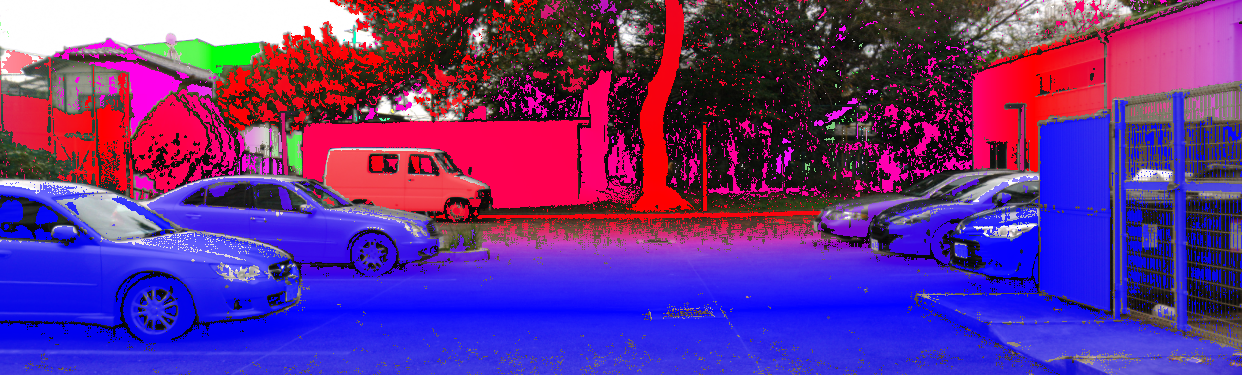}}
    \caption{Ground truth depth maps for Fig. \ref{fig:result_kitti_error} and \ref{fig:result_komaba}.}
    \label{fig:ref_gt}
\end{figure}
\begin{table}[t]
    \centering
    \caption{MAE variation by changing the radius in SSM}
    \begin{tabular}{c|ccc}
    \hline
    & \multicolumn{3}{c}{MAE [m]} \\
    \cline{2-4}
         & KITTI
         & KTTI w/ error
         & Komaba w/ error
         \\
         Radius [pixels]&
         & (\textit{blueprint})
         & (\textit{lines-64})\\
         \hline
         $r=2$ & 0.374 & 1.575 & 0.949 \\
         $r=5$ & \textbf{0.356} & 1.157 & 0.892 \\
         $r=10$ & 0.442 & 0.565 & 0.814 \\
         $r=15$ & 0.493 & \textbf{0.528} & \textbf{0.809} \\
         $r=20$ & 0.530 & 0.573 & 0.860 \\
         \hline
    \end{tabular}
    \label{tab:parameter}
\end{table}
\section{Conclusion}
\label{sec:limitation}
We proposed a non-learning stereo-aided depth completion method that is robust to mis-projection and preserves LiDAR precision in the long range.
Unlike previous methods, our method does not require accurate LiDAR-stereo extrinsic calibration parameters in any part of its process.
Therefore, it is applicable in the conditions that the calibration is difficult to conduct.
In the evaluations, our method demonstrated smaller MAEs than previous state-of-the-art stereo-aided depth completion methods.

Our proposal is composed of SSM and the framework combining SSM and B-ADT aided smoothing.
SSM searches for an optimal depth value for each pixel from its neighborly projected LiDAR points by an energy minimization approach, which can handle any type of mis-projection.
In addition, we apply B-ADT-aided smoothing \cite{yao2020discontinuous} to generate boundary-preseriving continuous depth maps since SSM is discrete optimization.

The current limitations of the proposed method include the accuracy dependency on the LiDAR scan density, as demonstrated by the evaluation discussed in Section \ref{sec:eval_kitti_error}.

We aim to extend our approach to run in real time for applying it to actual robotic systems.
Since most of our processing time comes from LBP of SSM candidate selection (0.943 out of 0.999 [s]), we consider improving the selection process to be able to adapt faster optimizers.

\bibliography{IEEEabrv, manuscript}

\begin{thebibliography}{10}
\providecommand{\url}[1]{#1}
\csname url@samestyle\endcsname
\providecommand{\newblock}{\relax}
\providecommand{\bibinfo}[2]{#2}
\providecommand{\BIBentrySTDinterwordspacing}{\spaceskip=0pt\relax}
\providecommand{\BIBentryALTinterwordstretchfactor}{4}
\providecommand{\BIBentryALTinterwordspacing}{\spaceskip=\fontdimen2\font plus
\BIBentryALTinterwordstretchfactor\fontdimen3\font minus
  \fontdimen4\font\relax}
\providecommand{\BIBforeignlanguage}[2]{{%
\expandafter\ifx\csname l@#1\endcsname\relax
\typeout{** WARNING: IEEEtran.bst: No hyphenation pattern has been}%
\typeout{** loaded for the language `#1'. Using the pattern for}%
\typeout{** the default language instead.}%
\else
\language=\csname l@#1\endcsname
\fi
#2}}
\providecommand{\BIBdecl}{\relax}
\BIBdecl

\bibitem{batlle1998recent}
J.~Batlle, E.~Mouaddib, and J.~Salvi, ``Recent progress in coded structured
  light as a technique to solve the correspondence problem: a survey,''
  \emph{Pattern recognition}, vol.~31, no.~7, pp. 963--982, 1998.

\bibitem{kopf2007joint}
J.~Kopf, M.~F. Cohen, D.~Lischinski, and M.~Uyttendaele, ``Joint bilateral
  upsampling,'' in \emph{ACM Transactions on Graphics (ToG)}, vol.~26,
  no.~3.\hskip 1em plus 0.5em minus 0.4em\relax ACM, 2007, p.~96.

\bibitem{ferstl2013image}
D.~Ferstl, C.~Reinbacher, R.~Ranftl, M.~R{\"u}ther, and H.~Bischof, ``Image
  guided depth upsampling using anisotropic total generalized variation,'' in
  \emph{International Conference on Computer Vision (ICCV)}, 2013, pp.
  993--1000.

\bibitem{diebel2006application}
J.~Diebel and S.~Thrun, ``An application of markov random fields to range
  sensing,'' in \emph{Advances in neural information processing systems}, 2006,
  pp. 291--298.

\bibitem{schneider2016semantically}
N.~Schneider, L.~Schneider, P.~Pinggera, U.~Franke, M.~Pollefeys, and
  C.~Stiller, ``Semantically guided depth upsampling,'' in \emph{German
  conference on pattern recognition}.\hskip 1em plus 0.5em minus 0.4em\relax
  Springer, 2016, pp. 37--48.

\bibitem{ma2019self}
F.~Ma, G.~V. Cavalheiro, and S.~Karaman, ``Self-supervised sparse-to-dense:
  Self-supervised depth completion from lidar and monocular camera,'' in
  \emph{International Conference on Robotics and Automation (ICRA)}.\hskip 1em
  plus 0.5em minus 0.4em\relax IEEE, 2019, pp. 3288--3295.

\bibitem{wong2020unsupervised}
A.~Wong, X.~Fei, S.~Tsuei, and S.~Soatto, ``Unsupervised depth completion from
  visual inertial odometry,'' \emph{IEEE Robotics and Automation Letters},
  vol.~5, no.~2, pp. 1899--1906, 2020.

\bibitem{yao2020discontinuous}
Y.~Yao, M.~Roxas, R.~Ishikawa, S.~Ando, J.~Shimamura, and T.~Oishi,
  ``Discontinuous and smooth depth completion with binary anisotropic diffusion
  tensor,'' \emph{IEEE Robotics and Automation Letters}, vol.~5, no.~4, pp.
  5128--5135, 2020.

\bibitem{liu2021fcfr}
L.~Liu, X.~Song, X.~Lyu, J.~Diao, M.~Wang, Y.~Liu, and L.~Zhang, ``Fcfr-net:
  Feature fusion based coarse-to-fine residual learning for depth completion,''
  in \emph{Proceedings of the AAAI Conference on Artificial Intelligence},
  vol.~35, no.~3, 2021, pp. 2136--2144.

\bibitem{park2020non}
J.~Park, K.~Joo, Z.~Hu, C.-K. Liu, and I.-S. Kweon, ``Non-local spatial
  propagation network for depth completion,'' in \emph{European Conference on
  Computer Vision, ECCV 2020}.\hskip 1em plus 0.5em minus 0.4em\relax European
  Conference on Computer Vision, 2020.

\bibitem{cheng2020cspn++}
X.~Cheng, P.~Wang, C.~Guan, and R.~Yang, ``Cspn++: Learning context and
  resource aware convolutional spatial propagation networks for depth
  completion,'' in \emph{Proceedings of the AAAI Conference on Artificial
  Intelligence}, vol.~34, no.~07, 2020, pp. 10\,615--10\,622.

\bibitem{zhao2021adaptive}
S.~Zhao, M.~Gong, H.~Fu, and D.~Tao, ``Adaptive context-aware multi-modal
  network for depth completion,'' \emph{IEEE Transactions on Image Processing},
  2021.

\bibitem{chen2019learning}
Y.~Chen, B.~Yang, M.~Liang, and R.~Urtasun, ``Learning joint 2d-3d
  representations for depth completion,'' in \emph{Proceedings of the IEEE/CVF
  International Conference on Computer Vision}, 2019, pp. 10\,023--10\,032.

\bibitem{qiu2019deeplidar}
J.~Qiu, Z.~Cui, Y.~Zhang, X.~Zhang, S.~Liu, B.~Zeng, and M.~Pollefeys,
  ``Deeplidar: Deep surface normal guided depth prediction for outdoor scene
  from sparse lidar data and single color image,'' in \emph{Proceedings of the
  IEEE/CVF Conference on Computer Vision and Pattern Recognition}, 2019, pp.
  3313--3322.

\bibitem{li2020multi}
A.~Li, Z.~Yuan, Y.~Ling, W.~Chi, C.~Zhang \emph{et~al.}, ``A multi-scale guided
  cascade hourglass network for depth completion,'' in \emph{Proceedings of the
  IEEE/CVF Winter Conference on Applications of Computer Vision}, 2020, pp.
  32--40.

\bibitem{van2019sparse}
W.~Van~Gansbeke, D.~Neven, B.~De~Brabandere, and L.~Van~Gool, ``Sparse and
  noisy lidar completion with rgb guidance and uncertainty,'' in \emph{2019
  16th international conference on machine vision applications (MVA)}.\hskip
  1em plus 0.5em minus 0.4em\relax IEEE, 2019, pp. 1--6.

\bibitem{xu2019depth}
Y.~Xu, X.~Zhu, J.~Shi, G.~Zhang, H.~Bao, and H.~Li, ``Depth completion from
  sparse lidar data with depth-normal constraints,'' in \emph{Proceedings of
  the IEEE/CVF International Conference on Computer Vision}, 2019, pp.
  2811--2820.

\bibitem{yan2020revisiting}
L.~Yan, K.~Liu, and E.~Belyaev, ``Revisiting sparsity invariant convolution: A
  network for image guided depth completion,'' \emph{IEEE Access}, vol.~8, pp.
  126\,323--126\,332, 2020.

\bibitem{eldesokey2019confidence}
A.~Eldesokey, M.~Felsberg, and F.~S. Khan, ``Confidence propagation through
  cnns for guided sparse depth regression,'' \emph{IEEE transactions on pattern
  analysis and machine intelligence}, vol.~42, no.~10, pp. 2423--2436, 2019.

\bibitem{schuster2021ssgp}
R.~Schuster, O.~Wasenmuller, C.~Unger, and D.~Stricker, ``Ssgp: Sparse spatial
  guided propagation for robust and generic interpolation,'' in
  \emph{Proceedings of the IEEE/CVF Winter Conference on Applications of
  Computer Vision}, 2021, pp. 197--206.

\bibitem{bai2020depthnet}
L.~Bai, Y.~Zhao, M.~Elhousni, and X.~Huang, ``Depthnet: Real-time lidar point
  cloud depth completion for autonomous vehicles,'' \emph{IEEE Access}, 2020.

\bibitem{shivakumar2019dfusenet}
S.~S. Shivakumar, T.~Nguyen, I.~D. Miller, S.~W. Chen, V.~Kumar, and C.~J.
  Taylor, ``Dfusenet: Deep fusion of rgb and sparse depth information for image
  guided dense depth completion,'' in \emph{2019 IEEE Intelligent
  Transportation Systems Conference (ITSC)}.\hskip 1em plus 0.5em minus
  0.4em\relax IEEE, 2019, pp. 13--20.

\bibitem{geiger2013vision}
A.~Geiger, P.~Lenz, C.~Stiller, and R.~Urtasun, ``Vision meets robotics: The
  kitti dataset,'' \emph{The International Journal of Robotics Research},
  vol.~32, no.~11, pp. 1231--1237, 2013.

\bibitem{geiger2012automatic}
A.~Geiger, F.~Moosmann, {\"O}.~Car, and B.~Schuster, ``Automatic camera and
  range sensor calibration using a single shot,'' in \emph{2012 IEEE
  International Conference on Robotics and Automation}.\hskip 1em plus 0.5em
  minus 0.4em\relax IEEE, 2012, pp. 3936--3943.

\bibitem{pandey2012automatic}
G.~Pandey, J.~McBride, S.~Savarese, and R.~Eustice, ``Automatic targetless
  extrinsic calibration of a 3d lidar and camera by maximizing mutual
  information,'' in \emph{Proceedings of the AAAI Conference on Artificial
  Intelligence}, vol.~26, no.~1, 2012.

\bibitem{ishikawa2018lidar}
R.~Ishikawa, T.~Oishi, and K.~Ikeuchi, ``Lidar and camera calibration using
  motions estimated by sensor fusion odometry,'' in \emph{2018 IEEE/RSJ
  International Conference on Intelligent Robots and Systems (IROS)}.\hskip 1em
  plus 0.5em minus 0.4em\relax IEEE, 2018, pp. 7342--7349.

\bibitem{john2015automatic}
V.~John, Q.~Long, Z.~Liu, and S.~Mita, ``Automatic calibration and registration
  of lidar and stereo camera without calibration objects,'' in \emph{2015 IEEE
  International Conference on Vehicular Electronics and Safety (ICVES)}.\hskip
  1em plus 0.5em minus 0.4em\relax IEEE, 2015, pp. 231--237.

\bibitem{cheng2019noise}
X.~Cheng, Y.~Zhong, Y.~Dai, P.~Ji, and H.~Li, ``Noise-aware unsupervised deep
  lidar-stereo fusion,'' in \emph{Proceedings of the IEEE/CVF Conference on
  Computer Vision and Pattern Recognition}, 2019, pp. 6339--6348.

\bibitem{zhang2005parameter}
L.~Zhang and S.~M. Seitz, ``Parameter estimation for mrf stereo,'' in
  \emph{2005 IEEE Computer Society Conference on Computer Vision and Pattern
  Recognition (CVPR'05)}, vol.~2.\hskip 1em plus 0.5em minus 0.4em\relax IEEE,
  2005, pp. 288--295.

\bibitem{hernandez2016embedded}
D.~Hernandez-Juarez, A.~Chac{\'o}n, A.~Espinosa, D.~V{\'a}zquez, J.~C. Moure,
  and A.~M. L{\'o}pez, ``Embedded real-time stereo estimation via semi-global
  matching on the gpu,'' \emph{Procedia Computer Science}, vol.~80, pp.
  143--153, 2016.

\bibitem{yamaguchi2014efficient}
K.~Yamaguchi, D.~McAllester, and R.~Urtasun, ``Efficient joint segmentation,
  occlusion labeling, stereo and flow estimation,'' in \emph{European
  Conference on Computer Vision}.\hskip 1em plus 0.5em minus 0.4em\relax
  Springer, 2014, pp. 756--771.

\bibitem{muresan2015improving}
M.~P. Muresan, M.~Negru, and S.~Nedevschi, ``Improving local stereo algorithms
  using binary shifted windows, fusion and smoothness constraint,'' in
  \emph{2015 IEEE International Conference on Intelligent Computer
  Communication and Processing (ICCP)}.\hskip 1em plus 0.5em minus 0.4em\relax
  IEEE, 2015, pp. 179--185.

\bibitem{spangenberg2014large}
R.~Spangenberg, T.~Langner, S.~Adfeldt, and R.~Rojas, ``Large scale semi-global
  matching on the cpu,'' in \emph{2014 IEEE Intelligent Vehicles Symposium
  Proceedings}.\hskip 1em plus 0.5em minus 0.4em\relax IEEE, 2014, pp.
  195--201.

\bibitem{zureiki2007stereo}
A.~Zureiki, M.~Devy, and R.~Chatila, ``Stereo matching using reduced-graph
  cuts,'' in \emph{2007 IEEE International Conference on Image Processing},
  vol.~1.\hskip 1em plus 0.5em minus 0.4em\relax IEEE, 2007, pp. I--237.

\bibitem{wang2021pvstereo}
H.~Wang, R.~Fan, P.~Cai, and M.~Liu, ``Pvstereo: Pyramid voting module for
  end-to-end self-supervised stereo matching,'' \emph{IEEE Robotics and
  Automation Letters}, vol.~6, no.~3, pp. 4353--4360, 2021.

\bibitem{cheng2020hierarchical}
X.~{Cheng}, Y.~{Zhong}, M.~{Harandi}, Y.~{Dai}, X.~{Chang}, H.~{Li},
  T.~{Drummond}, and Z.~{Ge}, ``Hierarchical neural architecture search for
  deep stereo matching,'' in \emph{Advances in Neural Information Processing
  Systems}, vol.~33, 2020, pp. 22\,158--22\,169.

\bibitem{chen2016transforming}
L.~Chen, Y.~He, J.~Chen, Q.~Li, and Q.~Zou, ``Transforming a 3-d lidar point
  cloud into a 2-d dense depth map through a parameter self-adaptive
  framework,'' \emph{IEEE Transactions on Intelligent Transportation Systems},
  vol.~18, no.~1, pp. 165--176, 2016.

\bibitem{hirata2019real}
A.~Hirata, R.~Ishikawa, M.~Roxas, and T.~Oishi, ``Real-time dense depth
  estimation using semantically-guided lidar data propagation and motion
  stereo,'' \emph{IEEE Robotics and Automation Letters}, vol.~4, no.~4, pp.
  3806--3811, 2019.

\bibitem{yang2019dense}
Y.~Yang, A.~Wong, and S.~Soatto, ``Dense depth posterior (ddp) from single
  image and sparse range,'' in \emph{Proceedings of the IEEE/CVF Conference on
  Computer Vision and Pattern Recognition}, 2019, pp. 3353--3362.

\bibitem{huber2011integrating}
H.~Badino, D.~Huber, T.~Kanade \emph{et~al.}, ``Integrating lidar into stereo
  for fast and improved disparity computation,'' in \emph{2011 International
  Conference on 3D Imaging, Modeling, Processing, Visualization and
  Transmission}.\hskip 1em plus 0.5em minus 0.4em\relax IEEE, 2011, pp.
  405--412.

\bibitem{maddern2016real}
W.~Maddern and P.~Newman, ``Real-time probabilistic fusion of sparse 3d lidar
  and dense stereo,'' in \emph{2016 IEEE/RSJ International Conference on
  Intelligent Robots and Systems (IROS)}.\hskip 1em plus 0.5em minus
  0.4em\relax IEEE, 2016, pp. 2181--2188.

\bibitem{park2018high}
K.~Park, S.~Kim, and K.~Sohn, ``High-precision depth estimation with the 3d
  lidar and stereo fusion,'' in \emph{2018 IEEE International Conference on
  Robotics and Automation (ICRA)}.\hskip 1em plus 0.5em minus 0.4em\relax IEEE,
  2018, pp. 2156--2163.

\bibitem{choe2021volumetric}
J.~Choe, K.~Joo, T.~Imtiaz, and I.~S. Kweon, ``Volumetric propagation network:
  Stereo-lidar fusion for long-range depth estimation,'' \emph{IEEE Robotics
  and Automation Letters}, vol.~6, no.~3, pp. 4672--4679, 2021.

\bibitem{park2019high}
K.~Park, S.~Kim, and K.~Sohn, ``High-precision depth estimation using
  uncalibrated lidar and stereo fusion,'' \emph{IEEE Transactions on
  Intelligent Transportation Systems}, vol.~21, no.~1, pp. 321--335, 2019.

\bibitem{newcombe2011dtam}
R.~A. Newcombe, S.~J. Lovegrove, and A.~J. Davison, ``Dtam: Dense tracking and
  mapping in real-time,'' in \emph{International Conference on Computer Vision
  (ICCV)}.\hskip 1em plus 0.5em minus 0.4em\relax IEEE, 2011, pp. 2320--2327.

\bibitem{yedidia2000generalized}
J.~S. Yedidia, W.~T. Freeman, Y.~Weiss \emph{et~al.}, ``Generalized belief
  propagation,'' in \emph{NIPS}, vol.~13, 2000, pp. 689--695.

\bibitem{menze2018object}
M.~Menze, C.~Heipke, and A.~Geiger, ``Object scene flow,'' \emph{ISPRS Journal
  of Photogrammetry and Remote Sensing}, vol. 140, pp. 60--76, 2018.

\bibitem{fischler1981random}
M.~A. Fischler and R.~C. Bolles, ``Random sample consensus: a paradigm for
  model fitting with applications to image analysis and automated
  cartography,'' \emph{Communications of the ACM}, vol.~24, no.~6, pp.
  381--395, 1981.

\bibitem{chambolle2011first}
A.~Chambolle and T.~Pock, ``A first-order primal-dual algorithm for convex
  problems with applications to imaging,'' \emph{Journal of mathematical
  imaging and vision}, vol.~40, no.~1, pp. 120--145, 2011.

\bibitem{radu20113d}
R.~B. Rusu and S.~Cousins, ``{3D is here: Point Cloud Library (PCL)},'' in
  \emph{{IEEE International Conference on Robotics and Automation (ICRA)}},
  Shanghai, China, May 9-13 2011.

\bibitem{uhrig2017sparsity}
J.~Uhrig, N.~Schneider, L.~Schneider, U.~Franke, T.~Brox, and A.~Geiger,
  ``Sparsity invariant cnns,'' in \emph{International Conference on 3D Vision
  (3DV)}.\hskip 1em plus 0.5em minus 0.4em\relax IEEE, 2017, pp. 11--20.

\end{thebibliography}
\bibliographystyle{IEEEtran}
\begin{IEEEbiography}[{\includegraphics[width=1in,height=1.25in,clip,keepaspectratio]{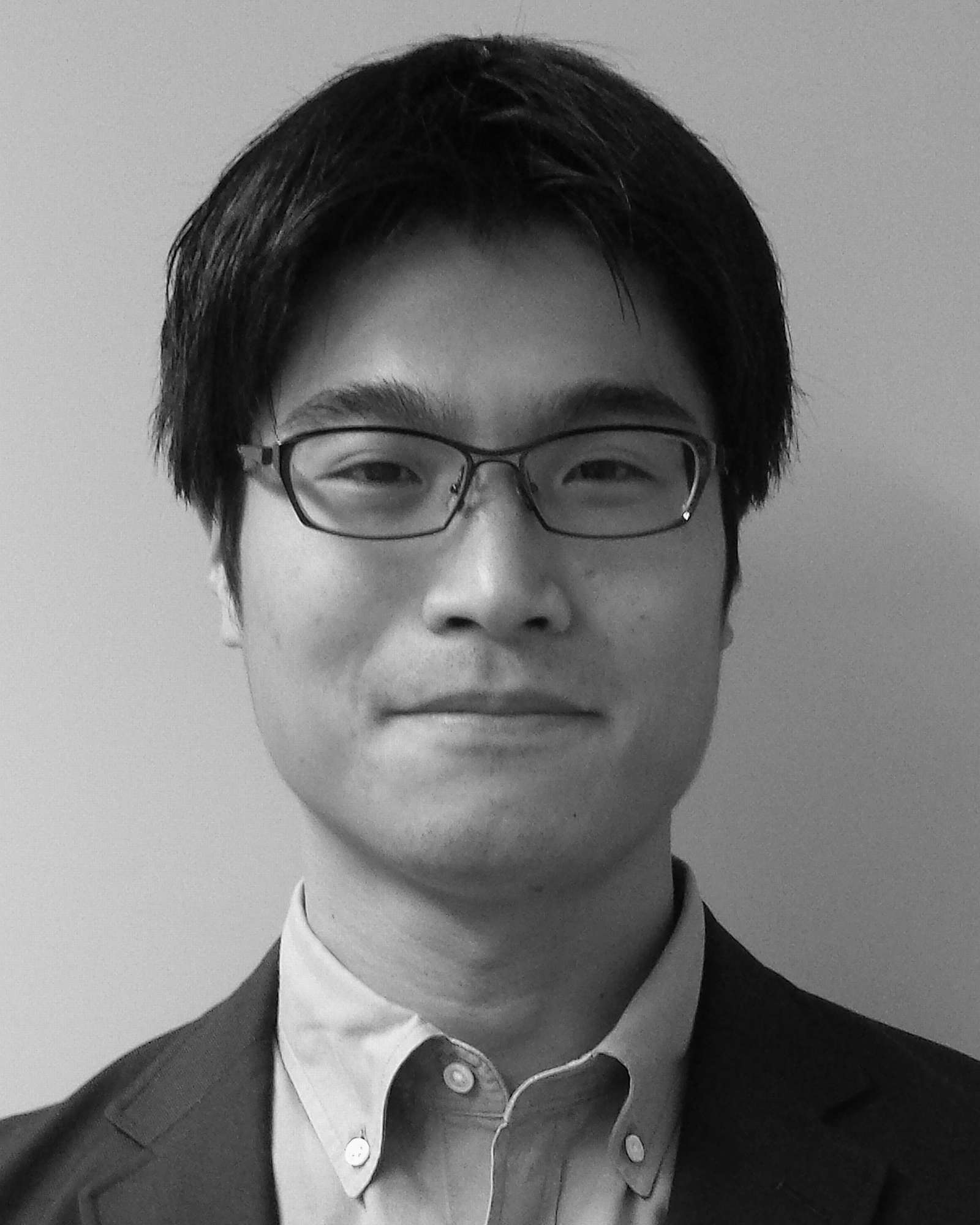}}]{Yasuhiro Yao}
received the B.S and the M.E. degree from the University of Tokyo, Japan in 2007, 2010, respectively.
In 2010, He joined NTT as a researcher.
From 2013 to 2016, he was a cloud solution architect at Dimension Data APAC, Singapore.
He is currently a Senior Research Engineer at NTT Human Informatics Laboratories and also pursuing a Ph.D. in Information Studies at the University of Tokyo.
His research interests include computer vision and sensor fusion.
\end{IEEEbiography}
\begin{IEEEbiography}[{\includegraphics[width=1in,height=1.25in,clip,keepaspectratio]{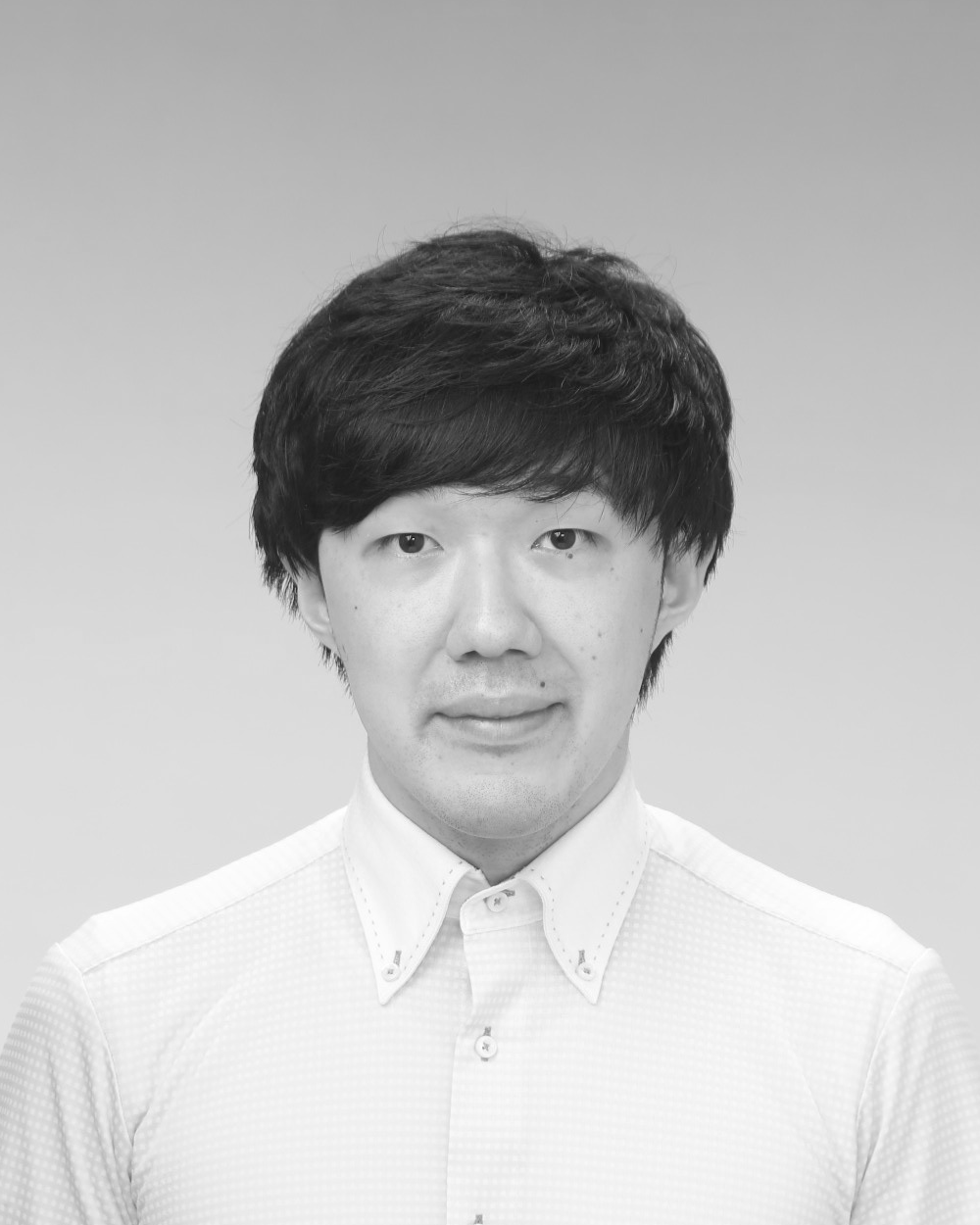}}]{Ryoichi Ishikawa}
received the B.E. degree from the Department of Electrical Engineering, The University of Tokyo, Japan, in 2014, and the M.E. and Ph.D. degrees in Electrical Engineering and Information Systems, The University of Tokyo, Japan, in 2016 and 2019, respectively.
From 2019, he is a Project Researcher at the Institute of Industrial Science, The University of Tokyo.
His research interests include robot vision, sensor fusion, and calibration.
\end{IEEEbiography}
\begin{IEEEbiography}[{\includegraphics[width=1in,height=1.25in,clip,keepaspectratio]{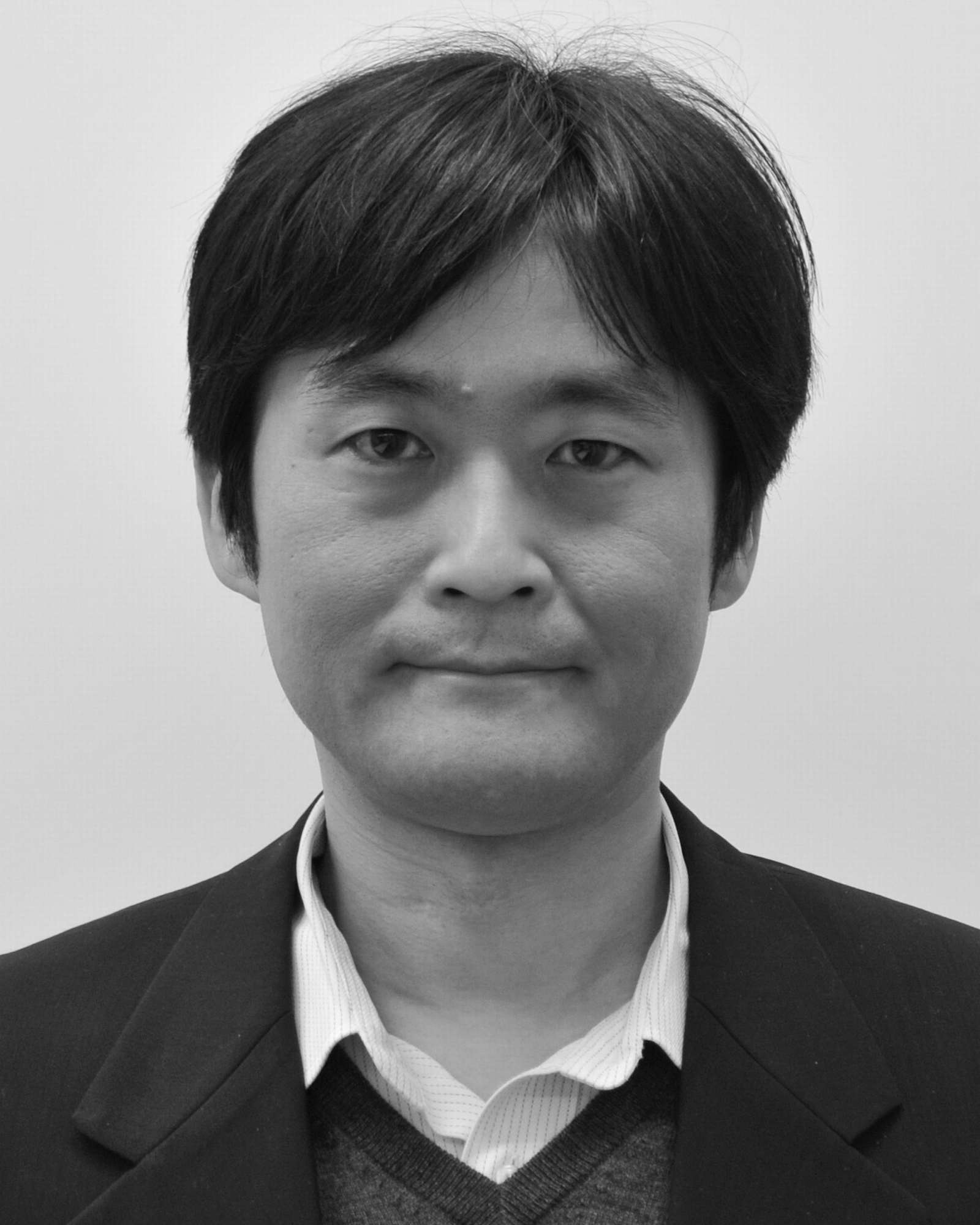}}]{Shingo Ando}
received a B.E. in electrical engineering from Keio University, Kanagawa, in 1998 and a Ph.D. in engineering from Keio University in 2003.
He joined NTT in 2003.
He has been engaged in research and practical application development in the fields of image processing, pattern recognition, and digital watermarks.
He is a member of IEICE and the Institute of Image Information and Television Engineers.
\end{IEEEbiography}
\begin{IEEEbiography}[{\includegraphics[width=1in,height=1.25in,clip,keepaspectratio]{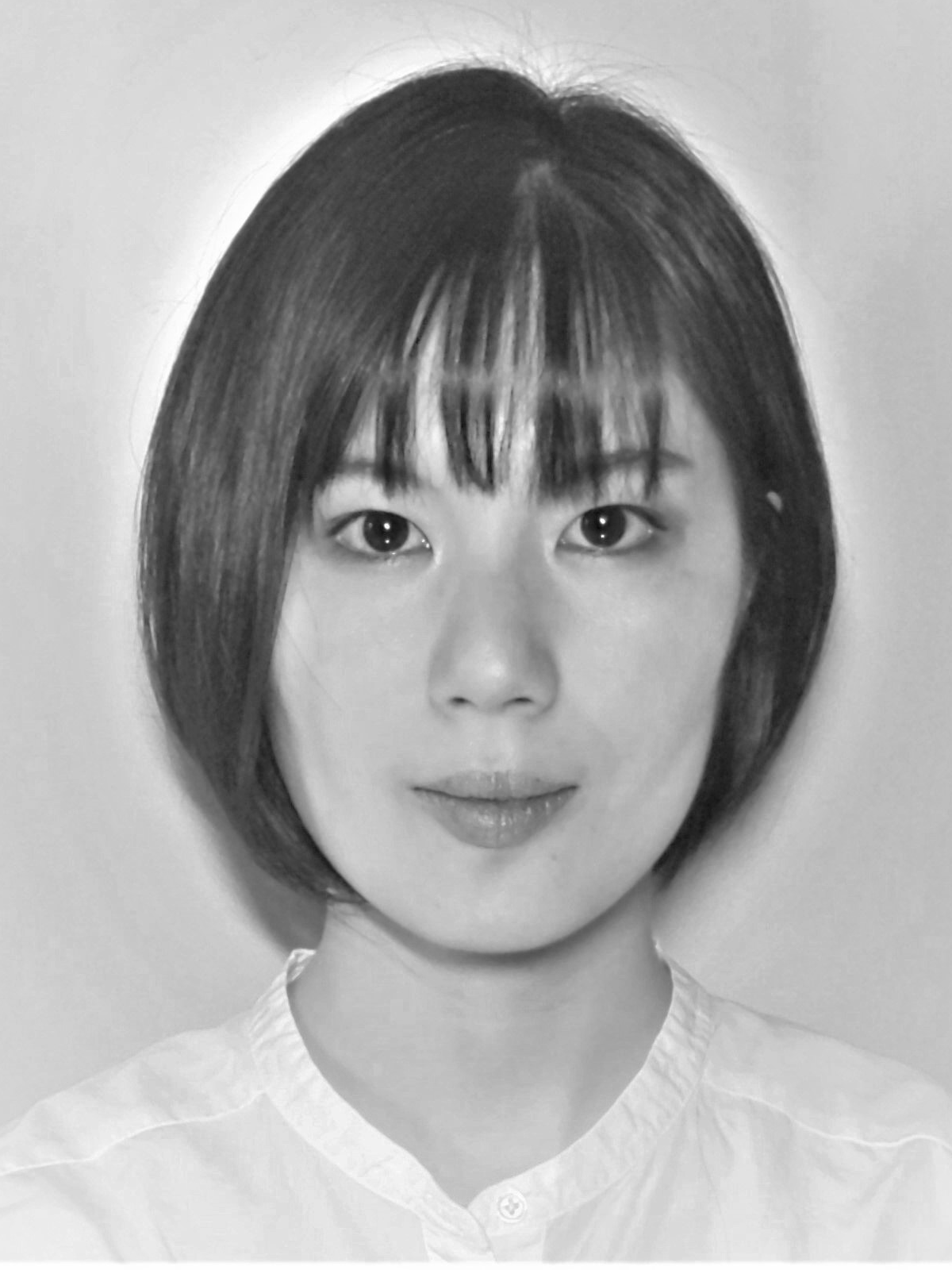}}]{Kana Kurata}
received a B.S. in earth and planetary sciences from the Nagoya University in 2016 and an M.E. in environmental studies from the Nagoya University in 2018.
She joined NTT in 2018.
She has been engaged in research in fields of computer vision and pattern recognition.
\end{IEEEbiography}
\begin{IEEEbiography}[{\includegraphics[width=1in,height=1.25in,clip,keepaspectratio]{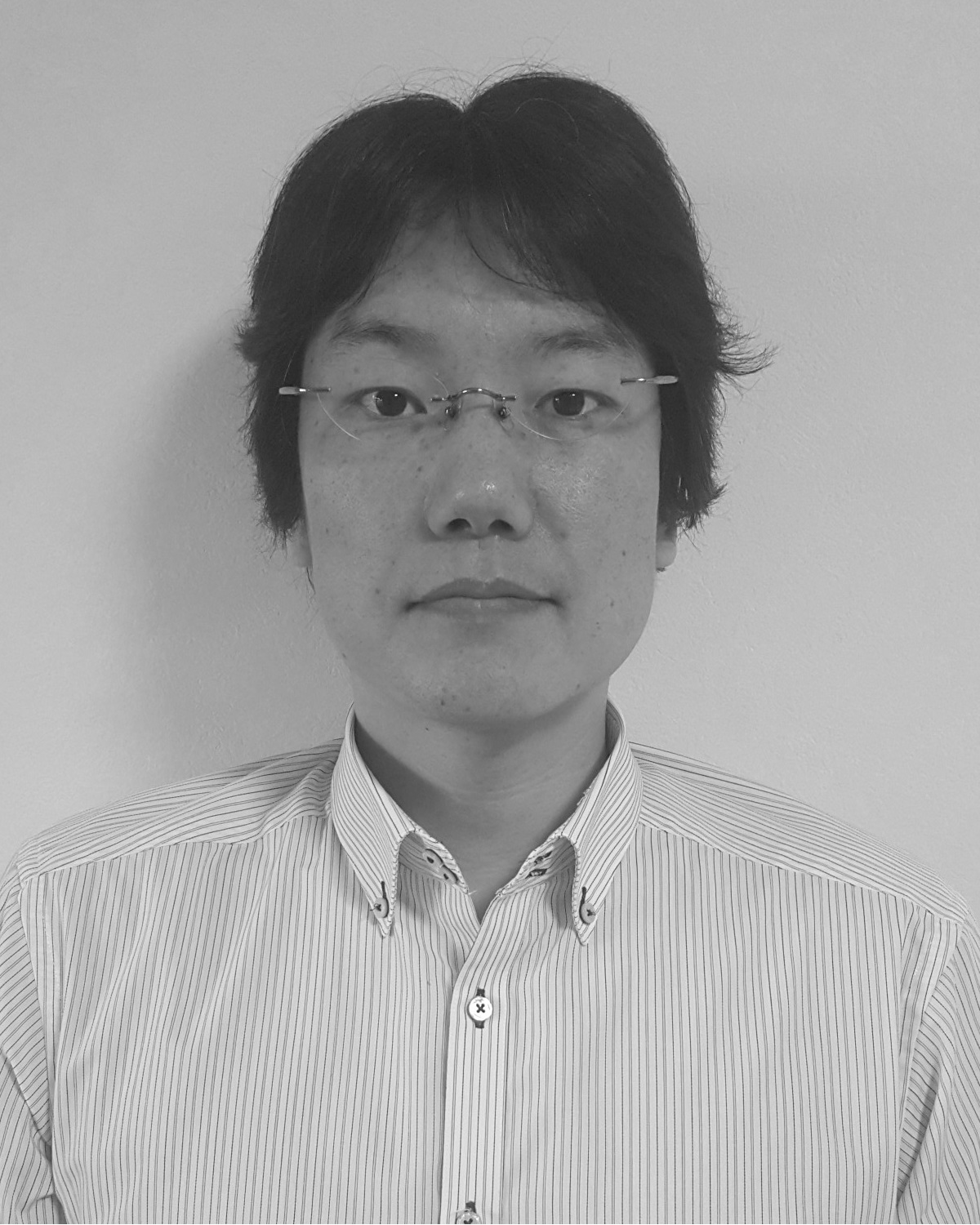}}]{Naoki Ito}
received the M.E. from Toyohashi University of Technology, Aichi, in 2001.
He joined NTT in 2001 and engaged in research on character recognition.
He moved to NTT EAST in 2004 and engaged in the development of security systems.
He moved to NTT Cyber Space Laboratories in 2008.
He has been working on the development of real-world digitalization technologies.
He is currently a senior research engineer at NTT Human Informatics Laboratories.
\end{IEEEbiography}
\begin{IEEEbiography}[{\includegraphics[width=1in,height=1.25in,clip,keepaspectratio]{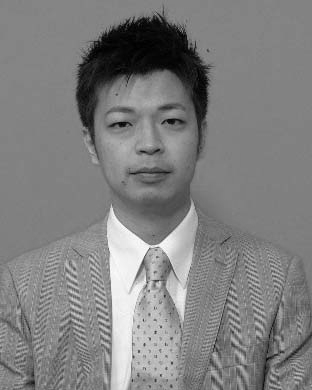}}]{Jun Shimamura}
received a B.E. in engineering science from Osaka University in 1998 and an M.E. and Ph.D. from Nara Institute of Science and Technology in 2000 and 2006.
He joined NTT Cyber Space Laboratories in 2000.
He is currently senior research engineer, supervisor of scene analysis technology at NTT Human Informatics Laboratories, Japan.
His research interests include computer vision and mixed reality.
\end{IEEEbiography}
\begin{IEEEbiography}[{\includegraphics[width=1in,height=1.25in,clip,keepaspectratio]{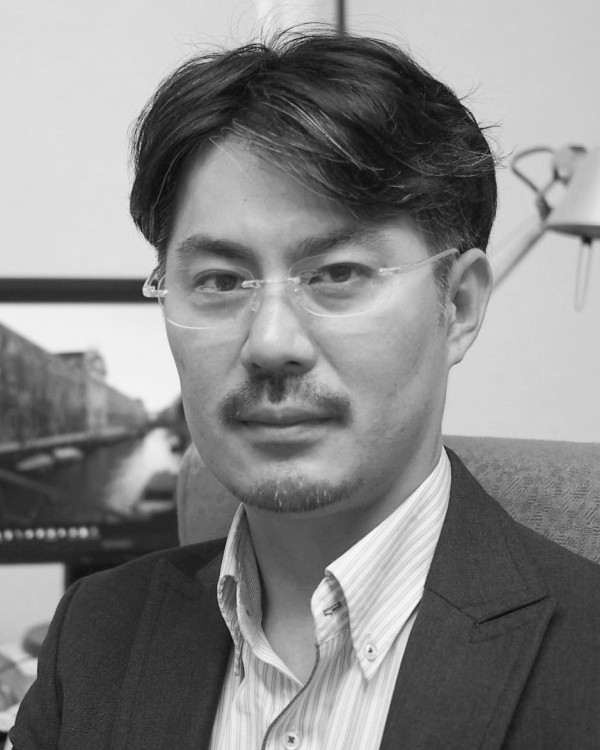}}]{Takeshi Oishi}
received the degree of B.Eng. in Electrical Engineering from Keio University in 1999, and the Ph.D. degree in Interdisciplinary Information Studies from the University of Tokyo in 2005.
He is currently an Associate Professor at Institute of Industrial Science, the University of Tokyo. His research interests are in 3D modeling from reality, digital archiving of cultural heritage assets and mixed/augmented reality.
\end{IEEEbiography}

\end{document}


\title{Supplementary to Non-learning Stereo-aided Depth Completion under Mis-projection via Selective Stereo Matching
}

\maketitle
\renewcommand{\thesection}{A-\arabic{section}}
\renewcommand{\thefigure}{A-\arabic{figure}}
\renewcommand{\thetable}{A-\arabic{table}}
\renewcommand{\theequation}{A-\arabic{equation}}

\setlength{\fboxsep}{0pt}
\begin{figure}[tb]
\centering
\subfloat[Left image]{\includegraphics[width=40mm]{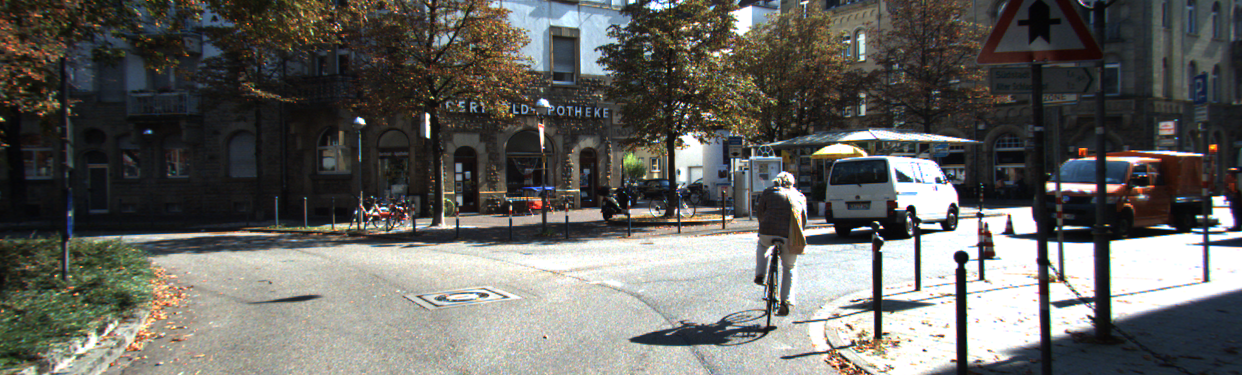}
}
\hfil
\subfloat[Right image]{\includegraphics[width=40mm]{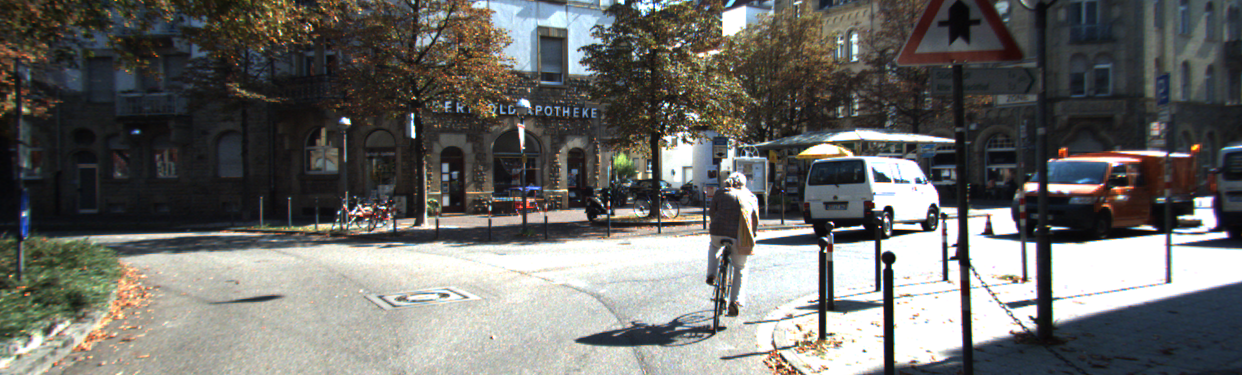}
}
\vfil
\vspace{-8pt}
\subfloat[Sparse depth map]{\fbox{\includegraphics[width=40mm]{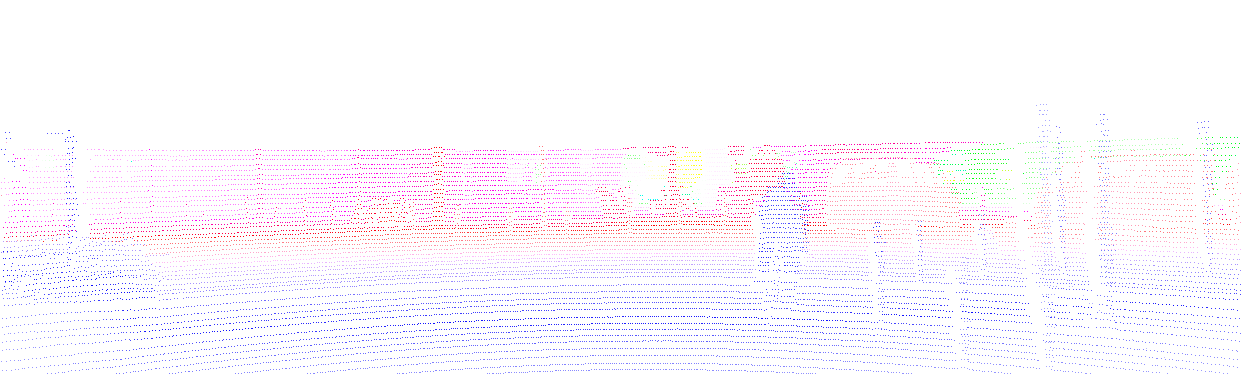}}
}
\hfil
\subfloat[Dense disparity map ]{\fbox{\includegraphics[width=40mm]{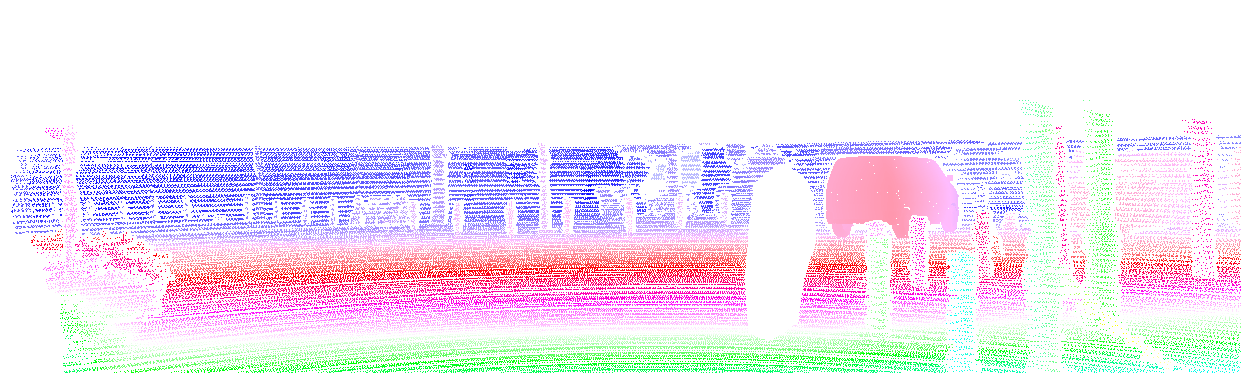}}
}
\vfil
\vspace{-8pt}
\subfloat[Dense depth map]{\fbox{\includegraphics[width=40mm]{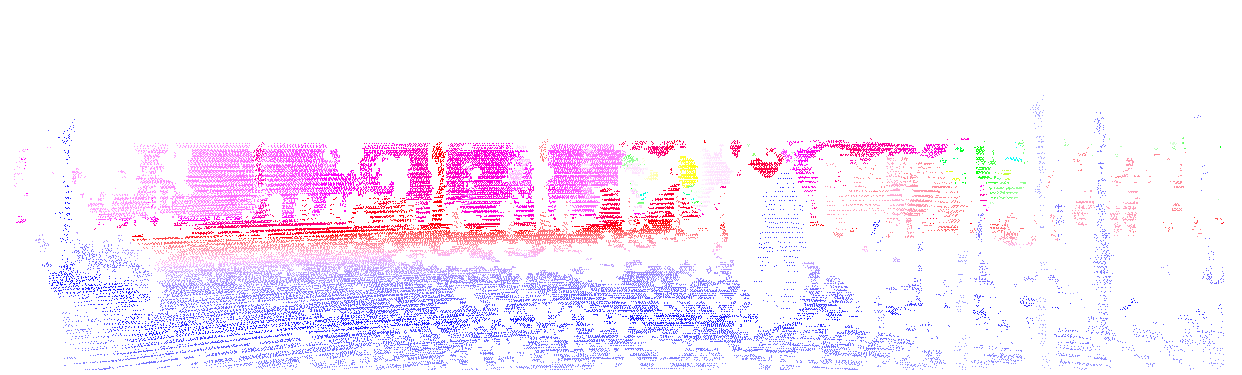}}
}

\caption{
A frame of the KITTI dataset.
(d) The ground truth dense disparity maps are from the stereo 2015 benchmark 
(e) The ground truth dense depth maps are from the depth completion benchmark 
}
\label{fig:kitti}
\end{figure}

\setlength{\fboxsep}{0pt}
\begin{figure}[t]
\centering
\subfloat[Image 1]{\includegraphics[width=40mm]{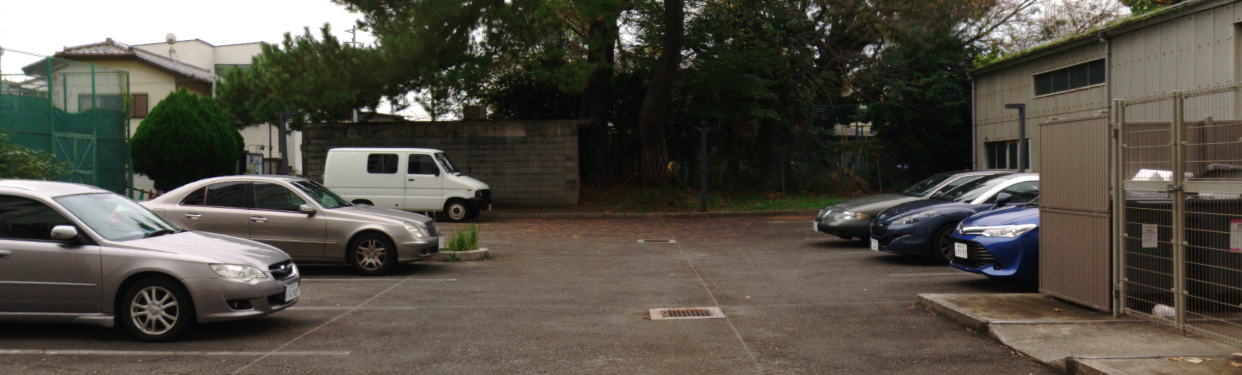}
}
\hfil
\subfloat[Image 2]{\includegraphics[width=40mm]{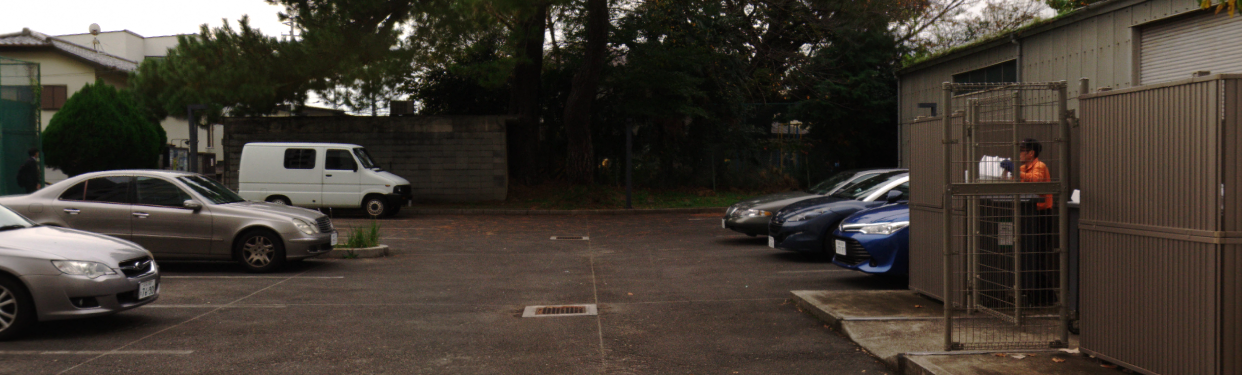}
}
\vfil
\vspace{-8pt}
\subfloat[Dense depth map]{\fbox{\includegraphics[width=40mm]{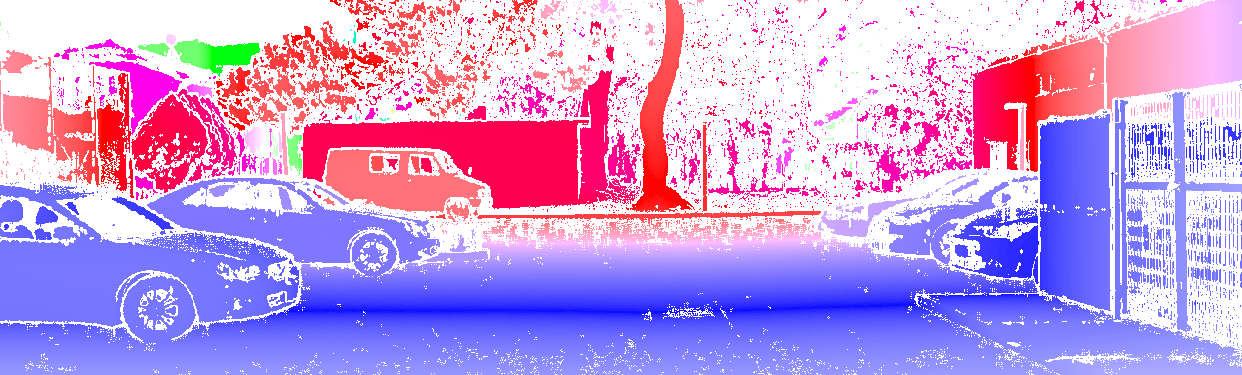}}}
\hfil
\subfloat[\textit{lines-16}]{\fbox{\includegraphics[width=40mm]{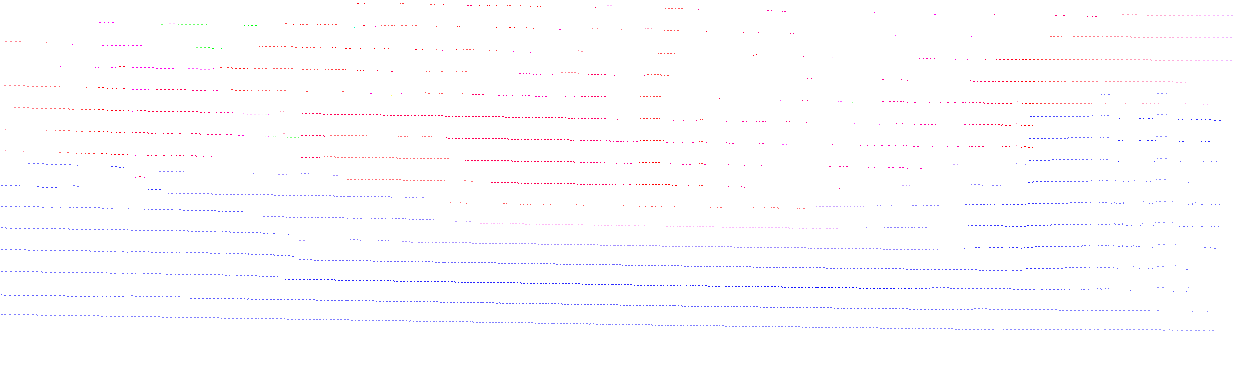}}}
\vfil
\vspace{-8pt}
\subfloat[\textit{lines-32}]{\fbox{\includegraphics[width=40mm]{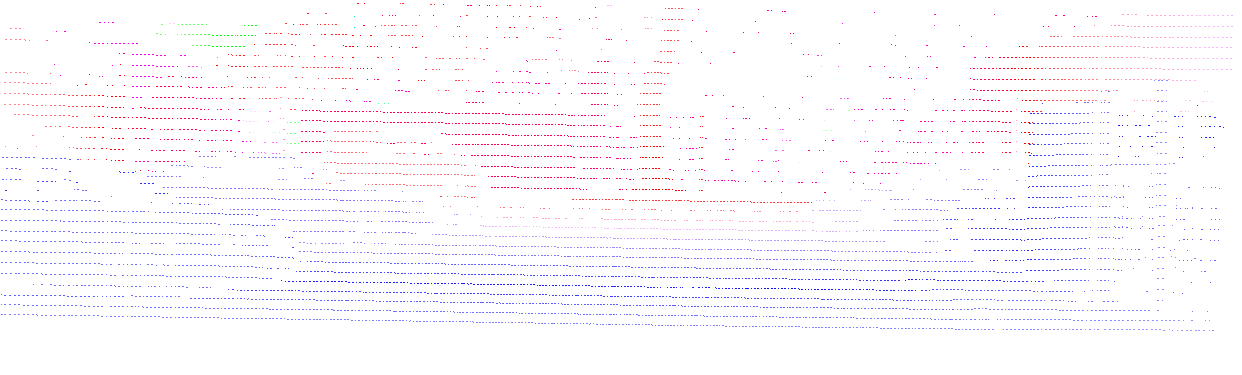}}}
\hfil
\subfloat[\textit{lines-64}]{\fbox{\includegraphics[width=40mm]{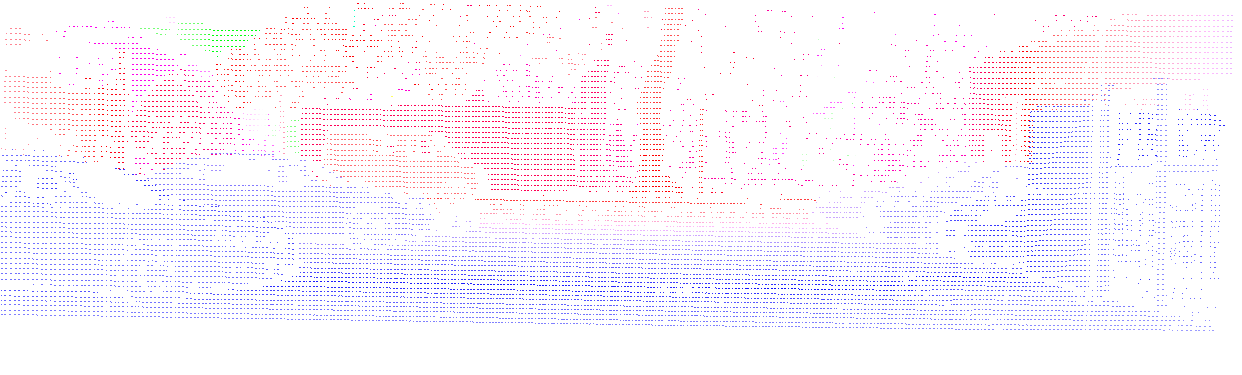}}
}

\caption{
A frame of the Komaba dataset.
}
\label{fig:komaba}
\end{figure}

\section{Dataset Figures}
Here, we visually show the examples of the dataset: KITTI in Fig. \ref{fig:kitti} and Komaba in Fig. \ref{fig:komaba}.

\begin{table}[t]
    \centering
    \caption{Errors in extrinsic parameters}
    \begin{tabular}{c|cc|cc}
    \hline
    &  \multicolumn{2}{c}{Avg. rot. error [deg.]} & \multicolumn{2}{c}{Avg. trans. error [m]} \\
    \cline{2-5}
    Cost &  
    \textit{error-1} &  \textit{error-2} &\textit{error-1} &  \textit{error-2} \\
    \hline
    $L^{\mathbf{x}}(d)$ &  \textbf{0.561} & 1.813 & 0.156 & 0.379\\
    $L^{\mathbf{x}}(d)+L^{\mathbf{x}}_B$ & 0.675 & \textbf{0.667} & \textbf{0.155} & \textbf{0.207}\\
    \hline
    \end{tabular}
    \label{tab:calib_errors}
\end{table}

\section{LiDAR stereo extrinsic calibration}
\label{sec:calib_method}
Here, we explain the single-frame-marker-less calibration method to create \textit{error-1} and \textit{error-2} in our evaluation with calibration errors.
We intended to emulate the maximum error expected in practical situations.
Hence, we extrinsically calibrated LiDAR and camera by using the same inputs as our depth completion, which are a single pair of images and a single scan of LiDAR.

Specifically, we search the rotation matrix $R^{\ast}$ and translation vector $\mathbf{t}^{\ast}$ to minimize the cost $L$ from the candidate set of rotation matrices $\mathcal{R} = \left\{R_i| i=0,1\,\ldots\, N_R -1 , R_i \in SO(3) \right\}$ and translation vectors $\mathcal{T} = \left\{ \mathbf{t}_i | i=0,1\,\ldots\,N_{\mathbf{t}}-1, \mathbf{t}_i \in \mathbb{R}^3 \right\}$ in a brute force manner with $N_R$ and $N_{\mathbf{t}}$ being the number of candidates, where $\mathcal{R}$ and $\mathcal{T}$ are generated around the initial values via grid sampling.
\begin{equation}
    R^{\ast}, \mathbf{t}^{\ast} = \argmin_{R \in \mathcal{R}, \mathbf{t} \in \mathcal{T}}{L}
\end{equation}

For each combination of $R\in \mathcal{R}$ and $\mathbf{t}\in \mathcal{T}$, we first project each LiDAR point $\mathbf{p} \in P$ to the left image by applying $R$ and $\mathbf{t}$ to derive set of the image locations and the depth of the point ($P^{R\mathbf{t}}$).
\begin{align}
&P^{R\mathbf{t}} = \left\{\left(x_0, x_1, d\right)^T\ |\ 0\leq x_0 < W, 0 \leq x_1 < H, d > 0 \right\} \mathrm{,}  \nonumber \\
    &\mathrm{where}
    \left\{
    \begin{array}{ll}
         \left(X_0, X_1, D\right)^T =  K\left( R  \mathbf{p} + \mathbf{t}\right) & \\
         x_0 = X_0 / D & \mathrm{for}\ \mathbf{p} \in P\\
         x_1 = X_1 / D & \\
         d = b f / D
    \end{array}
    \right.
    \label{eq:proj}
\end{align}
Here, $W$ and $H$ are the width and height of $I_1$, respectively.

To construct the cost, we introduce the $L^{\mathbf{x}}_B$ term to avoid LiDAR points to be projected on the areas where no object exists.
With using the notation of the stereo matching cost $L^{\mathbf{x}}(d)$ in Eq. (8) of the main paper, $L^{\mathbf{x}}_B$ is defined as follows:
\begin{equation}
\label{eq:loss_bg}
    L^{\mathbf{x}}_B = - \min{\left(L^\mathbf{x}(0), l_B\right)}.
\end{equation}
Note that $L^{\mathbf{x}}_B$ increases if a LiDAR point is projected onto an image location that appears to have infinite depth (or zero disparity).

Then, the cost $L_{\mathrm{calib}}$ for calibration is defined as the sum of $L^{\mathbf{x}}(d)$ and $L^{\mathbf{x}}_B$ over points in $P^{R \mathbf{t}}$.
\begin{equation}
\label{eq:loss_calib}
    L_{\mathrm{calib}} = \sum_{\left(\mathbf{x}, d\right)^T \in P^{R{\mathbf{t}}}} {L^{\mathbf{x}}(d) + L_B^{\mathbf{x}}}
\end{equation}

Table \ref{tab:calib_errors} shows the extrinsic parameter errors after calibration with or without the $L^{\mathbf{x}}_B$ term.
The results indicate that the $L^{\mathbf{x}}_B$ term has little effect when the initial calibration error is small as in \textit{error-1}.
However, the $L^{\mathbf{x}}_B$ term stabilizes calibration when the initial error is large as in \textit{error-2}.
